\documentclass[final]{cvpr}

\usepackage{times}
\usepackage{epsfig}
\usepackage{graphicx}
\usepackage{amsmath}
\usepackage{amssymb}
\usepackage{mathrsfs}
\usepackage{multirow}
\usepackage{float}

\usepackage[symbol]{footmisc}

\usepackage[table,xcdraw]{xcolor}
\usepackage{caption}

\usepackage[pagebackref=true,breaklinks=true,colorlinks,bookmarks=false]{hyperref}

\def\cvprPaperID{9907} 
\def\confYear{CVPR 2021}

\newcommand{\sota}{state-of-the-art\xspace}
\usepackage{pifont}%
\newcommand{\cmark}{\ding{51}}%
\newcommand{\xmark}{\ding{55}}%

\title{$\mathbb{X}$Resolution Correspondence Networks}

\author{Georgi Tinchev\footnotemark\\
	Oxford Robotics Insitute\\
	University of Oxford\\
	{\tt\small gtinchev@robots.ox.ac.uk}
	\and
	Shuda Li\\
	XYZ Reality\\
	{\tt\small shuda.li@xyzreality.com}
	\and
	Kai Han\\
	Visual Geometry Group\\
	University of Oxford\\
	{\tt\small khan@robots.ox.ac.uk}
	\and
	David Mitchell\\
	XYZ Reality\\
	{\tt\small david.mitchell@xyzreality.com}
	\and
	Rigas Kouskouridas\\
	XYZ Reality\\
	{\tt\small rigas.kousk@xyzreality.com}
	\and
	{\small\url{https://xyz-r-d.github.io/xrcnet}}
}

\begin{document}

	\maketitle


	\begin{abstract}
		In this paper, we aim at establishing accurate dense correspondences between a pair of images with overlapping field of 
		view under challenging illumination variation, viewpoint changes, and style differences. Through an extensive ablation study 
		of the \sota correspondence networks, we surprisingly discovered that the widely adopted 4D correlation tensor and its 
		related learning and processing modules could be de-parameterised and removed from training with merely a minor impact 
		over the final matching accuracy. Disabling these computational expensive modules dramatically speeds up the training 
		procedure and allows to use 4 times bigger batch size, which in turn compensates for the accuracy drop. Together with a 
		multi-GPU inference stage, our method facilitates the systematic investigation of the relationship between matching 
		accuracy and up-sampling resolution of the native testing images from 1280 to 4K. This leads to discovery of the existence 
		of an optimal resolution $\mathbb X$ that produces accurate matching performance surpassing the \sota methods 
		particularly over the lower error band on public benchmarks for the proposed network.
	\end{abstract}
	
	\begin{figure*}[h]
		\centering
		\includegraphics[width=1\textwidth]{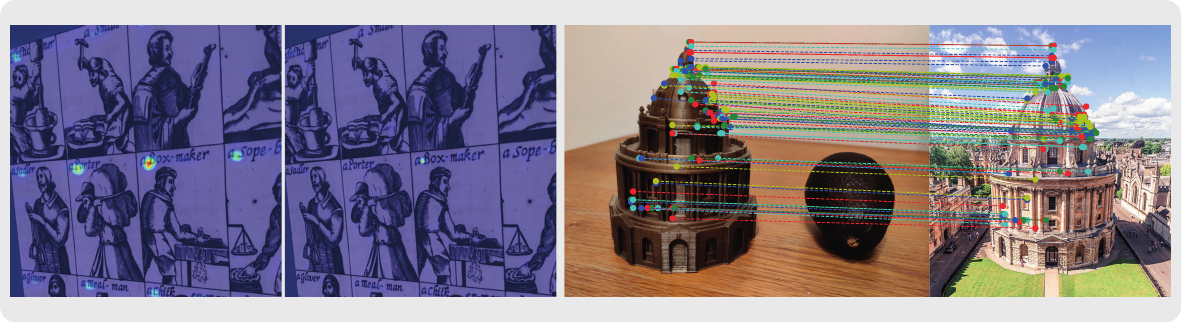}
		\caption{$\mathbb{X}$RCNet highlights: \textbf{Left tuple}: Using high resolution feature maps allows $\mathbb X$RCNet 
		to accurately pinpoint the target key point location given a query point in the source image. Note that the image contains 
		challenging repetitive patterns of the letter 'a' that confuse the network when using a low resolution image (left most). 
		\textbf{Right}: $\mathbb{X}$RCNet without using any geometric constraints is capable of producing \sota matching 
		accuracy and reliably match under extreme illumination and style differences (best viewed in PDF with zoom).}
		\vspace{-14pt}
		\label{fig:teaser}
	\end{figure*}

	
	\section{Introduction}
	
	\renewcommand{\thefootnote}{\fnsymbol{footnote}}
	\footnotetext[1]{Work done while on internship with XYZ Reality.}
	
	Establishing dense image correspondences is a fundamental problem for many computer vision applications, from 
	Structure-from-Motion (SfM)~\cite{Schonberger_CVPR16_SfSRevisited,Schonberger_ECCV16_PixelwiseViewSelection},   visual 
	Simultaneous Localisation and Mapping (SLAM)~\cite{Xinghui_NeurIPS20_DRC} to image 
	retrieval~\cite{Vassileios_CVPR20_UR2KiD}, image style transfer~\cite{VGG_NeurIPS20_D2D}, and scene 
	understanding~\cite{Oxford_CVPR20_ANCNet}. Traditionally, point correspondences between a pair of images are found 
	through a sparse detection-description and matching pipeline. Particularly, a key point 
	detector~\cite{Rosten_PAMI10_Fast,Gauglitz_IJCV11_EvaluateDetectors} is first used to collect a set of sparse interest points 
	from input images while a feature descriptor~\cite{Lowe_IJCV04_SIFT, Bay_ECCV06_SURF, Leutenegger_ICCV11_BRISK} 
	extracts a unique description of a local image patch centred at the detected key point location. In the end, the point 
	correspondences between the query image and the reference one are calculated by searching the candidate matching pairs for 
	small descriptor distances or using the ratio test between the best and the second best matches~\cite{Lowe_IJCV04_SIFT}. 
	
	In the last few years we have witnessed a dramatic improvement over all stages of the sparse correspondence pipeline mostly 
	using machine learning~\cite{MagicLeap_CVPR18_SuperPoint, Dusmanu_CVPR19_D2Net,Revaud_NeurIPS19_R2D2, 
	Vassilieos_CVPR19_SOSNet, Revaud_NeurIPS19_R2D2}. In addition to the feature detectors and descriptors, the matching 
	stage has also been extensively studied and new algorithms taking into account both inter-image and intra-image constraints 
	make the matching stage more reliable than before~\cite{MagicLeap_CVPR20_SuperGlue, 
	Luo_CVPR20_ASLFeat,Vassilieos_CVPR19_SOSNet, Zhang_ICCV19_OrderAwareCorrespondenceNet}. However, sparse 
	correspondence methods are not straightforward to be adapted to produce per pixel matches which are often required for 
	image warping, style transfer, or dense 3D reconstruction. A naive extension from sparse methods, for example, is to densely 
	extract feature descriptors and use brute force matching. However, this is prohibitively expensive for high resolution images. 
	Furthermore, to achieve the best performance, the detection-description and matching pipeline typically requires each stage to 
	be trained separately, which introduces extra difficulties when being deployed to new sensory data. For example, the top 
	performer on the visual localisation benchmark~\cite{Sattler_IJCV20_VisualLocalisationBenchMark} combines the SuperPoint 
	(SP)~\cite{MagicLeap_CVPR18_SuperPoint} and SuperGlue (SG)~\cite{MagicLeap_CVPR20_SuperGlue} and the SP 
	detector-descriptor has to be trained separately with SG.  
	
	In contrast, the dense correspondence methods~\cite{Liu_PAMI11_SiftFlow} and particularly Deep Correspondence Networks 
	(DCN)~\cite{Choy_NIPS16_UCN,Noh_ICCV17_DELF,Zanfir_CVPR18_DeepGM,VGG_NeurIPS20_D2D,Rocco_NIPS18_NCNet,Rocco_ECCV20_SparseNC}
	 that emerged in recent years, represent a highly competitive alternative for their capability of producing good quality per pixel 
	correspondences. DCNs also unify the detection-description and matching pipeline into one single architecture using standard 
	feature backbones such that it can be trained end-to-end. Moreover, DCNs are shown to be able to quickly adapt to images of 
	high resolution or larger feature maps while being deployed into consumer 
	products~\cite{Rocco_NIPS18_NCNet,Rocco_ECCV20_SparseNC,Xinghui_NeurIPS20_DRC,VGG_NeurIPS20_D2D,Lepetit_ECCV20s_S2DNet}.

	In this paper, we present a novel dense correspondence methodology that is capable of processing high resolution images 
	and produce reliable and highly accurate matching results as shown in Fig.~\ref{fig:teaser}. More importantly, light-weight 
	correspondence networks allows us to investigate an intriguing questions for all DCNs: The widely adopted testing protocol of 
	up-sampling the testing images seems always beneficial for improving matching accuracy. up-sampling the testing image 
	always lead to higher accuracy? If not, Does it exist an optimal resolution $\mathbb X$ given testing images in a dataset? In 
	this work, we introduce $\mathbb X$Resolution Correspondence Network ($\mathbb X$RCNet), a light-weight architecture 
	designed to answer these questions while achieving \sota performance. 
	
	Our work is directly inspired by the recently introduced strategy of using extensive ablation studies to either achieve more 
	accurate visual representations ~\cite{Google_010720_SimCLR} or highly impactful training procedures or architecture 
	refinements that improve model accuracy~\cite{He_CVPR19_BagOfTricks}. Approaching the dense correspondence problem 
	with the same strategy, we start by carrying out extensive ablation studies with various training configurations over the \sota 
	dense correspondence networks and made several key observations. First, the widely adopted 4D correlation tensor and its 
	related filtering modules~\cite{Rocco_NIPS18_NCNet,Rocco_ECCV20_SparseNC, 
	VGG_NeurIPS20_D2D,Lepetit_ECCV20s_S2DNet} can be de-parameterised and even removed from the training stage at the 
	cost of a small drop in accuracy. Second, switching to a much shallower feature backbone also has limited impact to the 
	overall matching results. Third, the combination of the first two discoveries results in the  light-weight $\mathbb X$RCNet. 
	During the training stage, $\mathbb X$RCNet enjoys a significantly smaller memory footprint and much faster speed than the 
	\sota methods (see Tab.~\ref{tab:ablation}). This allows us to use 4 times larger batch size and increase the number of epochs 
	within roughly the same amount of training time using the same hardware. The latter compensates for the slight deterioration 
	of the accuracy levels. When combined with a multi-GPU inference method, it allows us to evaluate the matching accuracy of 
	$\mathbb{X}$RCNet using image size up to 4K, by which we discover the existence of an optimal up-sampling resolution for 
	$\mathbb{X}$RCNet to achieve the best accuracy. Interestingly, increasing the resolution is not always beneficial possibly 
	because the relative size of the receptive field to the image might decrease, which then renders the network prediction less 
	accurate. 
	Overall, the contributions of the paper can be summarised as follow:
	\begin{itemize}
		\item We carried out extensive tests and a thorough ablation study over the \sota DCNs and made several key observations 
		that lead to the introduction of a simple and light-weight multi-resolution neural network architecture named as the 
		$\mathbb{X}$Resolution Network~($\mathbb{X}$RCNet).
		\item $\mathbb{X}$RCNet is capable of training with much larger batch size and faster per image learning speed. During 
		inference, $\mathbb X$RCNet can take in images with higher resolution than most of the previous work and allows us to 
		search for the optimal resolution $\mathbb X$ to up-sample the testing image for a correspondence task. 
		\item $\mathbb{X}$RCNet achieves \sota accuracy on two challenging datasets --- 
		HPatches~\cite{Vassilieios_CVPR17_HPatches} and InLoc~\cite{taira2018inloc}, while performing competitively on Aachen 
		Day-Night~\cite{Sattler_2018_CVPR,Sattler2012BMVC}. 
	\end{itemize}
	\begin{figure*}[h]
		\centering
		\includegraphics[width=1.0\textwidth]{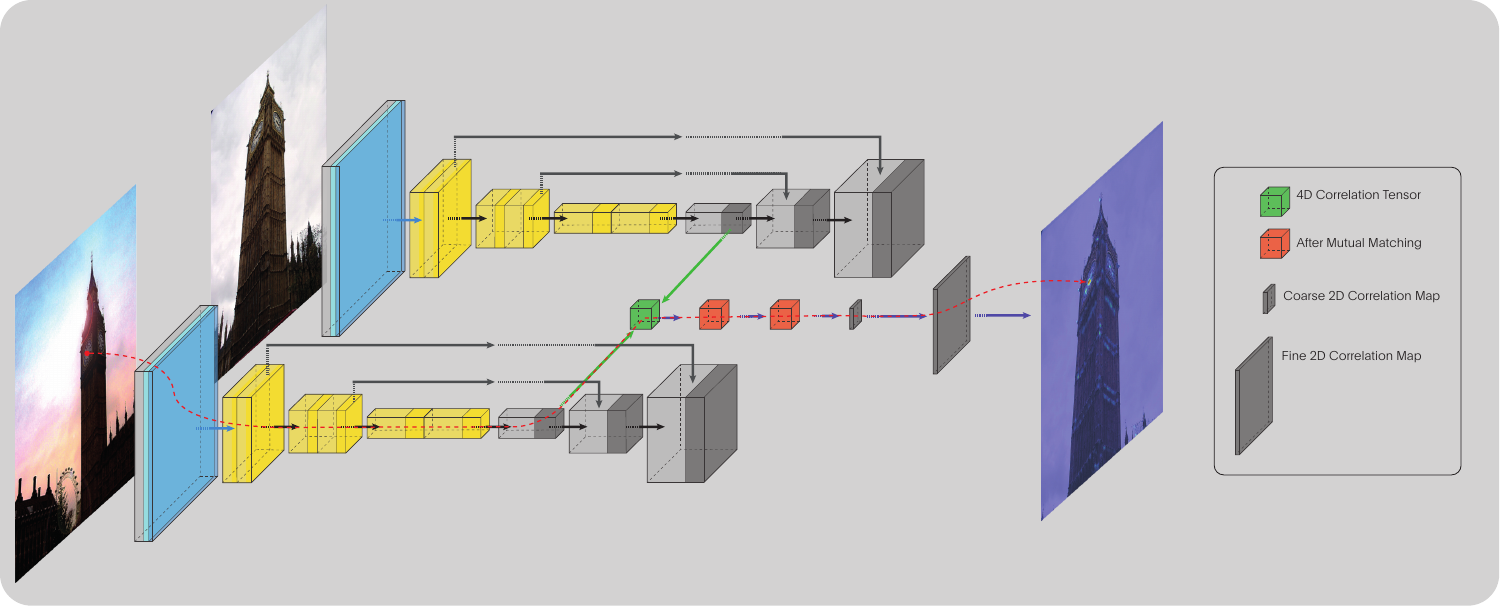}
		\caption{The architecture of $\mathbb{X}$RCNet. The deep correspondence neural network takes a Siamese-like structure. 
		Each branch is composed of a feature encoder (gray and yellow from the left) and FPN-like decoder (the gray blocks to the 
		right). From the coarse layer of FPN, a 4D correlation tensor is calculated (green) and filtered by the mutual matching (MM) 
		layer to get the filtered 4D tensor (orange). After that the tensor is filtered again by the MM layer. Given a query key point 
		from the source image (bottom left), the corresponding features are selected from the FPN coarse layers and query into the 
		4D tensor.} 
		\label{fig:xrcnet}
		\vspace{-14pt}
	\end{figure*}
	
	\section{Related work}
	As the correspondence algorithm is a basic building block in computer vision, there is a huge volume of related works that can 
	be found in the literature. 
	Existing methods range from sparse to dense correspondence estimation.
	Sparse correspondence algorithms typically adopt the three-stage pipeline of detection-description and matching. Each stage 
	has received extensive research focus over the last two decades. 
	For key point detection and description, handcrafted methods SIFT, SURF, 
	BRIEF~\cite{Lowe_IJCV04_SIFT,Bay_ECCV06_SURF,Calonder_ECCV10_BRIEF} and their  
	variants~\cite{Leutenegger_ICCV11_BRISK,MarcPollefeys_CVPR17_EvalDescriptors} were introduced for first detecting, then 
	describing and finally matching a sparse set of key points. Taking into account the local region around each key point, a 
	feature vector of floating points or binary numbers can be extracted to uniquely represent the key points for feature matching 
	or scene description~\cite{Galvez-Lopez_ToR12_BagOfBinary}. Most of the modern 
	descriptors~\cite{CAS_CVPR17_L2Net,Lepetit_CVPR18_LGCNet,Dusmanu_CVPR19_D2Net,Vassilieos_CVPR19_SOSNet,Revaud_NeurIPS19_R2D2,Luo_CVPR20_ASLFeat}
	 focus on data-driven learning approaches, while evaluating the matching performance of descriptors is performed either by 
	measuring the distances between a pair of descriptors or through the ratio test ~\cite{Lowe_IJCV04_SIFT}. Modern matching 
	approaches take into account the constraints between feature descriptors to enhance the matching success 
	rate~\cite{Zhang_ICCV19_OrderAwareCorrespondenceNet,MagicLeap_CVPR20_SuperGlue}. Particularly, 
	SuperGlue~\cite{MagicLeap_CVPR20_SuperGlue} represents the \sota matching success rate against efficiency by exploring 
	the inter/intra-image information. 
	
	Sparse correspondence algorithms achieve efficiency by attending to a small set of salient points in the images, however, for 
	applications such as SfM~\cite{Schonberger_CVPR16_SfSRevisited, Schonberger_ECCV16_PixelwiseViewSelection}, style 
	transfer~\cite{VGG_NeurIPS20_D2D} or view synthesis \cite{Microsoft_CVPR19_InvertingSfM} where per-pixel correspondence 
	maps are often required, simply scaling up the sparse approach becomes prohibitively expensive. In contrast, dense 
	correspondence approaches focus on bridging this gap. One of the earliest dense methods~\cite{Liu_PAMI11_SiftFlow} uses 
	dense feature descriptors and regularising within the local region to achieve a consistent dense flow field. In recent years, 
	deep semantic correspondence 
	networks~\cite{han2017scnet,INRIA_CVPR19_SFNet,Rocco_NIPS18_NCNet,Min_ICCV19_HPF,ShanghaiTech_ICCV19_DCCNet,Oxford_CVPR20_ANCNet}
	 have demonstrated the potential of densely associating key points between a pair of input images. However, as these 
	approaches focus on matching high level regions, they either require a large number of feature channels, typically larger than 
	1024~\cite{INRIA_CVPR19_SFNet,Min_ICCV19_HPF,Ponce_ECCV20_LearnHypercolumn}, or build on top of the 4D correlation 
	tensor and expensive 4D filtering~\cite{Rocco_NIPS18_NCNet,ShanghaiTech_ICCV19_DCCNet,Oxford_CVPR20_ANCNet}. This 
	fact makes it very difficult to scale up to higher image resolutions, which is critical for accurate data 
	association~\cite{Rocco_NIPS18_NCNet,Rocco_ECCV20_SparseNC,Xinghui_NeurIPS20_DRC}. SparseNC overcomes the 
	scalability problem by projecting the memory consuming 4D correlation tensor into a sub-manifold and uses the Minkowski 
	convolution~\cite{Stanford_CVPR19_MinkowskiConv} to approximate the 4D filtering, however the approximation reduces the 
	performance of the network. DualRC ~\cite{Xinghui_NeurIPS20_DRC} keeps the original 4D correlation tensor in its original 
	space, but relies on a coarse to fine re-weighting mechanism to guide the search in a fine resolution correlation map for the 
	best match. In this work, we further reduce the network redundancy by limiting the operation of the 4D correlation tensor. 
	Combined with a much shallower feature backbone, our proposed approach can process images with higher resolution than all 
	previous dense networks on the same hardware setup.
	
	Establishing dense correspondences is also relevant to stereo networks~\cite{Niantic_ECCV20_StereoFromSingle}, deep visual 
	odometry~\cite{TUMCremers_CVPR20_D3VO} and optical flow networks ~\cite{Niantic_ECCV20_StereoFromSingle} since these 
	algorithms involve calculating a dense flow field that associates two individual images. However, a stereo architecture typically 
	assumes that the input views are rectified and the images are captured under roughly the same illumination environment, while 
	the viewpoints are relatively close to each other, which can be viewed as a simplified version of the correspondence problem. 
	Similarly, both the tasks of optical flow estimation and visual odometry are considered to be much more constrained than the 
	general correspondence problem, since both assume that the viewpoints of the input images are close both temporally and 
	spatially in terms of the 6D manifold of the camera poses.   
	\section{Methodology}
	
	In this work we present a new dense correspondence methodology  working with input images of higher resolution than any 
	other \sota dense method and attaining higher accuracy particularly for small error bands. In this section, we first describe the 
	DCN framework illustrated in Fig.~\ref{fig:xrcnet}, then the redundant module is ablated to form the $\mathbb X$RCNet. 
	
	Given a pair of images $\mathbf I$ and $\mathbf I'$, we want to estimate a per-pixel correspondence map that associates a 2D 
	key point from the source image $(x,y) \in \mathbf I$ to a point in the target image $(x',y')\in\mathbf I'$. To reliably associate 
	the point, we first adopt a standard multi-level feature backbone $\mathbf F =f(\mathbf I; \theta_0)$, where $\theta_0$ are 
	learnable variables. Particularly, $\mathbf F=\{\mathbf F_f,\mathbf F_c\}$ where $\mathbf F_f\subset \mathbb 
	R^{C\times\Omega_f}$ is one layer of the feature map within a 2D domain $\Omega_f$ and $\mathbf F_c\subset \mathbb 
	R^{C\times\Omega_c}$ is another layer of the feature map within $\Omega_c$. Subscripts $f$ and $c$ represent the fine and 
	coarse resolutions, while $C$ stands for the number of feature channels. Previous 
	works~\cite{Rocco_NIPS18_NCNet,Rocco_ECCV20_SparseNC,VGG_NeurIPS20_D2D,Xinghui_NeurIPS20_DRC} make use of a 4D 
	correlation tensor $\mathbf C\subset\mathbb R^{\Omega\times\Omega'}$, where $\mathbf C_{(x,y,x',y')} = \mathbf F(x,y)^\top 
	\mathbf F'(x',y')$. Note that all values in the feature maps are positive due to the ReLU activation layer in the feature backbone 
	$f(\cdot)$ immediately before calculating the  correlation. The features are typically normalised along the channels --- 
	$\|\mathbf F(x,y)\|_2 \triangleq 1$ and thus the dot product of two feature vectors is within the range $[0,1]$. The 4D 
	correlation tensor represents all possible candidate matching pairs from the source  to the target image.
	\subsection{Neighbourhood consensus}
	Initially introduced in NCNet~\cite{Rocco_NIPS18_NCNet}, a set of 4D convolutions with learnable variables is trained to filter 
	the noise from the raw correlation tensor. The local 4D volume contains all possible matching pairs within the neighbourhood 
	of the source and target image from which a filtering process is employed in order to collect consensus from them. 
	Neighbourhood Consensus (NC) filtering can be formulated as $\mathbf {\hat C} = N(\mathbf C;\theta_1) + N(\mathbf 
	C^\top;\theta_1)^\top$ where $N(\cdot)$ represents the NC filtering consisting of a sequence of 4D convolution layers. 
	$\mathbf C^\top$ is the permutation operation such that $\mathbf C_{(x,y,x',y')}=\mathbf C^\top_{(x',y',x,y)}$ and $\theta_1$ 
	are the learnable parameters in the NC filtering. The first term corresponds to the matching direction from source to the target 
	and the second term from target to the source. Since the matching direction is independent of the filter weights, $\theta_1$ is 
	shared by the two filtering stages. The result $\mathbf{\hat C}$ has the same dimensionality as $\mathbf C$ that contains the 
	filtered correlation scores. 
	
	To improve accuracy soft Mutual Matching (MM) filtering layers can also be applied before and after the NC filtering to 
	dynamically adjust the scale of the correlation tensor:
	\begin{align}
		\mathbf M_{(x,y,x',y')} &= \frac{\mathbf C_{(x,y,x',y')}}{\max_{\forall (x',y')\in\Omega'} \mathbf C_{(x,y,x',y')}+\epsilon} 
		\label{eqn:1}\\
		\mathbf M'_{(x,y,x',y')}&= \frac{\mathbf C_{(x,y,x',y')}}{\max_{\forall (x,y)\in\Omega}\mathbf C_{(x,y,x',y')}+\epsilon}    
		\label{eqn:2}\\
		\mathbf {\hat C}_{(x,y,x',y')} &= \mathbf M_{(x,y,x',y')} \mathbf C_{(x,y,x',y')} \mathbf M'_{(x,y,x',y')}  \label{eqn:3}   
	\end{align}
	where $\epsilon$ is a infinite small value to improve the numerical stability and prevent errors during the degenerating scenario 
	that the maximum correlation in a domain is 0. It can be seen that the MM layer contains no learnable parameters. As shown in 
	equations~\ref{eqn:1} and~\ref{eqn:2} the MM layer first converts the correlation scores into probabilities by normalising using 
	the maximum correlations with respect to the target domain $\Omega'$ and source domain $\Omega$, respectively. The 
	multiplication of $\mathbf M_{(x,y,x',y')}$ and $\mathbf M'_{(x,y,x',y')}$ can be viewed as the joint probability of matching from 
	source to target and from target to source providing the matching along both directions are independent. Ablating MM layer 
	reduces matching accuracy possibly because the MM layer adjusts the scores in the correlation tensor. More detailed 
	discussion will be provided in Section~\ref{sec:ablation}.
	
	In the end, given a query point $(x,y)\in \mathbf I$, the best matches can be found at $(\hat x',\hat y')=\arg\max_{\forall 
	(x',y')\in \Omega'}\mathbf{\hat{C}}_{(x,y,x',y')}$. The dense correspondence map can be established by calculating $(\hat x', 
	\hat y')$ for every pixel in the source image. In addition, the maximum correlation scores $\mathbf S\subset \mathbb 
	R^{\Omega}$ represent a good indication of the matching reliability, where $\mathbf{S}(x,y) = \max_{\forall (x',y')\in 
	\Omega'}\mathbf{ \hat{C}}_{(x,y,x',y')}$. A sub-set of top $k$ reliable matches $\mathbb S=\{\mathbf S\}_k$ can be collected 
	accordingly, or alternatively set a threshold to remove unreliable 
	matches~\cite{Rocco_NIPS18_NCNet,Rocco_ECCV20_SparseNC,Oxford_CVPR20_ANCNet}. 
	
	\subsection{Correlation re-weighting}
	The main bottleneck for the aforementioned NC filtering and MM layer lies in the fact that the 4D correlation is very expensive 
	to calculate and difficult to scale up. To deal with the problem, SparseNC~\cite{Rocco_ECCV20_SparseNC} projects the 
	correlation tensor onto a sub-manifold that contains the top $k$ highest correlation scores for each source or target pixels. 
	The 4D filtering is then approximated using the Minkowski operation~\cite{Stanford_CVPR19_MinkowskiConv}. In this way, the 
	memory footprint can be dramatically reduced. Higher resolution images can fit into the memory leading to improved 
	performance. However, such an approximation affects the accuracy as shown in ~\cite{Xinghui_NeurIPS20_DRC}. Li et al 
	~\cite{Xinghui_NeurIPS20_DRC} propose to use a hierarchical architecture where the coarse resolution feature map $\mathbf 
	F_c$ is used to calculate the 4D tensor for NC filtering and MM filtering. Then, the 2D correlation map $\mathbf 
	C^c(x,y)\subset \mathbb R^{\Omega'_c}$ at location $(x,y)\in \mathbb R^{\Omega_c}$ is used to guide the searching for the 
	best matches in the fine feature map by re-weighting the correlation map at the fine resolution $\mathbf C^f(x,y)$. 
	Specifically, $\mathbf{\hat{C}}^{f}(x,y) = U(\mathbf C^c(x,y))\cdot \mathbf C^f(x,y)$, where $U()$ is a de-parameterised 
	up-sampling function, $\cdot$ represents the element-wise multiplication, and  $\mathbf{\hat{C}}^f(x,y)$ is the re-weighted 
	correlation map with the fine resolution. More accurate matches can be localised at $(\hat x',\hat y')_f=\arg\max_{\forall 
	(x',y')\in \Omega_f'}\mathbf{\hat{C}}^{f}_{(x,y,x',y')}$. Note that the correlation re-weighting contains no learnable parameters.
	
	\begin{table}
		\begin{center}
			\caption{Ablation study. Training is performed on a single Tesla V100-SXM2 GPU. We use batch size of 16 and Adam 
			optimiser~\cite{Kingma_ICLR15_Adam} for all configurations. We run the training 15 epochs in total. Strong key point 
			supervision is adopted for all methods. The test image up-sample resolution is 1.6K. The Sum of Area under the curve of 
			MMA represent the overall accuracy over multiple error bands. All approaches are trained using ResNet18. 
			}\label{tab:ablation}
			\resizebox{1.0\linewidth}{!}{
				\begin{tabular}{lccccccccc}
					\hline
					\small Component/Method & DualRC & SparseNC & NCNet & $\mathbb X$RC$_1$ & $\mathbb X$RC$_2$ & 
					$\mathbb X$RC$_3$ & $\mathbb X$RC$_4$ \\ 
					\hline
					\small{4D correlation tensor} &\cmark&\cmark&\cmark&\cmark&\cmark&\cmark& \xmark \\
					\small{NC filtering}          &\cmark&\cmark&\cmark&\xmark&\xmark&\xmark& \xmark \\
					\small{Mutual matching}       &\cmark&\xmark&\cmark&\cmark&\cmark&\xmark& \xmark \\
					\small{DualRC re-weighting}   &\cmark&\xmark&\xmark&\cmark&\xmark&\cmark& \cmark \\
					\small{Memory (GB)}           & 6.78 & 3.73 & 5.40 & 4.57 & 4.25 & 4.36 & 4.21 \\
					\small{Training time (s)}     & 2.73 & 0.49 & 0.73 & 0.48 & 0.26 & 0.35 & 0.27 \\
					\small{Sum of Area}           & 3.90 & 3.20 & 3.61 & 3.65 & 2.49 & 3.36 & 3.26 \\
					\hline
				\end{tabular}\label{tab:ablation}
			}
		\end{center}
		\vspace{-14pt}
	\end{table}
	
	\begin{figure}[]
		\centering
		\includegraphics[width=.8\linewidth]{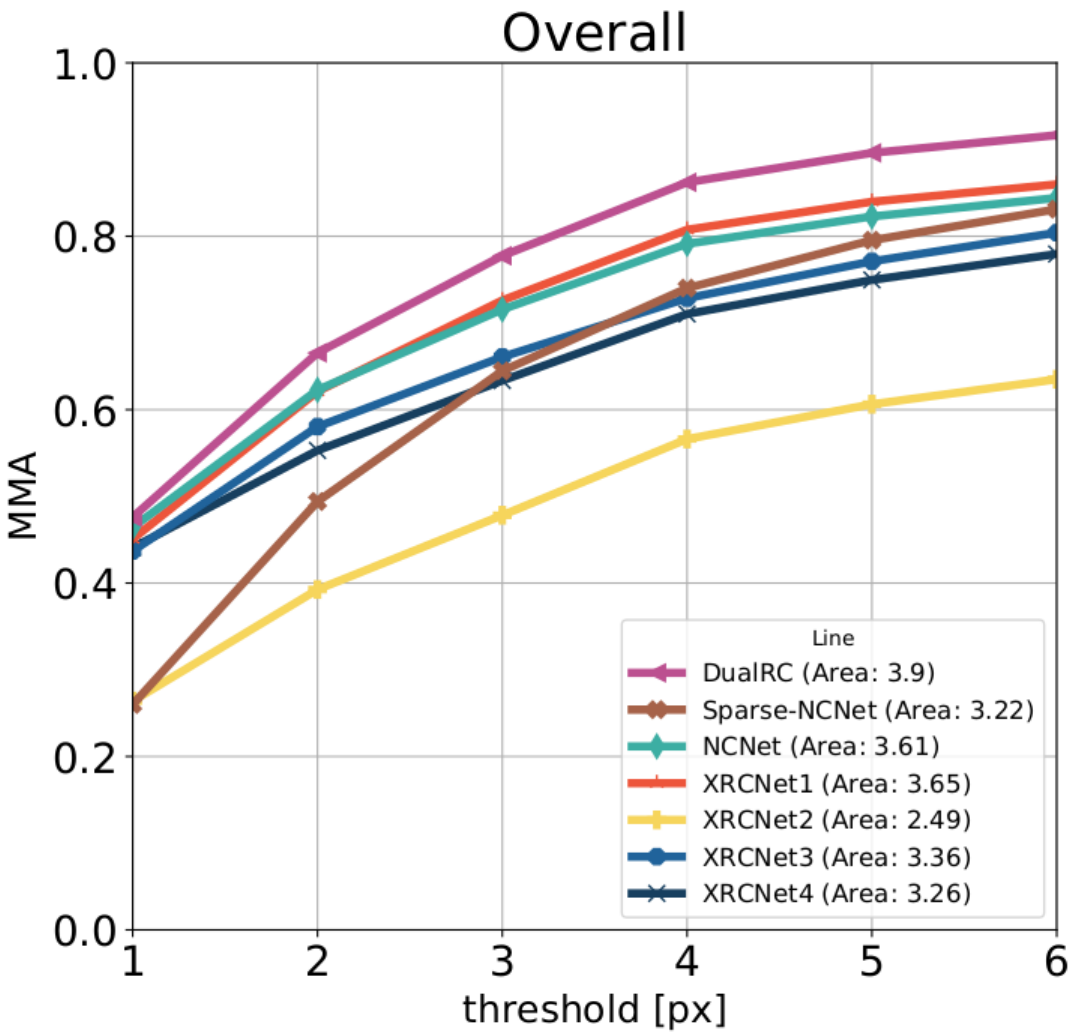}
		\caption{Quantitative evaluation of Neighbourhood Consensus architectures on the HPatches dataset.}
		\label{fig:hpatches_ablation}
		\vspace{-14pt}
	\end{figure}
	
	\subsection{Ablation study and $\mathbb{X}$RCNet} \label{sec:ablation}
	To better understand the pros and cons of the mainstream DCN architectures, we conducted an ablation study over several 
	\sota methods, namely, NCNet~\cite{Rocco_NIPS18_NCNet}, SparseNC~\cite{Rocco_ECCV20_SparseNC}, and 
	DualRC~\cite{Xinghui_NeurIPS20_DRC}. The left column in Tab.~\ref{tab:ablation} lists the key modules shared by the DCNs. 
	We tested the performance of all possible combinations of the key modules using the same training protocol on the 
	MegaDepth dataset~\cite{MegaDepthLi18} following the work of~\cite{Dusmanu_CVPR19_D2Net} - all baselines are trained with 
	strong supervision with ResNet18 with 256 channels and hard relocalization~\cite{Rocco_ECCV20_SparseNC} for a fair 
	comparison. The evaluation over feature backbones is discussed in Fig.~\ref{fig:backbone}. For each configuration, we record 
	the average memory consumption, training speed, and the overall matching accuracy. The accuracy measurements are the 
	sum of the area below the Mean Matching Accuracy (MMA) curve on the HPatches dataset, comprised of challenging scenes 
	with illumination and viewpoint variation. In addition to Tab.~\ref{tab:ablation}, we also plot the accuracy of MMA in 
	Fig.~\ref{fig:hpatches_ablation} for completeness.
	
	From the experiments, we observe that although all the modules of DCN contribute to the accuracy, they come with a variety 
	of costs. Particularly, 1) the 4D NC filter consumes nearly 50\% more memory and is more than 5 times slower comparing 
	DualRC and $\mathbb X$RC$_1$. SparseNC reduces the expense of NC filter using the sparse 4D correlation with Minkowski 
	convolution~\cite{Stanford_CVPR19_MinkowskiConv} but at the cost of degrading accuracy.
	2) The DualRC re-weighting often plays an important role to the accuracy comparing DualRC with NCNet and 
	$\mathbb{X}$RC$_1$ with $\mathbb{X}$RC$_2$. 3) Mutual matching layer contributes relative less to accuracy but is also 
	cheap to calculate comparing $\mathbb{X}$RC$_1$ with $\mathbb{X}$RC$_3$ and therefore we do not remove it. 4) Removing 
	both NC filtering and DualRC re-weight dramatically increases the speed but also decreases the accuracy for 
	$\mathbb{X}$RC$_2$ significantly. 5) Removing the 4D correlation tensor, similar to UCN~\cite{Choy_NIPS16_UCN}, hurts the 
	performance for $\mathbb{X}$RC$_4$ compared to $\mathbb{X}$RC$_3$. 
	To summarise, we select $\mathbb X$RC$_1$ as the default architecture of $\mathbb X$RCNet, which is about 5 times faster 
	than DualRC but nearly 50\% smaller in terms of memory costs. Also, it allows us to adopt 4 times larger batch size during 
	training and can run up to 40 epochs in about same amount of time of training DualRC for 15 epochs. Fig.~\ref{fig:teaser} 
	(left) shows qualitative examples of removing the NC module and the proposed alternative. 
	
	
	The prediction of $\mathbb X$RCNet is 2D correlation maps as illustrated in Fig.~\ref{fig:xrcnet}. The loss can be then 
	calculated using the F-norm between the prediction and the ground truth 
	distribution~\cite{Oxford_CVPR20_ANCNet,Xinghui_NeurIPS20_DRC,VGG_NeurIPS20_D2D,Lepetit_ECCV20s_S2DNet}. 
	Particularly, we get the ground truth distribution by converting a 2D key point coordinate into a Probability Density Function 
	(PDF). Specifically, we assign the probability of 4 pixels that are the nearest neighbours of the ground truth key point 
	according to their normalised 2D distance. Then the PDF is further filtered by a 3$\times$3 Gaussian 
	kernel~\cite{Oxford_CVPR20_ANCNet}. To keep the ablation test fair, all the networks in Tab.~\ref{tab:ablation} are supervised 
	using the same loss term.
	
	\textbf{$\mathbb X$RCNet Inference} We distribute various key modules illustrated in Fig.~\ref{fig:xrcnet} over multiple GPUs 
	during inference to allow images with much larger resolution to be processed efficiently. Together with the low-cost but 
	relatively accurate $\mathbb X$RC$_1$, we can address the critical question of how the input image resolution affects the 
	matching accuracy of a DCN. To this end, we evaluate $\mathbb X$RC$_1$ using various image resolutions ranging from 1280 
	to 4K. More details are provided in section~\ref{sec:optimal}. The source code for both training and evaluation is attached in 
	the supplementary materials. 
	
	\section{Experiments}
	In this section, we describe the experiments we conducted to evaluate the performance of $\mathbb X$RCNet, the training 
	strategy and we discuss the question of the relationship between the input image resolution and matching accuracy. 
	
\begin{figure}[t!]
		\centering
		\includegraphics[width=0.85\linewidth]{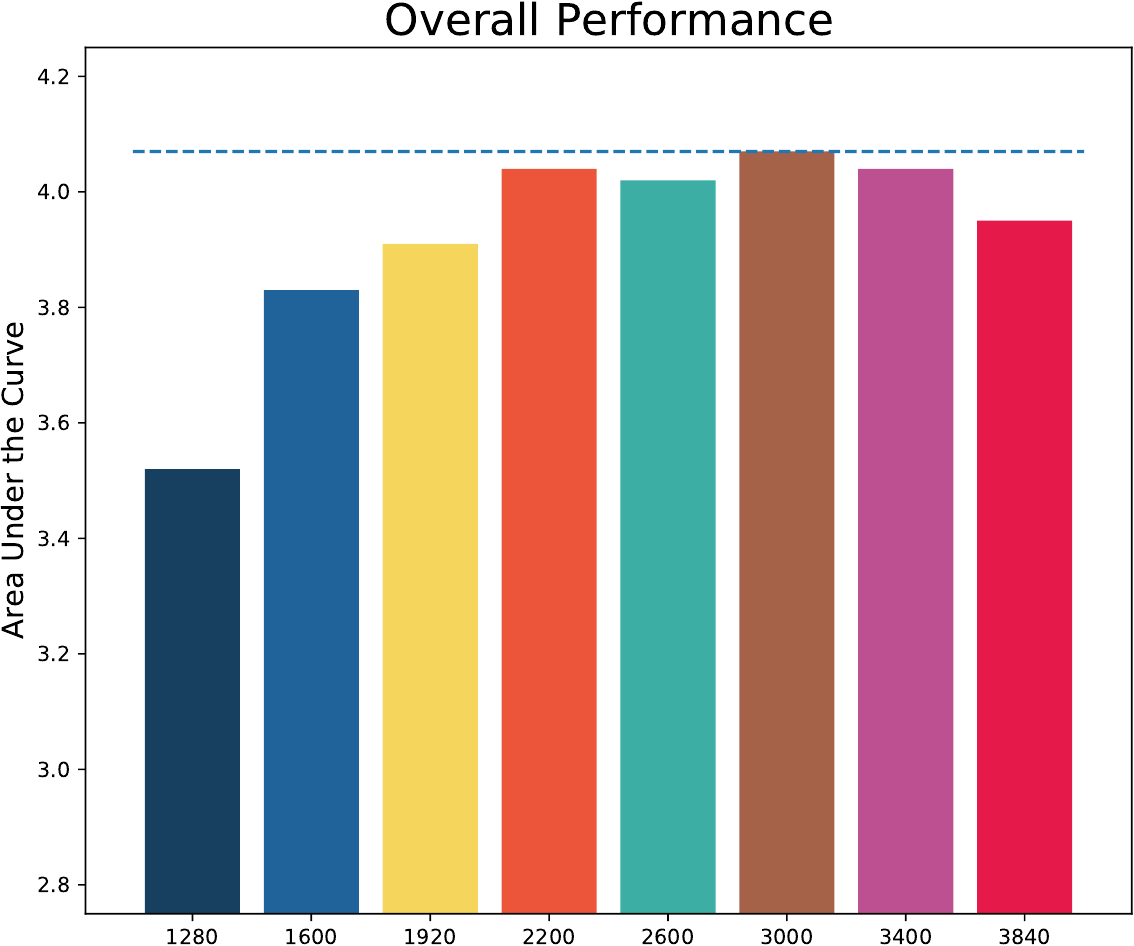}
		\caption{Evaluation of the optimal up-sampling resolution. The height of each bar is the Sum of Area under the curve of the 
		MMA, which represent the overall accuracy. It can be seen that both the low and high up-sampling resolution have a 
		negative impact over the $\mathbb X$RCNet's accuracy.}
		\label{fig:bar_chart}
		\vspace{-14pt}
	\end{figure}

	\textbf{Implementation details}  The $\mathbb X$RCNet training and evaluation code is implemented using 
	PyTorch~\cite{NEURIPS2019_Pytorch}. For the feature backbone, we mainly evaluated the 
	ResNet101/50/18~\cite{He_CVPR16_ResNet}, HRNet64/32/18~\cite{Microsoft_PAMI20_HRNet,Microsoft_CVPR20_HigherNet} 
	and the FPN256/128~\cite{Lin_CVPR17_FPN}. The ResNet and HRNet are pre-trained on 
	ImageNet~\cite{Feifei_CVPR09_ImageNet} and kept fixed during all training procedures. The parameters of the FPN layers are 
	trained from scratch. The configuration of ResNet101 is adopted from~\cite{Rocco_NIPS18_NCNet}, the ResNet18 is truncated 
	after the 3rd layer, the coarse feature map is extracted from the 3rd layer in the ResNet18 and the coarse layer is taken from 
	the output of layer 1. The FPN architecture is identical to the original work of FPN~\cite{Lin_CVPR17_FPN} except that a ReLU 
	layer is inserted before the feature normalisation. For HRNet we tested 18, 32, and 64 channels configurations. We truncated 
	HRNet after the third stage in order to keep the input image ratio identical to ResNet. Here we considered both including and 
	excluding the fusing (transition) stages. In addition, we use the output of the first branch as the fine feature map and the 
	output of the third branch as the coarse feature map in order to be consistent to the fine to coarse ratio we used for ResNet. 
	We train our model using the Adam optimiser with an initial learning rate of 0.01 and momentum 0.9. The batch size is 64. The 
	learning rate is halved for every 5 epoch until 15 and remain constant till the 40th epoch. The model with lowest validation 
	error is adopted for the final evaluation. It is worth pointing out that comparing with the training in Tab.~\ref{tab:ablation} 
	which only runs 15 epochs and uses batch size of 16 as previous work, the training with more epochs and a larger batch size 
	result in a much higher accuracy which can be seen by comparing Fig.~\ref{fig:bar_chart} and the bottom row of 
	Tab.~\ref{tab:ablation}. 
	
	\textbf{Training data} We adopt the same training protocol as D2Net~\cite{Dusmanu_CVPR19_D2Net} on the MegaDepth 
	dataset~\cite{MegaDepthLi18}. MegaDepth includes 196 scenes and the corresponding 3D point clouds created using 
	SfM~\cite{Schonberger_CVPR16_SfSRevisited,Schonberger_ECCV16_PixelwiseViewSelection}. The camera internal and external 
	parameters are also jointly estimated and provided by the dataset. We follow the same methodology 
	of~\cite{Dusmanu_CVPR19_D2Net} to extract a sparse set of ground truth correspondent points. Only image pairs with more 
	than 50\% of overlapping field of view are selected as training pairs (15,070). The validation image pairs (14,638) are selected 
	from scenes containing more than 500 good pairs. All training pairs are randomly shuffled to avoid over-fitting to specific 
	scenes.
	
	\subsection{Correspondence Evaluation}
	
	\begin{table}[]
		\caption{The size statistics of the image size in the 3 testing datasets. The minimum, mean, and maximum size over height 
		and width recorded. HPatches has the lowest mean image resolution and InLoc has the highest resolution.}
		\begin{tabular}{|c|c|c|c|c|c|c|}
			\hline
			{\begin{tabular}[c]{@{}c@{}} \end{tabular}} & \multicolumn{2}{c|}{HPatches} & \multicolumn{2}{c|}{InLoc} & 
			\multicolumn{2}{c|}{\begin{tabular}[c]{@{}c@{}}Aachen \\ Day-Night\end{tabular}} \\ \cline{1-7} 
			& h & w & h & w & h & w \\ 
			min & 380 & 512 & 1200 & 1200 & 1063 & 1063 \\
			mean & 780 & 980 & 2397 & 2531 & 1268 & 1498 \\
			max & 1411 & 1536 & 4032 & 4032 & 1600 & 1600 \\
			\hline
		\end{tabular}
		\label{tab:dataset}
		\vspace{-10pt}
	\end{table}
	
\begin{figure*}[t]
		\centering
		\includegraphics[width=0.9\textwidth]{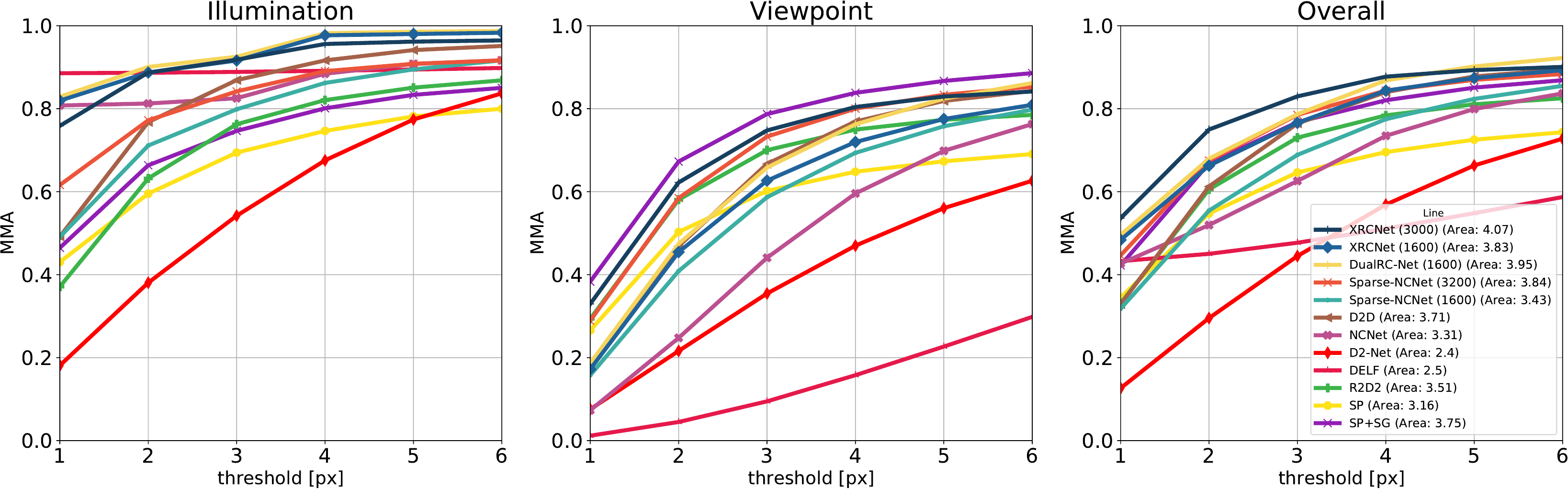}
		\caption{The comparison of $\mathbb{X}$RCNet with the \sota correspondence networks on the HPatches dateset. It can 
		be seen that $\mathbb X$RCNet when using the up-sampled testing images of 3000 surpassed all other methods, i.e. Deep 
		Local Feature Network (DELF)~\cite{Noh_ICCV17_DELF}, DualRC~\cite{Xinghui_NeurIPS20_DRC}, 
		Sparse-NCNet~\cite{Rocco_ECCV20_SparseNC}, SP~\cite{MagicLeap_CVPR18_SuperPoint}, 
		SG~\cite{MagicLeap_CVPR20_SuperGlue}.}
		\label{fig:hpatches_performance}
	\end{figure*}

	\textbf{HPatches} is widely used for evaluating sparse feature matching and dense correspondence 
	algorithms~\cite{Vassilieios_CVPR17_HPatches}. It contains two main challenges --- the viewpoint and illumination variations 
	consisting of 56 and 52 sequences of testing images respectively. Each sequence contains 6 images and the first image is 
	matched against the remaining ones. The native image size is reported in Tab~\ref{tab:dataset}. Testing images contain both 
	indoor and outdoor scenes. The ground truth homography is provided so that the correspondences can be densely evaluated. 
	The evaluation procedure is adopted 
	from~\cite{Dusmanu_CVPR19_D2Net,Revaud_NeurIPS19_R2D2,Rocco_ECCV20_SparseNC,Xinghui_NeurIPS20_DRC} to allow 
	direct comparison with these baseline methods. The evaluation metric used is the Mean Matching Accuracy (MMA) that 
	estimates the average number of correct matches over the total number of matches using top 2000 proposed matches by the 
	testing neural networks, where the correct matches are defined as the distance between the predicted 2D key points to 
	ground truth. $\mathbb X$RCNet sets a new accuracy standard as it can be seen at the comparative evaluation graph shown 
	in Fig.~\ref{fig:hpatches_performance}.
	
	\begin{table*}[]
		\centering
		\caption{Evaluation on the Aachen dataset.} 
		\begin{tabular}{ccccccccc}
			\hline
			Error Band    & \multicolumn{1}{l}{ASLFeat+OANet} & 
			\multicolumn{1}{l}{D2-Net} & \multicolumn{1}{l}{SparseNC} & \multicolumn{1}{l}{R2D2} & \multicolumn{1}{l}{DualRC-Net} & 
			\multicolumn{1}{l}{SP + SG} & \multicolumn{1}{l}{} & \multicolumn{1}{l}{$\mathbb X$RCNet-1.6k} \\
			\hline
			0.25m \& 2$^{\circ}$ & 77.6 & 74.5 & 76.5 & 76.5 & 79.6 & 79.6 & & 76.5 \\
			0.5m \& 5$^{\circ}$  & 89.8 & 86.7 & 84.7 & 90.8 & 88.8 & 90.8 & & 85.7 \\
			5m \& 10$^{\circ}$  & 100.0 & 100.0 & 98.0 & 100.0 & 100.0 & 100.0 & & 100.0 \\
			\hline                
		\end{tabular}
		\vspace{-15pt}
		\label{table:aachen}
	\end{table*}
	
	\textbf{Aachen Day-Night} dataset~\cite{Sattler_2018_CVPR,Sattler2012BMVC} is a challenging outdoor relocalisation dataset. 
	The Day-Night relocalisation challenge contains 98 night query images to be relocalised with respect to 20 day-time candidate 
	images. The performance of $\mathbb X$RCNet compared to the baselines on the night query images is shown in 
	Tab.~\ref{table:aachen}, while an example qualitative comparison and the produced 2D heatmap in the reference image are 
	shown in Fig.~\ref{fig:nc_nonc_comparison} (right). We provide 3D reconstruction results of our $\mathbb{X}$RCNet and 
	DualRC in the supplementary material. $\mathbb{X}$RCNet achieves commensurate performance to the \sota, while also 
	having a much smaller memory footprint for the used input resolution size and faster inference speed (see 
	Tab.~\ref{tab:ablation}).
	
	\begin{table*}[t]
		\centering
		\caption{Evaluation on the InLoc dataset. Best result is shown in \textbf{bold} and second best is \underline{underlined}.}
		\resizebox{1.0\textwidth}{!}{
			\begin{tabular}{ccccccccccc}
				\hline
				Error Band    & \multicolumn{1}{l}{DualRC} & 
				\multicolumn{1}{l}{SparseNC} & \multicolumn{1}{l}{NCNet} & \multicolumn{1}{l}{InLoc} & \multicolumn{1}{l}{DensePE} & 
				\multicolumn{1}{l}{D2-Net} & \multicolumn{1}{l}{R2D2} & \multicolumn{1}{l}{$\mathbb X$RCNet-1.6k} & 
				\multicolumn{1}{l}{$\mathbb X$RCNet-3k} & \multicolumn{1}{l}{$\mathbb X$RCNet-4k} \\
				\hline
				0.25m \& 10$^{\circ}$ & 44.1 & 45.6 & 44.1 & 38.9 & 35.3 & 43.2 & \underline{47.3} & 44.7 & 46.2 & \textbf{50.2} \\
				0.5m \& 10$^{\circ}$ & 67.5 & 66.3 & 63.8 & 56.5 & 47.4 & 61.1 & 67.2 & 66.6 & \underline{67.8} & \textbf{68.7} \\\
				1m \& 10$^{\circ}$ & \textbf{82.4} & 79.9 & 76.0 & 69.9 & 57.1 & 74.2 & 73.3 & 79.6 & \textbf{82.4} & \underline{81.2} 
				\\
				\hline
			\end{tabular}
		}
		\label{table:inloc}
		\vspace{-10pt}
	\end{table*}
	
	\begin{figure}[]
		\centering
		\includegraphics[width=.75\linewidth]{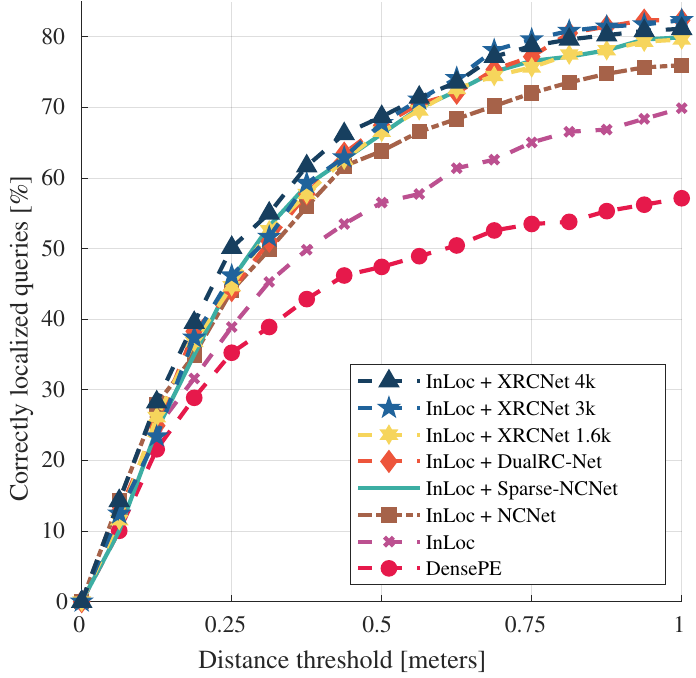}
		\caption{Pose accuracy of InLoc measured by the percentage of correctly localised queries over different distances.}
		\label{fig:inloc_performance}
		\vspace{-14pt}
	\end{figure}
	
	\textbf{InLoc} mainly contains indoor images captured with a different type of sensors~\cite{taira2018inloc}. It is a popular 
	benchmark for evaluating the accuracy of camera localisation with respect to large variety of indoor scenes. Reference images 
	are obtained with a 3D scanner and the query images are captured using a mobile phone several months later to introduce 
	extra non-static challenges. InLoc contains significant viewpoint changes and illumination variation. We adopt the evaluation 
	procedure of~\cite{taira2018inloc} to find the top 10 candidate database images for each query image. $\mathbb{X}$RCNet is 
	used to calculate the matches between them, and the final 6D camera pose is estimated using P$n$P~\cite{Gao_PAMI03_P3P} 
	and dense pose verification~\cite{taira2018inloc}. The results are provide in Fig.~\ref{fig:inloc_performance} and 
	Tab.~\ref{table:inloc} which show $\mathbb{X}$RCNet significantly outperform the others.  
	
	\begin{figure*}[t!]
		\centering
		\includegraphics[width=.95\textwidth]{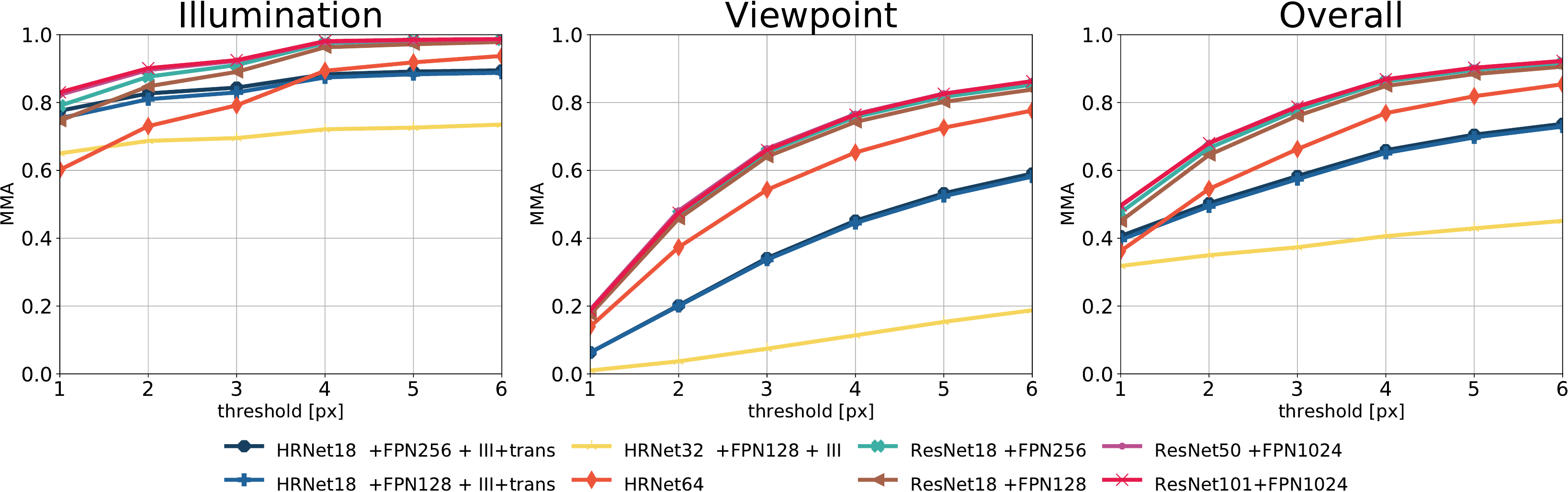}
		\caption{Comparison of different backbone architectures considered for dense correspondence matching.}
		\label{fig:backbone}
		\vspace{-14pt}
	\end{figure*}
	
	\subsection{Optimal Resolution $\mathbb{X}$}\label{sec:optimal}
	A common practice to achieve better accuracy in previous works, is to up-sample the original images to a higher resolution 
	that almost consistently improve the final matching accuracy 
	~\cite{Rocco_NIPS18_NCNet,Rocco_ECCV20_SparseNC,Xinghui_NeurIPS20_DRC} as long as the network can fit into the GPU 
	memory. However, up-sampling the input image to infinity causes issues because the information contained in the original 
	image is fixed. Increasing the image size implies the receptive field of a deep neural network will reduce and so will the 
	descriptiveness of the feature maps. Therefore, there must be an optimal resolution $\mathbb X$ for a network to achieve its 
	best performance for a given input. Thanks to the light-weight design of $\mathbb X$RCNet and the multi-GPU inference, we 
	ran a series of experiments to confirm the existence of the optimal $\mathbb X$ given a pre-trained $\mathbb X$RCNet by 
	varying the up-sampling rate of the testing images.
	
\begin{figure}[]
		\centering
		\includegraphics[width=1.0\linewidth]{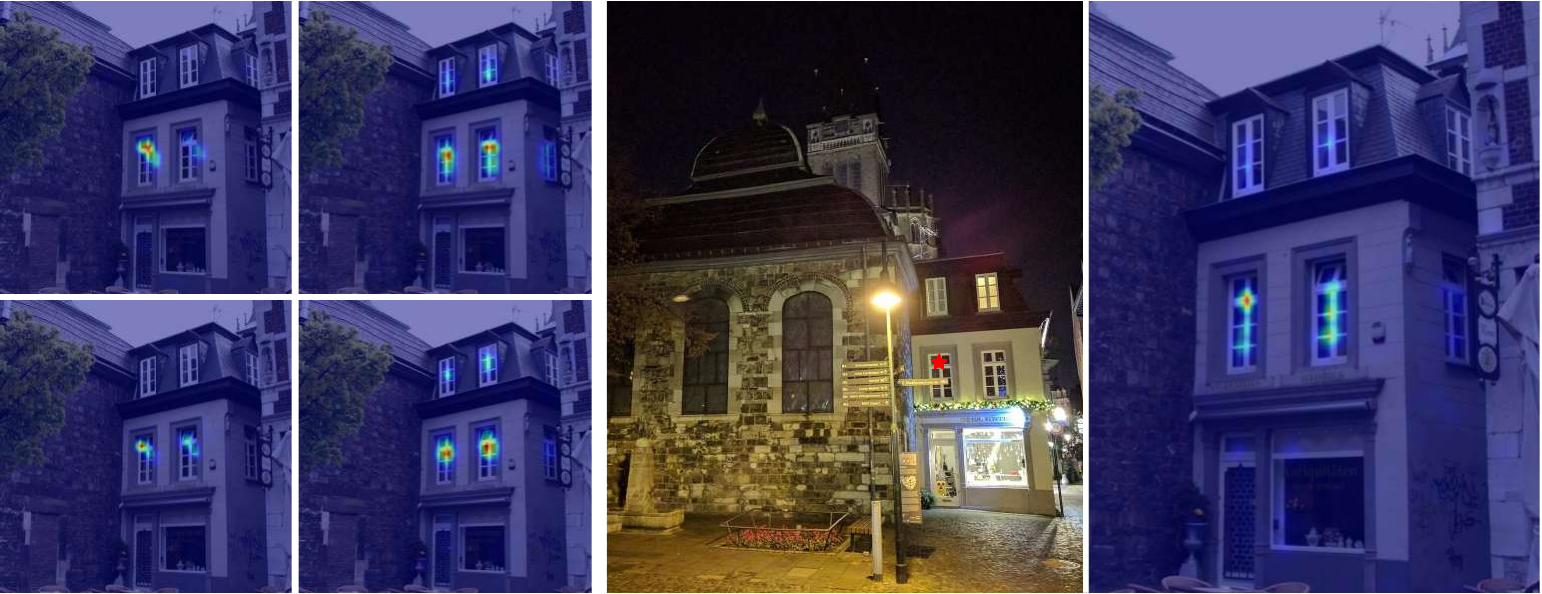}
		\caption{\textbf{Left quadrant}: top row represents a heatmap from ResNet101, while the bottom row is from a light-weight 
		ResNet18. The left column was produced from a model with NC filtering, while the right column has no NC filtering. 
			Removing the NC reduces the capability of differentiating repetitive patterns such as the windows and therefore affects 
			the matching accuracy. \textbf{Right}: the query night image with the chosen keypoint location marked with a red star 
			and the heatmap produced from a ResNet18 model without NC filtering but using a high resolution image. It can be seen 
			that the increased resolution balanced the issue of being sensitive to repetitive patterns without NC.}
		\label{fig:nc_nonc_comparison}
		\vspace{-10pt}
	\end{figure}

	Particularly, we resize the image of HPatches from 1280 pixels up to 4K (3840$\times$2160) with fixed aspect ratio and 
	evaluate. Fig.~\ref{fig:bar_chart} shows the total area under the accuracy curve of MMA. We discover that the best matching 
	accuracy increases with respect to the image resolution and gradually saturates around the resolution 3k. 
	Using an image resolution higher than 3k reduces the matching accuracy. The accuracy at $\mathbb{X}$RCNet-3k surpassed 
	the \sota DCN performance in the same error band on both HPatches (Fig.~\ref{fig:hpatches_performance}). For InLoc, 
	$\mathbb{X}$RCNet-4k further surpassed $\mathbb{X}$RCNet-3k (See Tab.~\ref{table:inloc} and 
	Fig.~\ref{fig:inloc_performance}) possibly because of the relatively larger mean native resolution shown in 
	Tab~\ref{tab:dataset}. 
	Unfortunately, the $\mathbb{X}$RCNet-1.6k gives the best performance on Achen Day-Night which is inconsistent. However, 
	we observe individual cases illustrated in Fig.~\ref{fig:nc_nonc_comparison}, up-sampling remains effective as the heatmap of 
	the $\mathbb{X}$RCNet-3k (right) is less ambiguous in the repetitive regions over $\mathbb{X}$RCNet-1.6k (left). The 
	inconsistent results on Achen Day-Night is possibly due to the a much smaller number of testing pairs compared with other 
	datasets (98 pairs in Achen Day-Night vs 108x5 pairs in HPatches and 329x10 pairs in InLoc).
	
	\subsection{Feature Backbone}
	
	In the end, we show experiments with different feature backbone architectures on HPatches. We have evaluated the matching 
	accuracy using variant of both the ResNet and the HRNet backbones. In Fig.~\ref{fig:backbone}, it can be seen that when 
	using the ResNet18 and ResNet50, the performance of the DualRC is almost identical to the original DualRC with ResNet101. 
	The HRNet is another candidate we consider to replace the original feature backbone for the DualRC. However, HRNet seems 
	less competitive when integrated with the correspondence network. This is possibly because of the small number of channels 
	in the lower layer and reduces the descriptiveness of the feature map for the lower layers. We also tested FPN with 128 and 
	256 channels. Although the 128 channel does not affect the accuracy much, the channel number is a relatively small cost to 
	afford. We adopt 256 as default channel number.
	
	\section{Conclusion}
	In this paper, we propose the $\mathbb X$Resolution Correspondence Network, which is the result of a systematic study of 
	the \sota dense correspondence networks. We noticed that a key component of these networks --- the learned 4D correlation 
	tensor --- does not have a huge impact on the performance. Therefore, removing the 4D filtering with learnable parameters 
	allows $\mathbb X$RCNet to learn quicker and enables it to process input images with resolution up to 4K. The proposed a 
	DCN architecture outperforms \sota on HPatches and InLoc, and enables us to investigate the intriguing question if increasing 
	the input image resolution is always beneficial to matching accuracy. Through extensive experimentation and a thorough 
	ablation study we observe a saturation of the matching performance over the optimal resolution $\mathbb{X}$. We hope this 
	work can shed light on how to design efficient and effective correspondence networks, while acting as a first step towards the 
	interesting problem how the scale differences in input images affect DCNs.
	
	\paragraph{Acknowledgements}
	We would like to thank Umar Ahmed, Guy Newsom and the rest of the XYZ Reality team for helping out with various design 
	concepts and fruitful discussions. In addition, we are grateful to Olivia Wiles for kindly providing the raw data for comparison.
	
	{\small
		\bibliographystyle{ieee_fullname}
		\bibliography{reference}
	}
	
\end{document}


	\title{$\mathbb{X}$Resolution Correspondence Networks \\ -Supplementary Material-}
	
	\maketitle
	\begin{figure}[h]
		\centering
		\includegraphics[width=.85\textwidth]{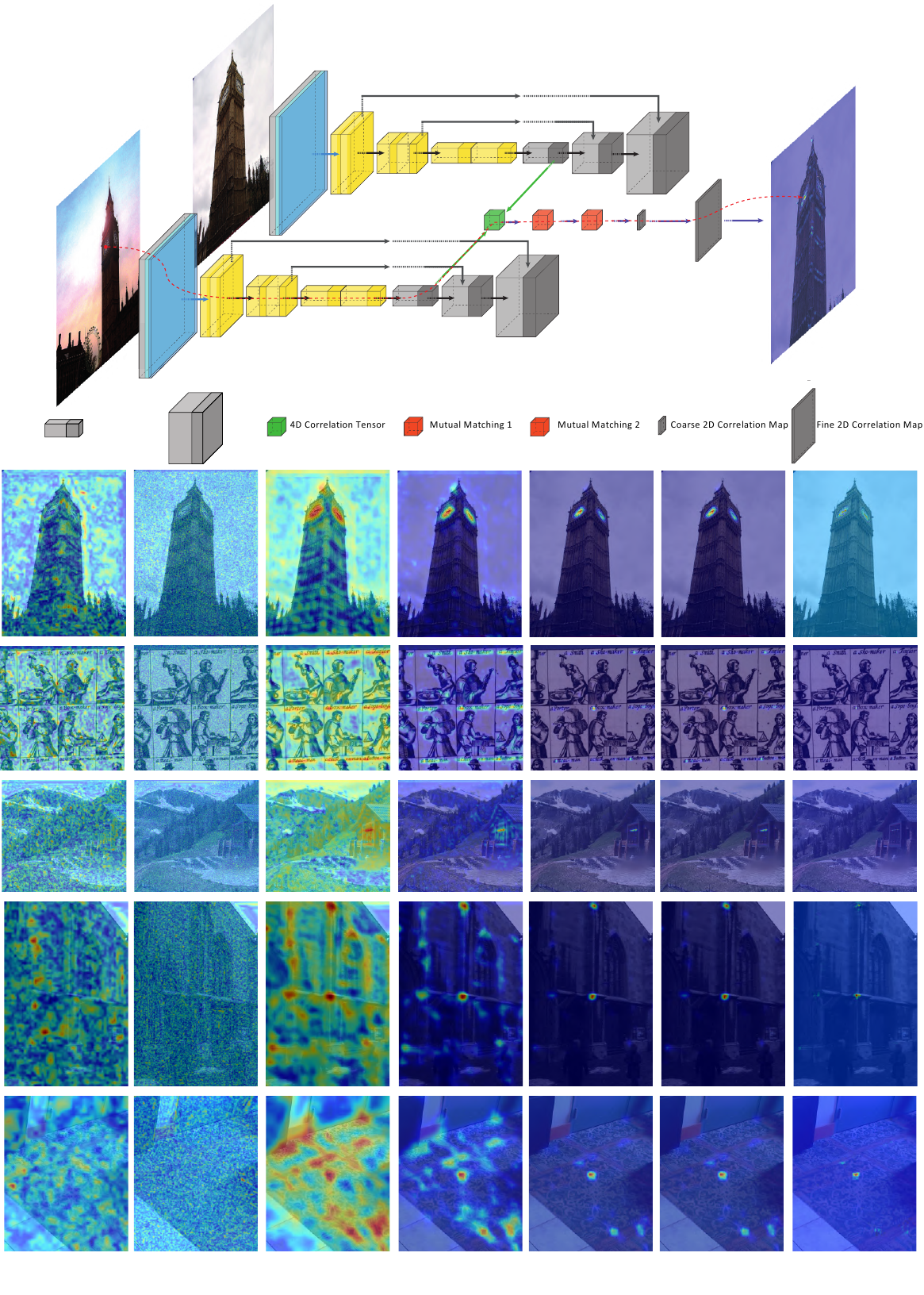}
		\caption{Visualisation of the feature maps and correlation maps of key components in $\mathbb{X}$RCNet.}
		\label{fig:visualisation}
	\end{figure}
	
	This supplementary material provides extra details which are not presented in the main paper due to space limitations. In the 
	following document we discuss the effects of using various re-sampling resolutions during testing in Sec.~\ref{Sec_1}. In 
	Sec.~\ref{Sec_2}, we provide more in depth comparison on HPatches and propose a novel evaluation criterion to demonstrate 
	the effectiveness of the $\mathbb X$RCNet. In Sec.~\ref{Sec_3}, we show more qualitative results on InLoc and Aachen 
	Day-Night dataset, which further demonstrate the quality of the proposed model. Moreover, in Sec.~\ref{Sec_4} we 
	demonstrate the accuracy of our $\mathbb X$RCNet in the challenging task of 3D reconstruction using the Aachen Day-Night 
	dataset. We conclude with a brief description of the source code released in Sec.~\ref{Sec_5}. 
	
	First of all, we visualise the output feature maps and correlation maps of the key modules in $\mathbb X$RCNet in 
	Fig~\ref{fig:visualisation} to illustrate the effectiveness of the key modules when solving a correspondence task. We plot 5 
	examples of various training and testing images. Each is superimposed with the colour map representing the output feature 
	maps or correlation maps of the corresponding modules. From left to right, we plot the coarse features maps from the FPN 
	decoder, the fine feature maps, the 2D coarse correlation map calculated by querying the key point in the source image into 
	the 4D correlation tensor, the same coarse correlation map querying into the 4D tensor after the first mutual matching layer 
	and after the second mutual matching layer respectively. In the end, we plot the final 2D coarse correlation map and the fine 
	correlation map after the re-weighting. For feature maps, we simply visualise the max values alone the channels. It can be 
	noticed that the coarse and fine feature map contains similar patterns except the resolution difference possibly due to the 
	original design of FPN layers. The raw 4D correlation tensor does show a peak around the ground truth point location but also 
	contains significant amount of noise. After two rounds of mutual matching filtering, most of the noise are suppressed except 
	a few ambiguous candidates, and the final re-weighting allows the network to look into the local area in detail so that 
	$\mathbb X$RCNet can make correct predictions in the end. 
	
	\begin{figure}[h]
		\centering
		\includegraphics[width=1.0\textwidth]{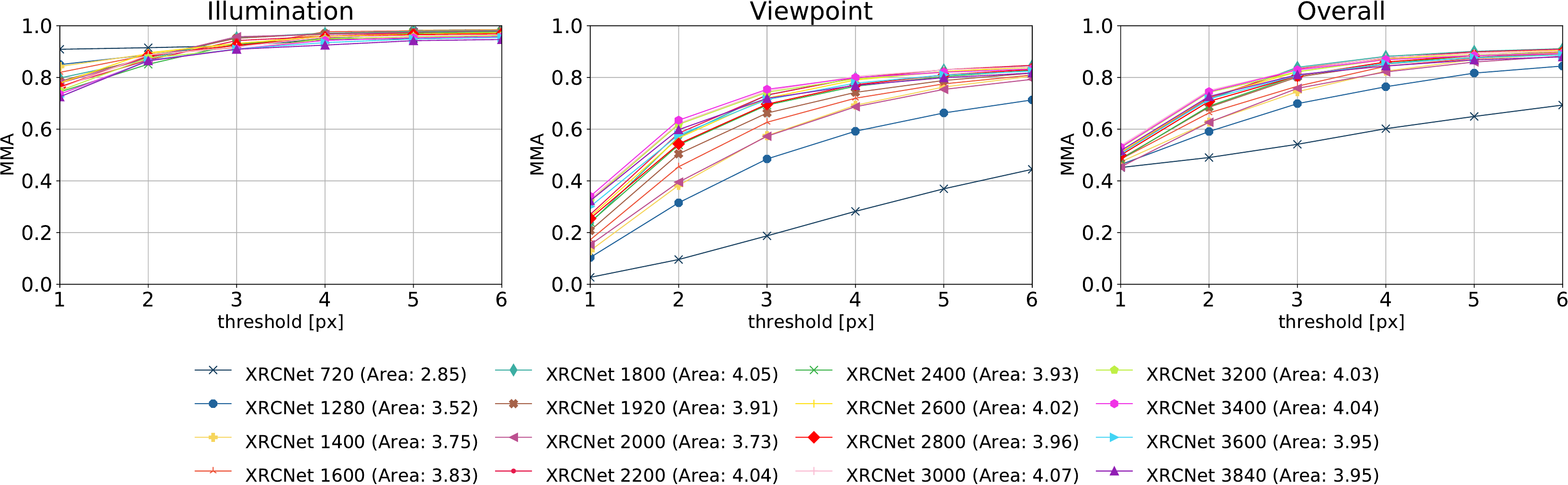}
		\caption{Comparison of $\mathbb X$RCNet with respect to the up-sampled input image resolution evalauted on the 
		HPatches dataset.}
		\label{fig:optimal_resolution}
	\end{figure}
	
	\begin{figure}[h]
		\centering
		\includegraphics[width=0.33\textwidth]{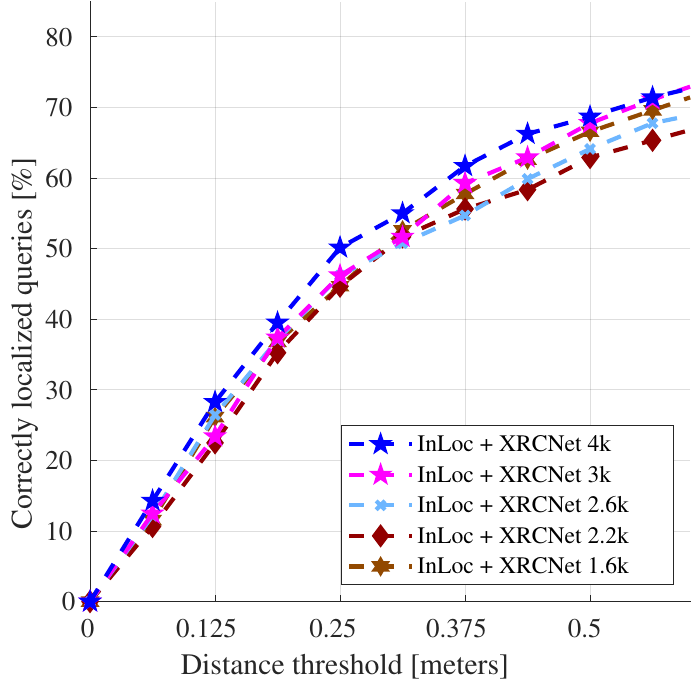}
		\caption{Comparison of $\mathbb X$RCNet with respect to the up-sampled input image resolution on the InLoc dataset 
		after geometric verification.}
		\label{fig:inloc_resolution_comparison}
	\end{figure}
	
	\section{Effects of Different Re-sampling Resolutions}\label{Sec_1}
	In this section we present both qualitative and quantitative analysis on HPatches when re-sampling the testing images into 
	various resolutions. As shown in Fig.~\ref{fig:optimal_resolution}, we varied the input image resolution from 720 to 3840 (4K) 
	with a step size of about 200. From left to right, we show the Mean Matching Accuracy 
	(MMA)~\cite{Vassilieios_CVPR17_HPatches,Xinghui_NeurIPS20_DRC} plots for the cases of illumination challenges, viewpoint 
	challenges, and overall. The native resolution of the HPatches dataset is reported in Tab. 2 of the main paper.
	
	We observe that the low re-sampling resolution has a major impact on the accuracy in the viewpoint challenges. In contrast, 
	for illumination challenges, low resolution performs relatively well for the low error band ($<3$ pixels). However, the increased 
	re-sampling resolution leads to better performance on the illumination challenge at the cost of a small decrease at the low 
	error band. This is possibly due to stronger ambiguity in the local region in illumination scenarios. For example, the lighting 
	changes introduce blur around many key points when transitioning from day to night. As the resolution increases, the 
	predicted key point locations are more likely to converge towards more repeatable but less accurate areas. As far as the large 
	error band is concerned, the performance of our method saturates for the illumination scenario while increasing for the 
	viewpoint challenges as the re-sampling resolution increases. The area under the MMA curve is also provided to measure the 
	overall accuracy. It can be seen that the performance gain using higher re-sampling resolution saturated around 2600 to 3400 
	with the peak performance at resolution 3000. Note that Fig. 8 in the main paper provides a clear visualisation of the overall 
	performance and Fig~\ref{fig:optimal_resolution} provides individual plots for each tested resolution.
	
	We have also evaluated different re-sampling resolutions on the InLoc~\cite{taira2018inloc} dataset in 
	Fig.~\ref{fig:inloc_resolution_comparison}. It can be seen that high-resolution images result in better relocalisation accuracy in 
	terms of the translation error.
	
\begin{figure}[h]
		\centering
		\includegraphics[width=1.0\textwidth]{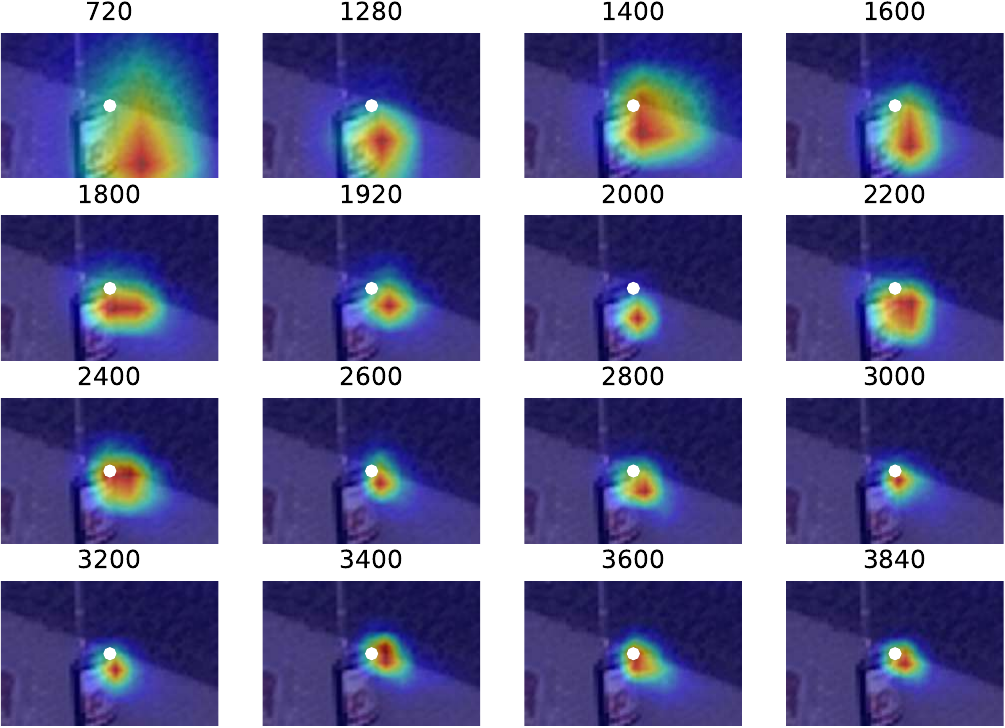}
		\caption{Produced keypoint heatmap from the correlation tensor overlayed at a reference image. The ground truth location 
		of the query keypoint is denoted in white.}
		\label{fig:hpatches_heatmap}
	\end{figure}

	Fig.~\ref{fig:hpatches_heatmap}  shows the heatmaps of predicted target point using input images of various resolutions. The 
	ground truth match is marked with a white dot. It can be seen that higher re-sampling resolutions consistently reduce the 
	uncertainty indicated by the size of the coloured blob. However, as the resolution further increases over 3000, the prediction 
	becomes over-confident towards a close but inaccurate location. This is possibly because of the reduced receptive field of the 
	feature backbone relative to the original image. 
	
\begin{figure}[h]
		\centering
		\includegraphics[width=1.0\textwidth]{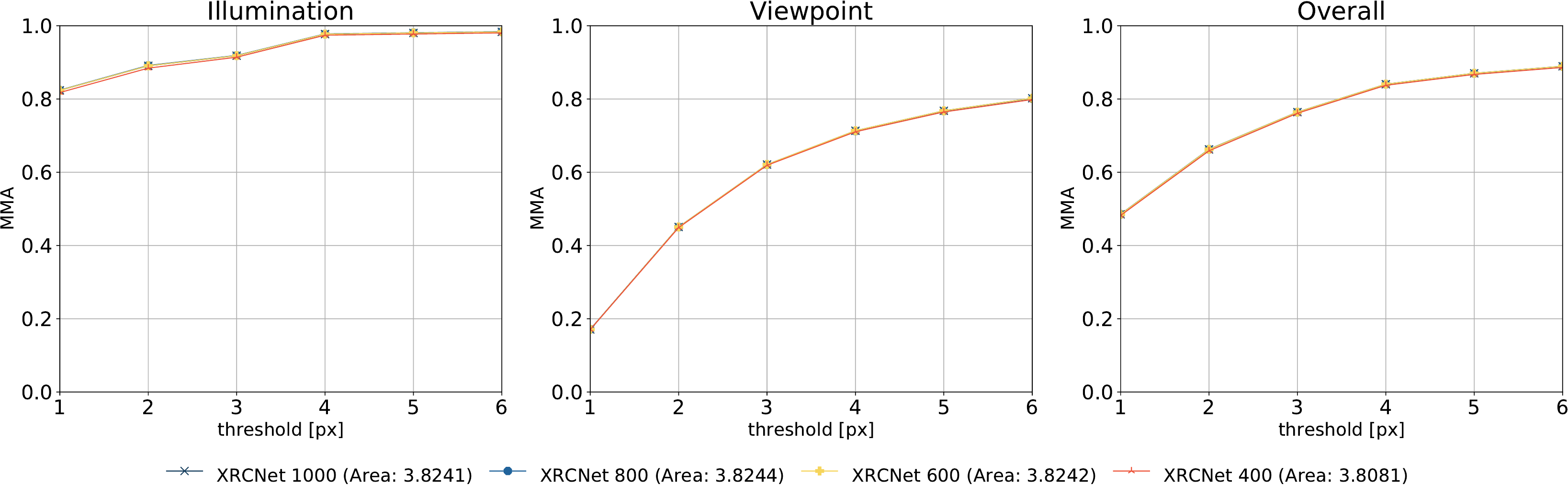}
		\caption{Training $\mathbb{X}$RCNet with re-sampled image resolution of 400px to 1,000 px at every 200px.}
		\label{fig:train_resolution_comparison}
	\end{figure}

	In addition to evaluating the re-sampling impact for inference, we also trained our correspondence network using various 
	training image resolutions. Surprisingly, increasing the input resolution during training does not improve performance, as 
	shown in Fig.~\ref{fig:train_resolution_comparison}. We hypothesise this is because various training image resolutions contain 
	a fixed amount of information that a correspondence network can use. Therefore, we choose to use 400 px resolution during 
	training in order to achieve a fair comparison with other baseline methods. Please note that all methods are trained with a 
	batch size of 16 to accommodate higher resolution in the feature maps.
	
	\section{Qualitative Analysis --- HPatches}\label{Sec_2}
\begin{figure}[t]
		\hspace{0.08\textwidth}
		DualRCNet\hspace{0.16\textwidth}%
		XRCNet\hspace{0.18\textwidth}%
		DualRCNet\hspace{0.16\textwidth}
		XRCNet
		\\[0.25em]
		\par
		
		\noindent
		\includegraphics[width=0.24\textwidth]{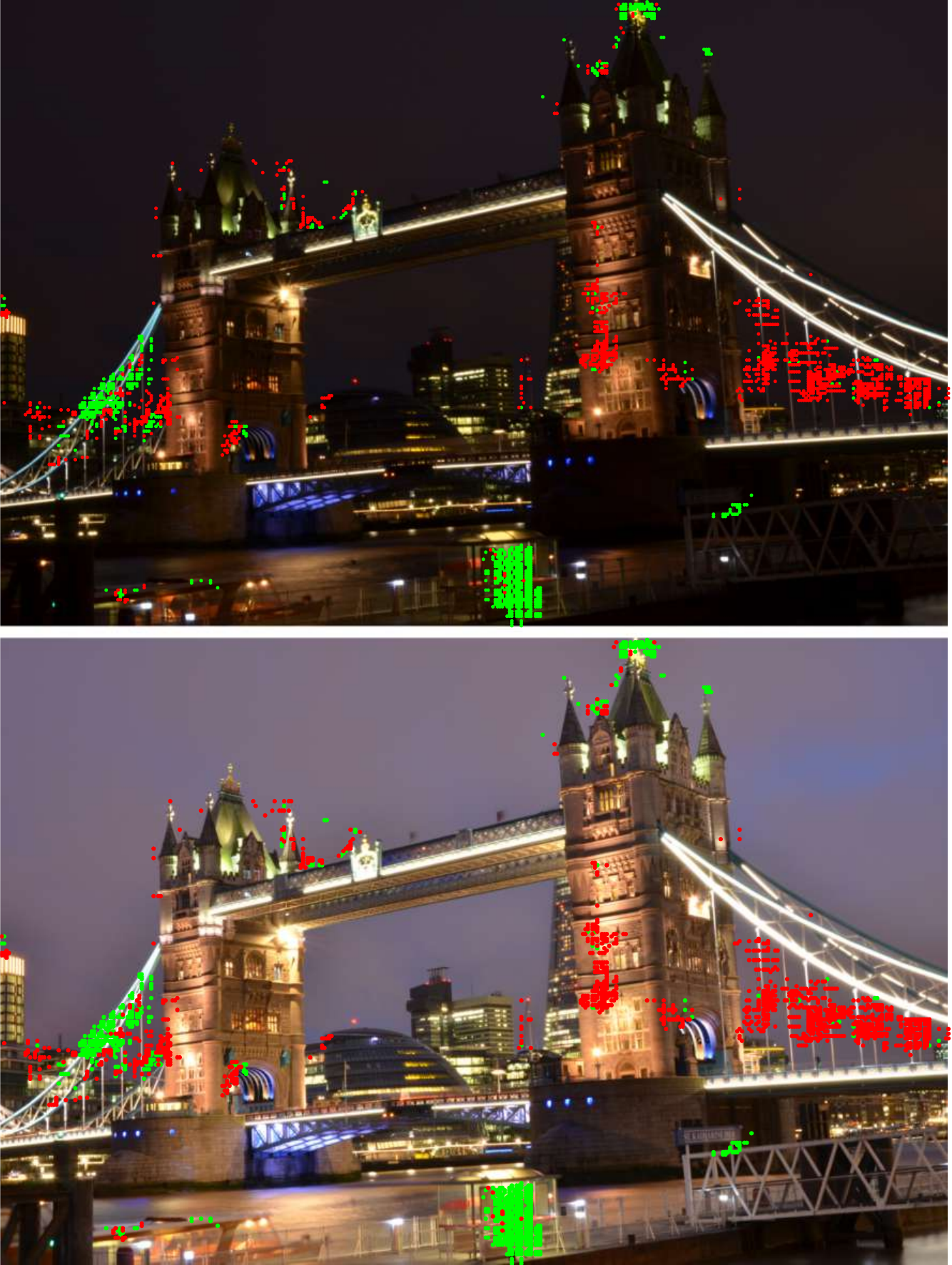}\hspace{0.01\textwidth}%
		\includegraphics[width=0.24\textwidth]{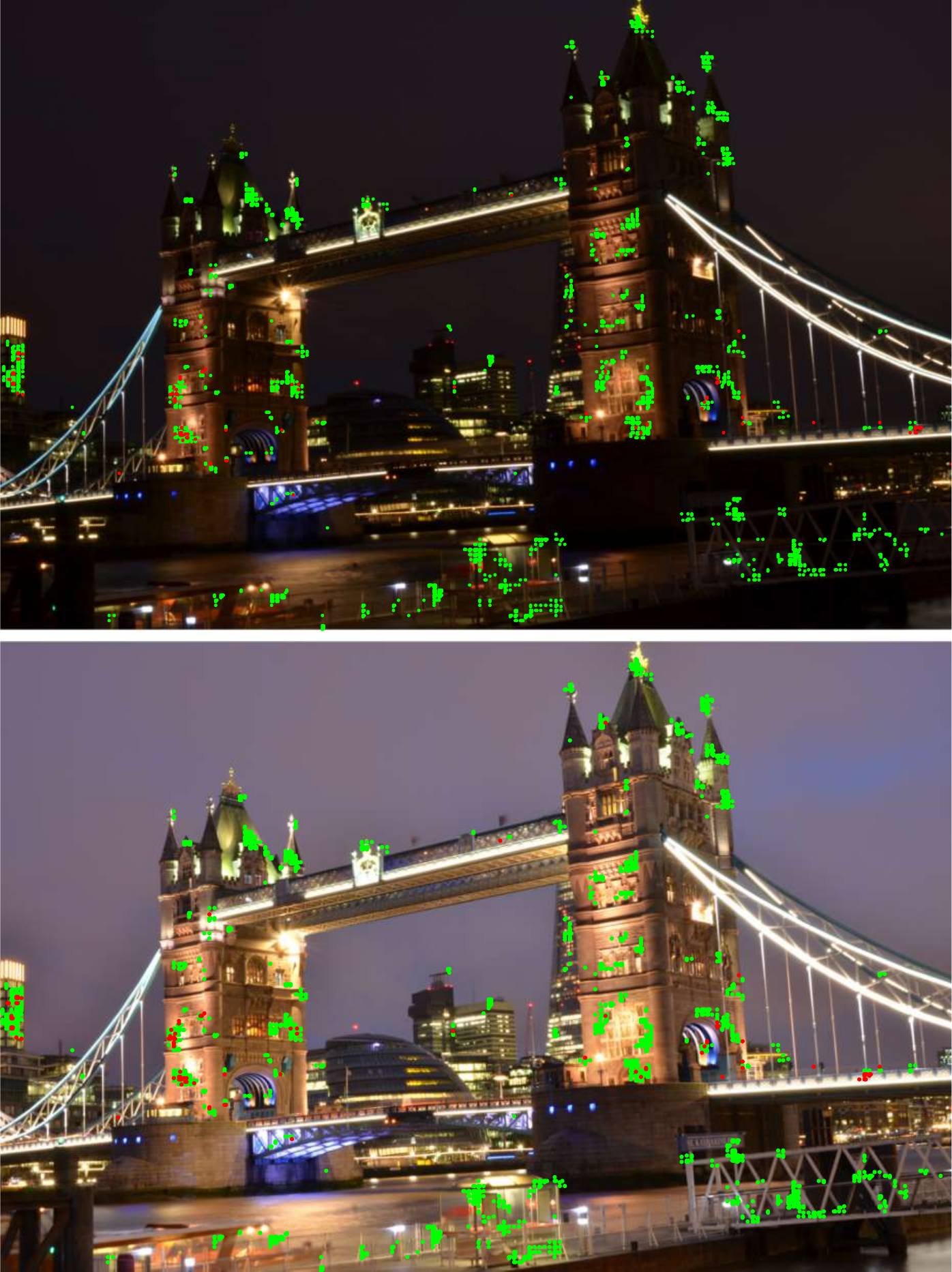}\hspace{0.01\textwidth}%
		\includegraphics[width=0.24\textwidth]{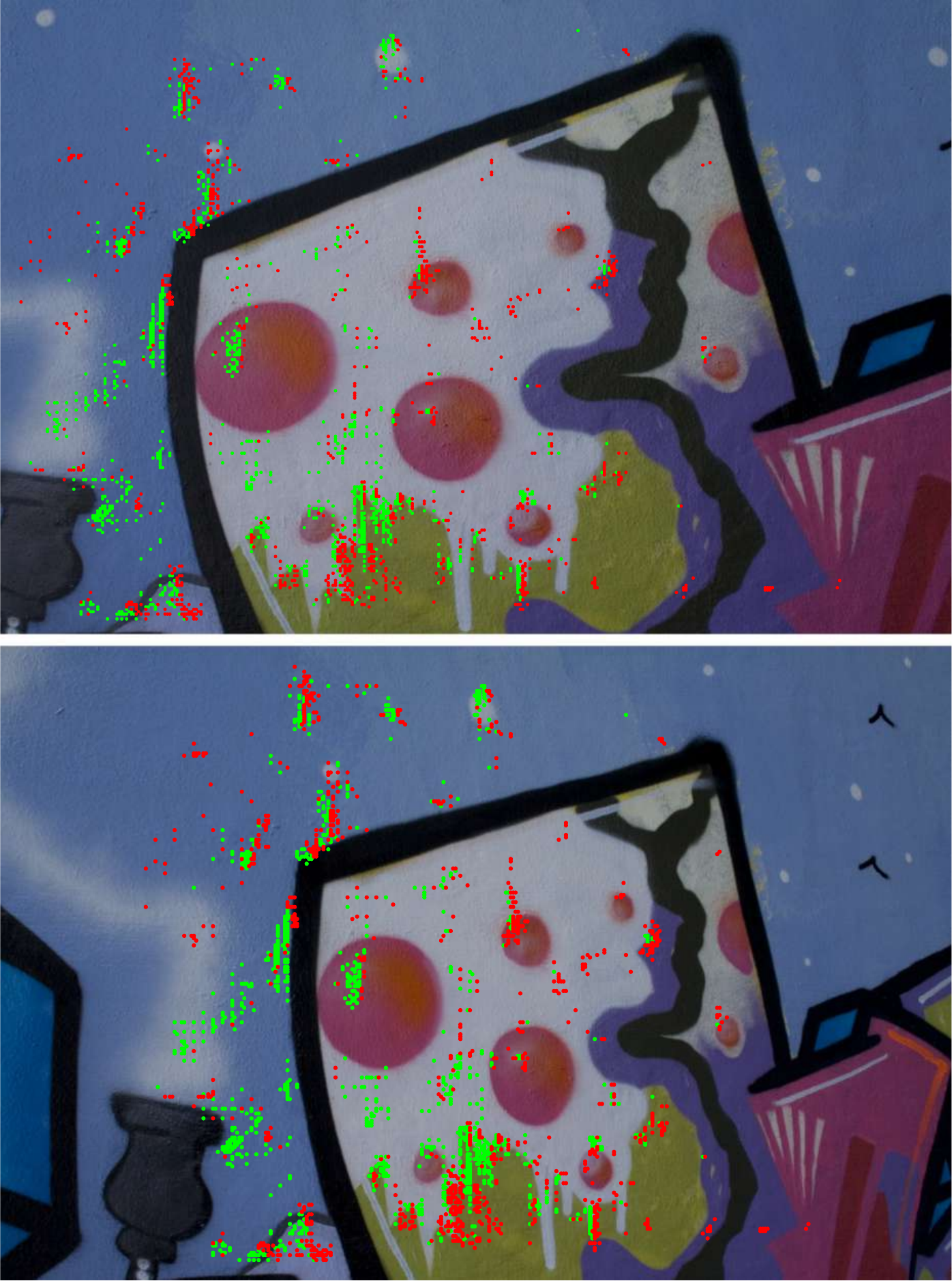}\hspace{0.01\textwidth}%
		\includegraphics[width=0.24\textwidth]{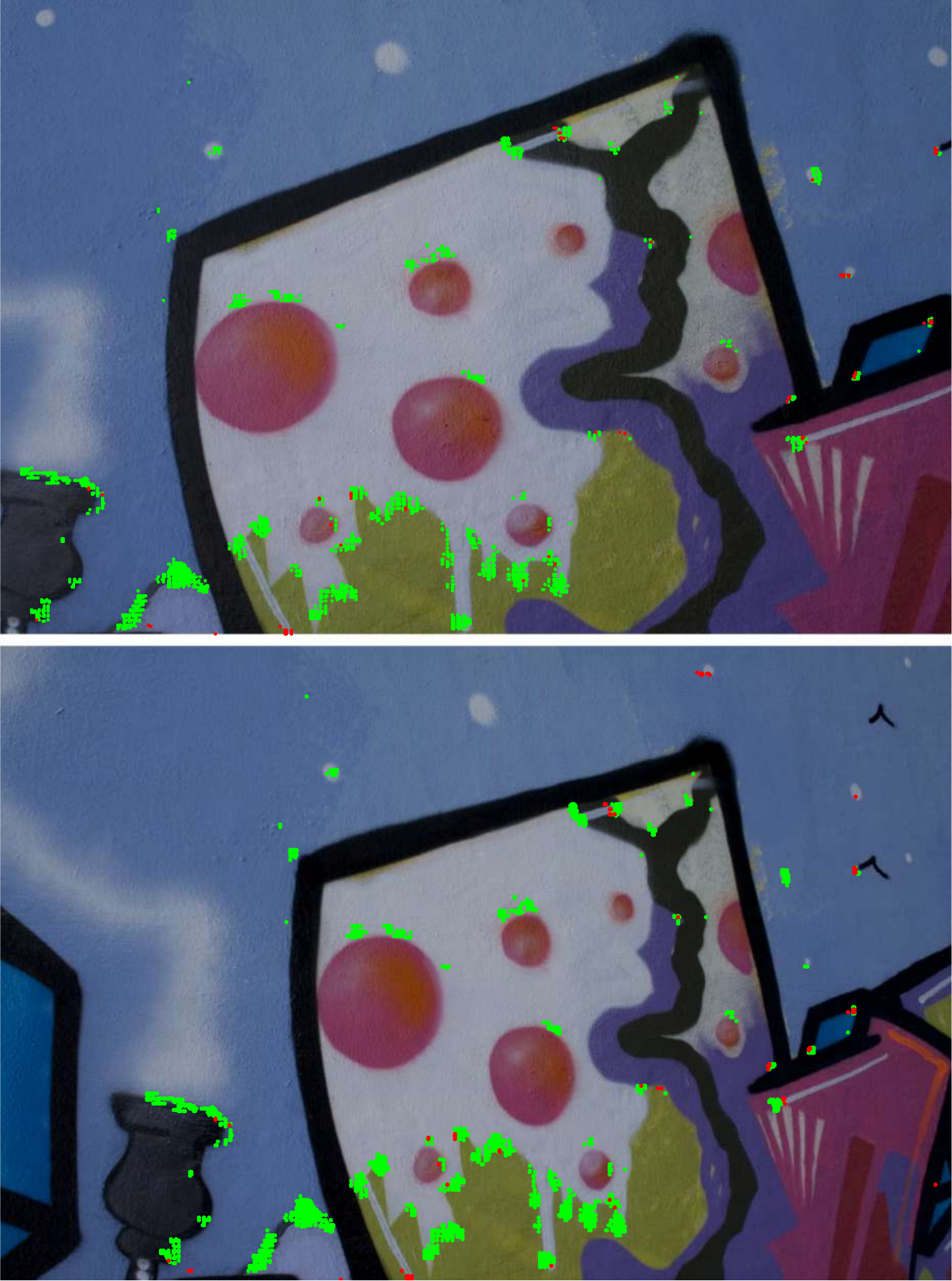}
		\\[1em]
		\includegraphics[width=0.24\textwidth]{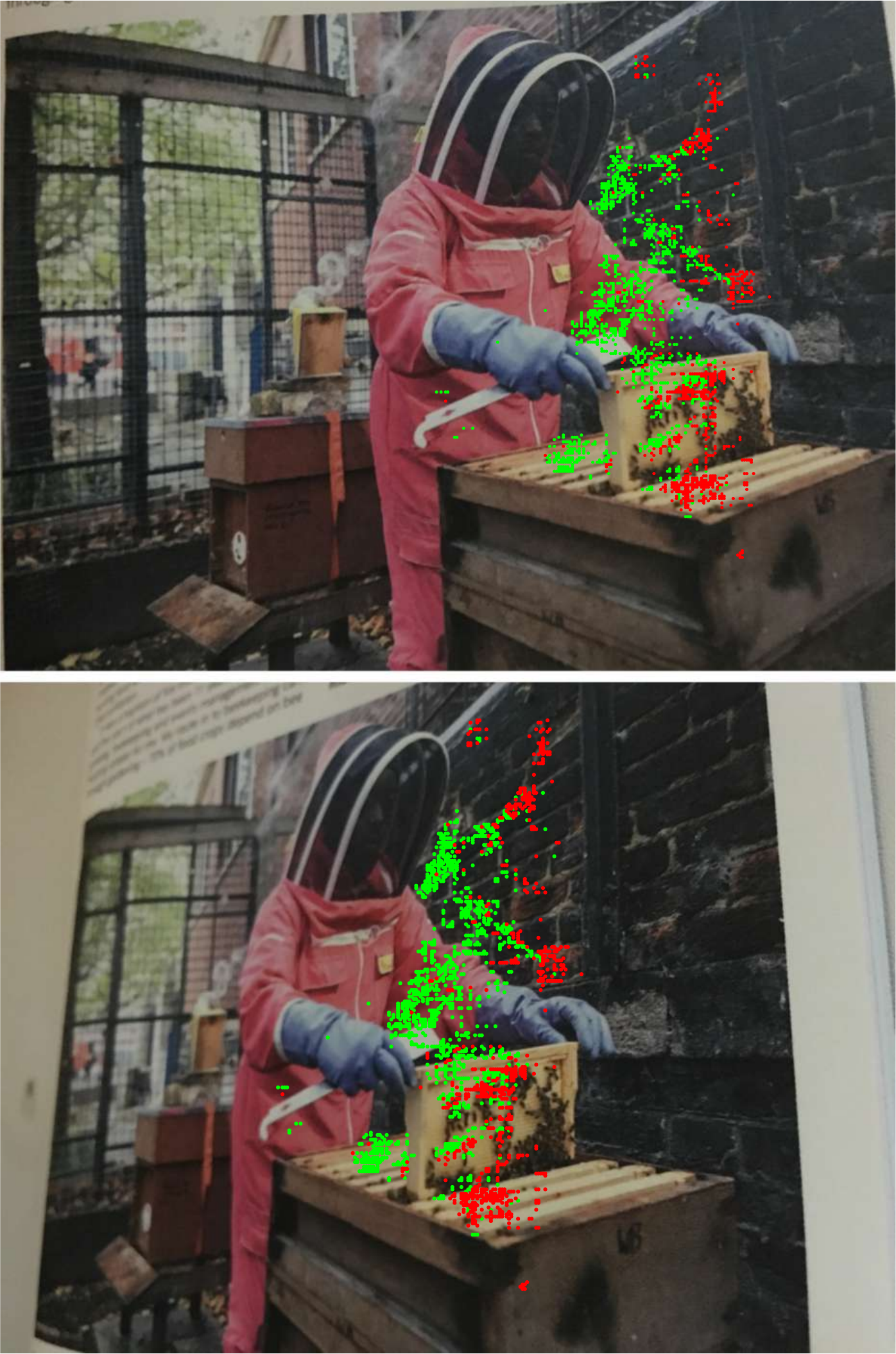}\hspace{0.01\textwidth}%
		\includegraphics[width=0.24\textwidth]{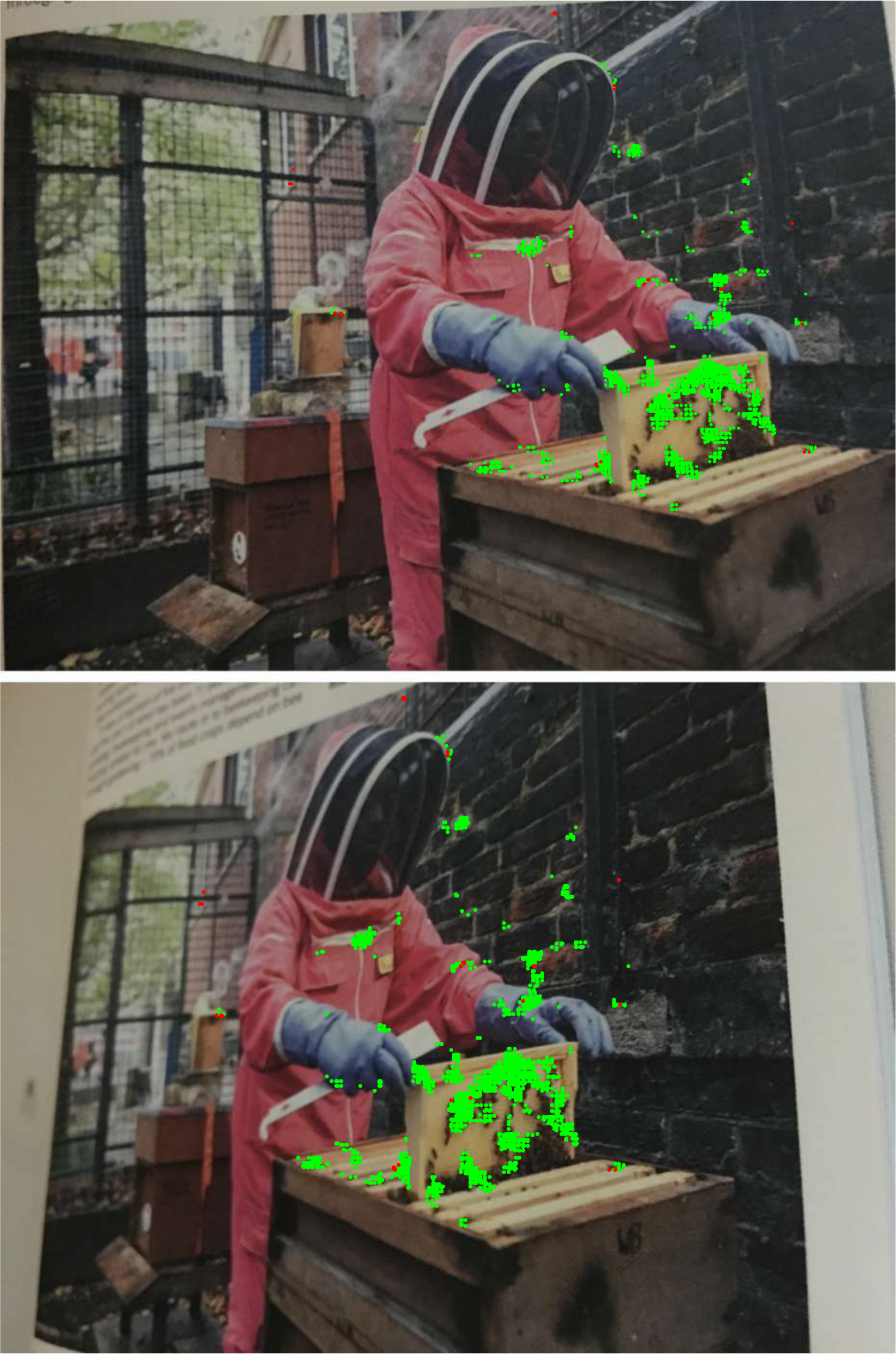}\hspace{0.01\textwidth}%
		\includegraphics[width=0.24\textwidth]{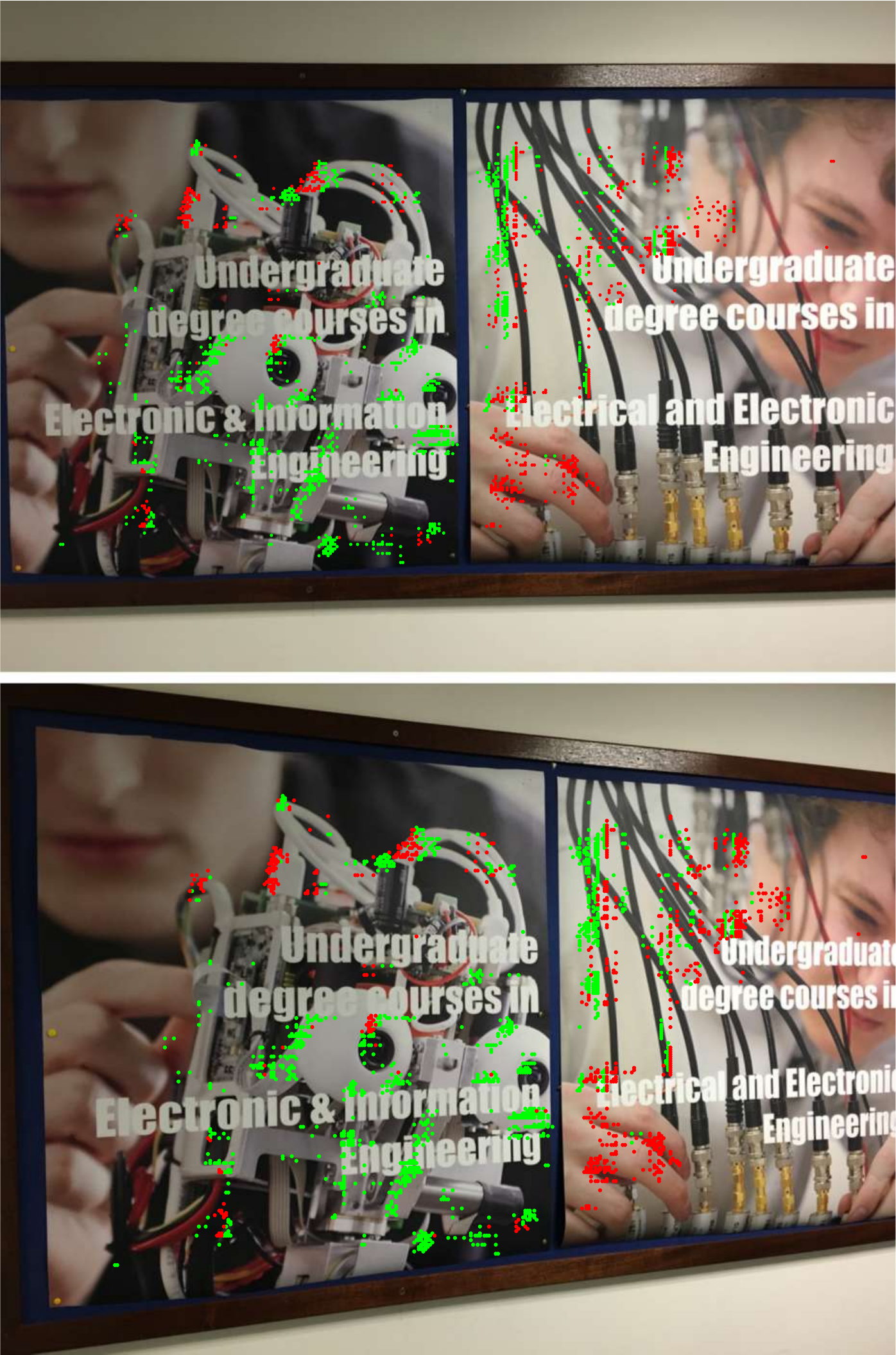}\hspace{0.01\textwidth}%
		\includegraphics[width=0.24\textwidth]{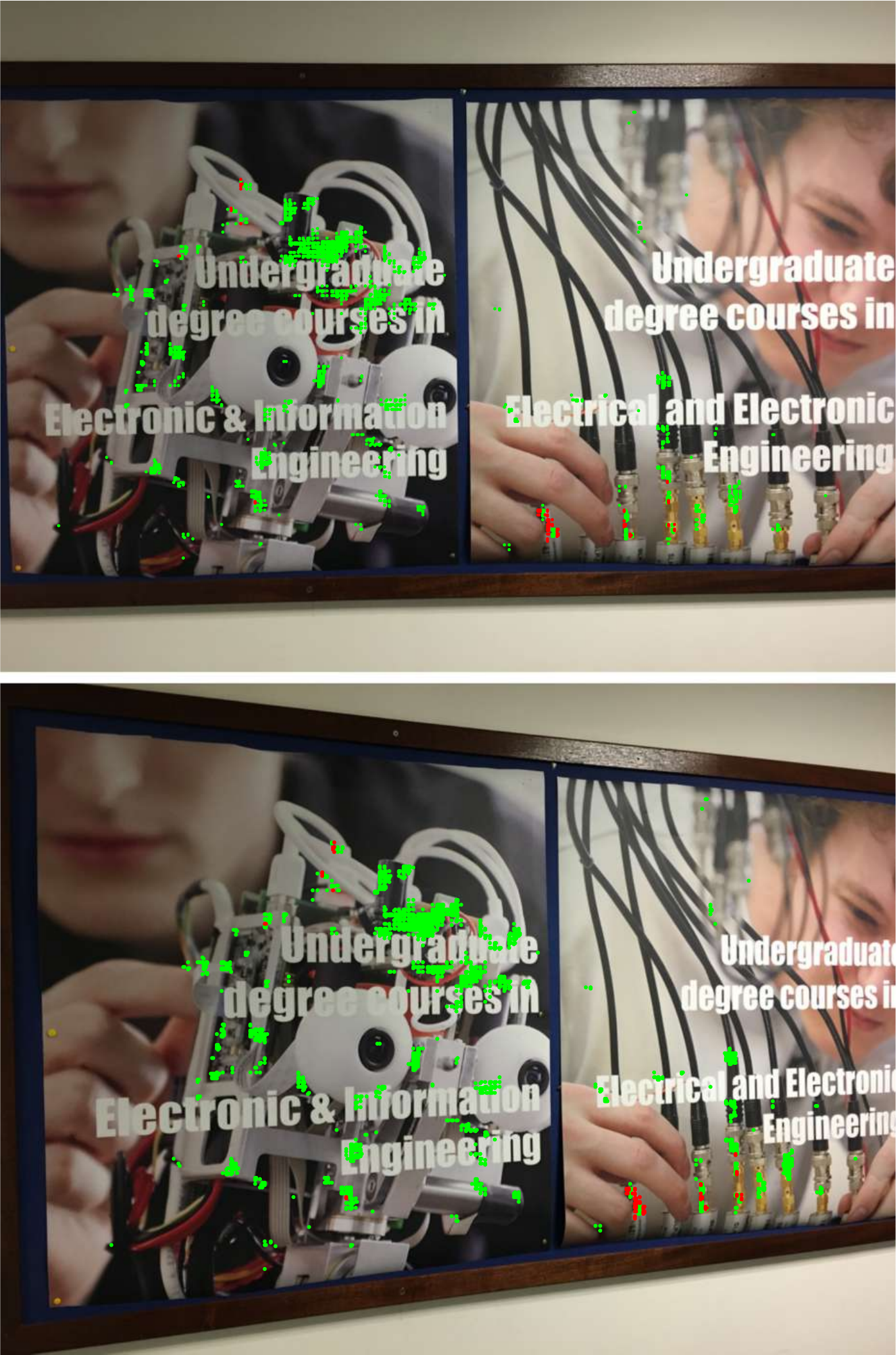}
		\\[1em]
		\includegraphics[width=0.24\textwidth]{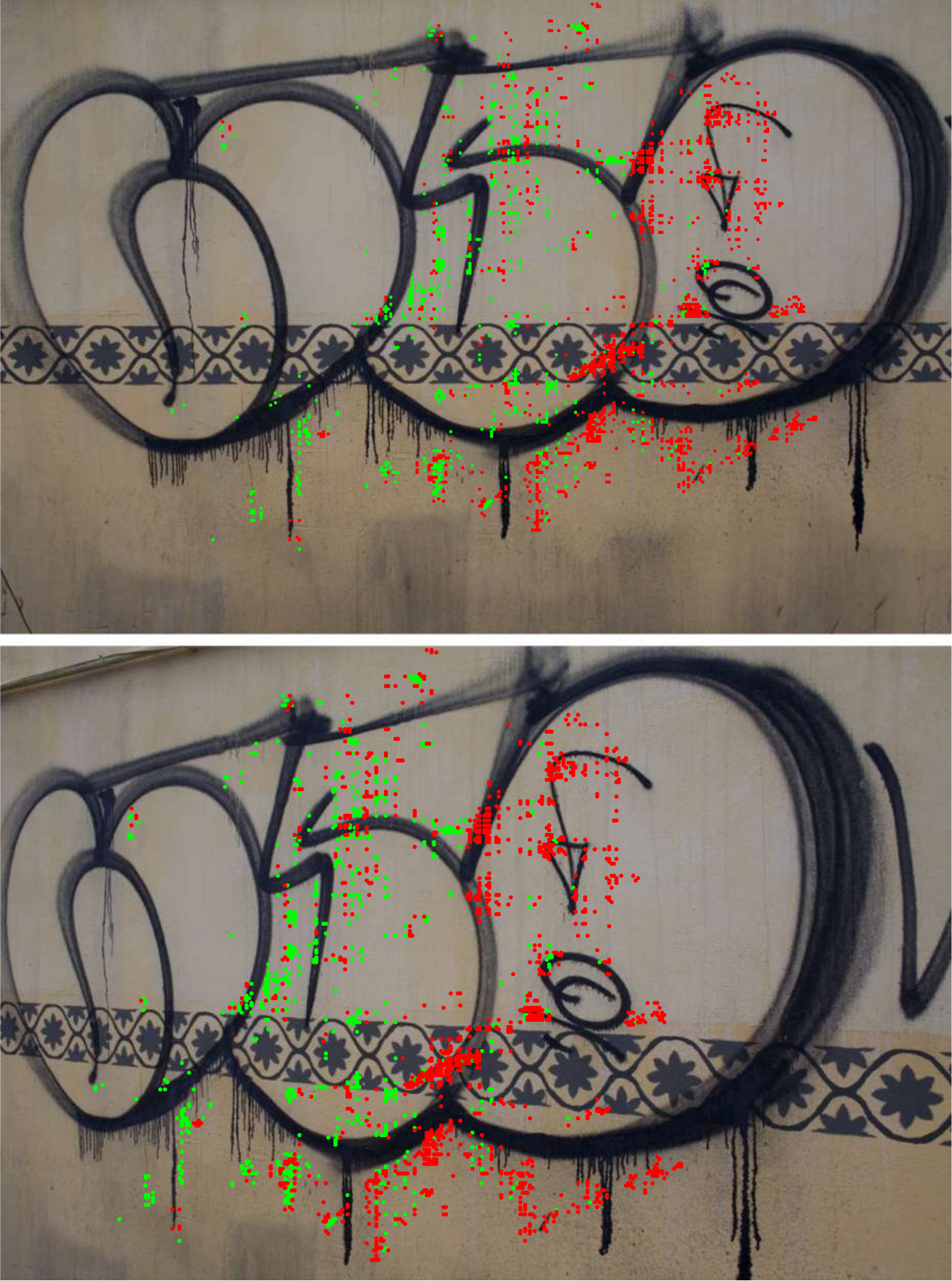}\hspace{0.01\textwidth}%
		\includegraphics[width=0.24\textwidth]{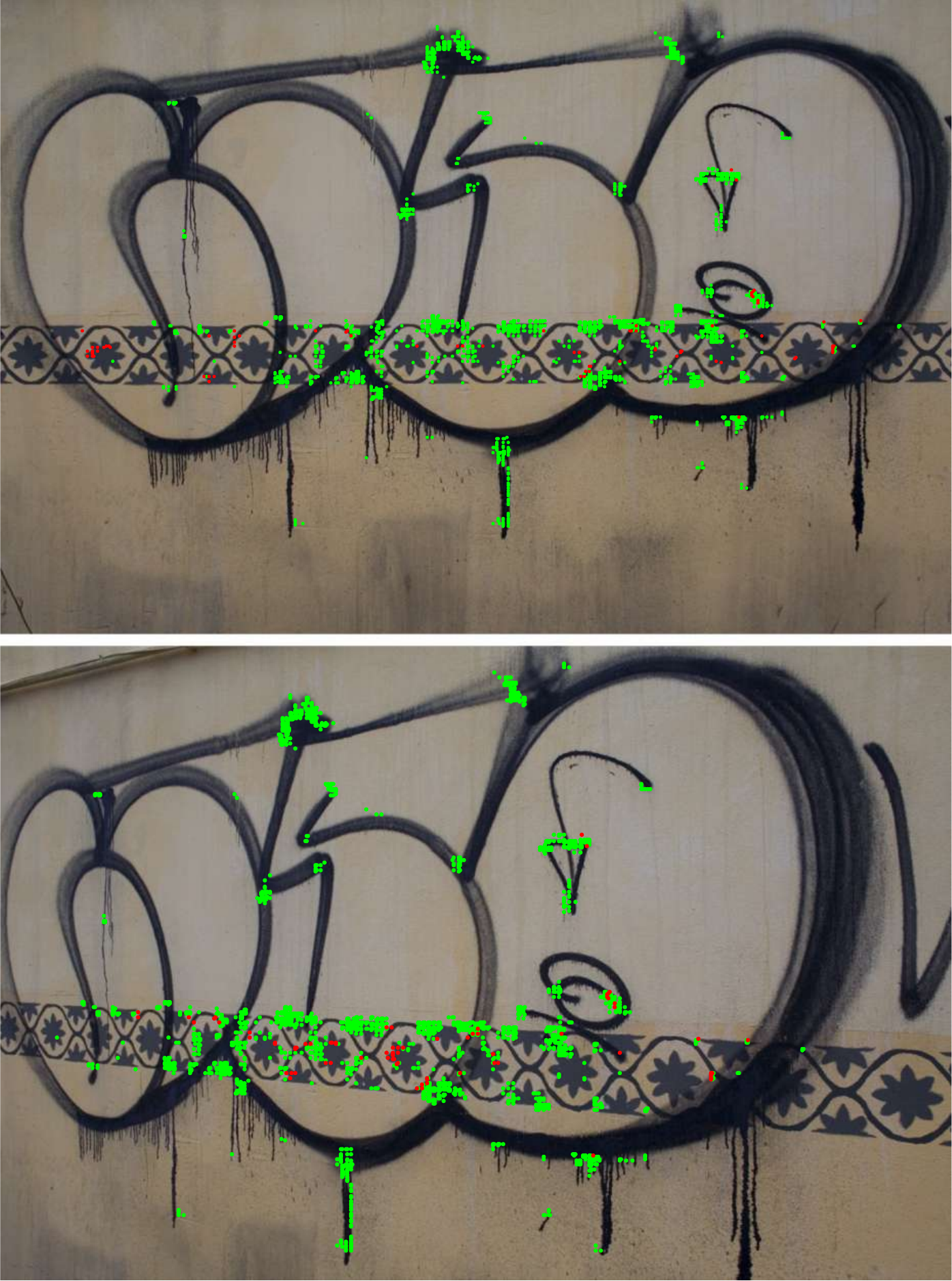}\hspace{0.01\textwidth}%
		\includegraphics[width=0.24\textwidth]{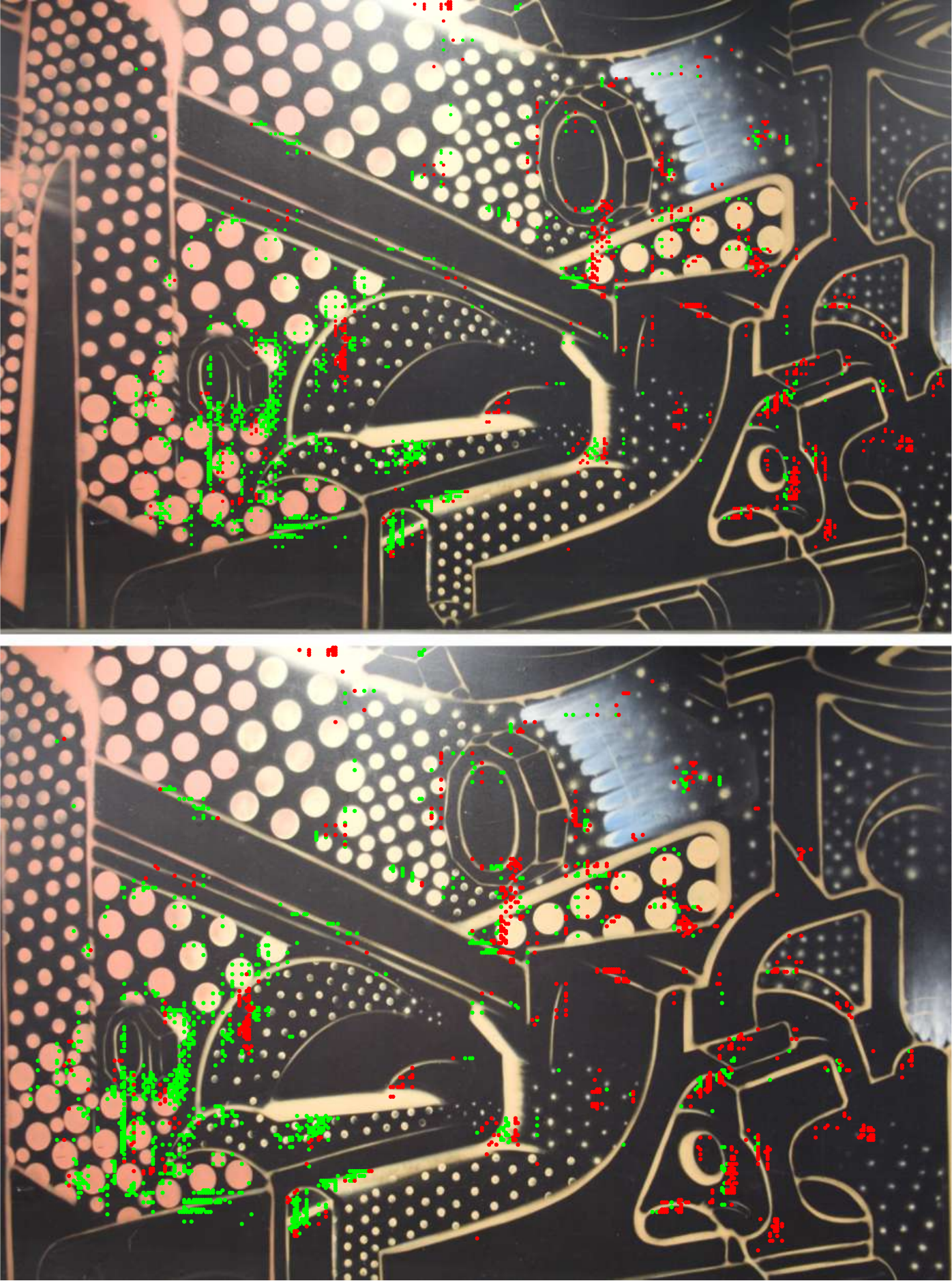}\hspace{0.01\textwidth}%
		\includegraphics[width=0.24\textwidth]{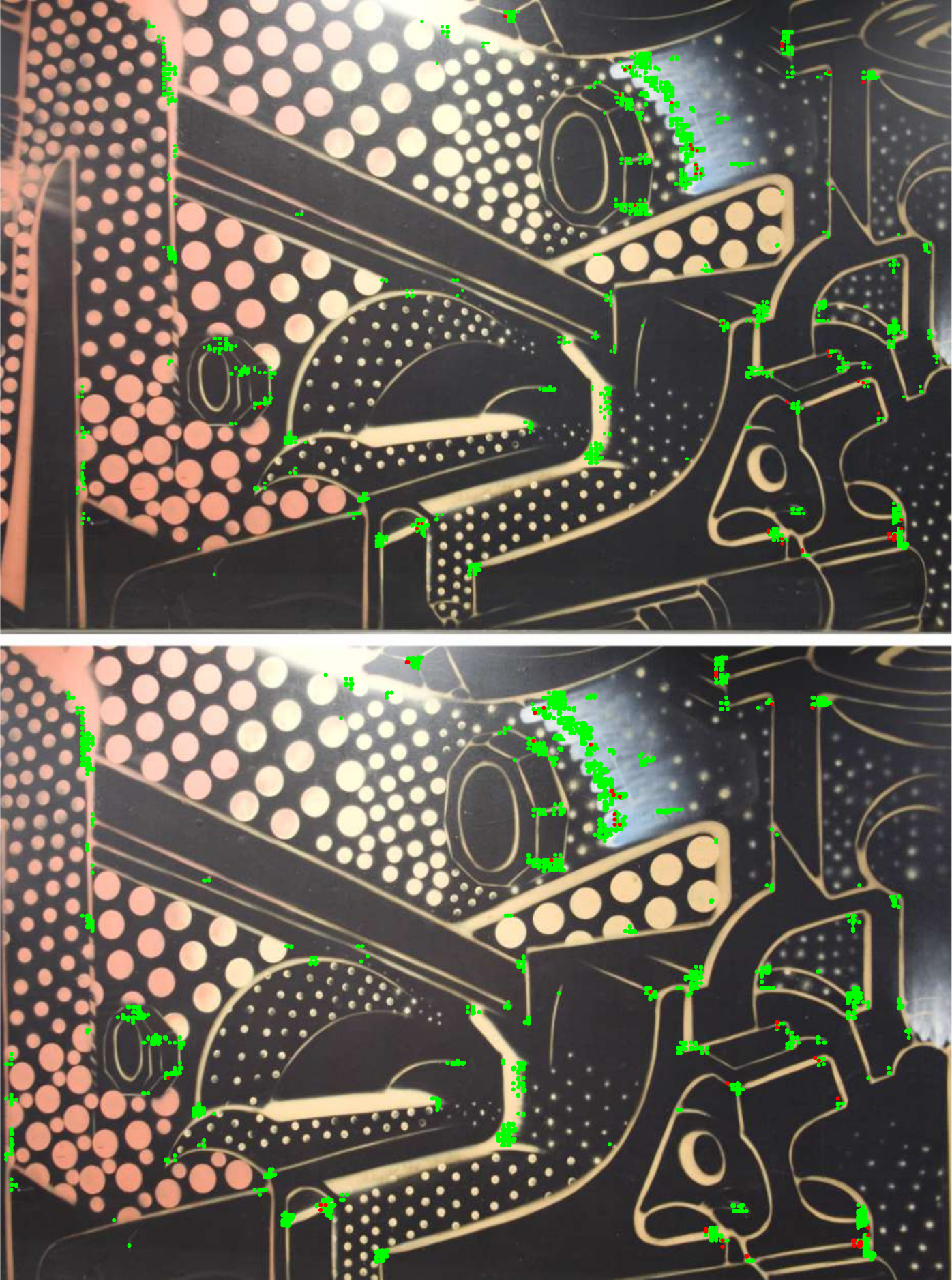}
		\par
		\caption{Qualitative comparison between $\mathbb X$RCNet and DualRCNet on HPatches. The green dots represent the 
		correct matches whose errors are within 3 pixels, and red dots the incorrect matches. $\mathbb X$RCNet produces more 
		correct matches out of the top 2000 matches than DualRCNet.}
		\label{fig:hpatches_qualitative}
	\end{figure}

\begin{figure}[t]
		\hspace{0.08\textwidth}
		SparseNC\hspace{0.16\textwidth}%
		XRCNet\hspace{0.18\textwidth}%
		SparseNC\hspace{0.16\textwidth}%
		XRCNet
		\\[0.25em]
		\par
		
		\noindent
		\includegraphics[width=0.24\textwidth]{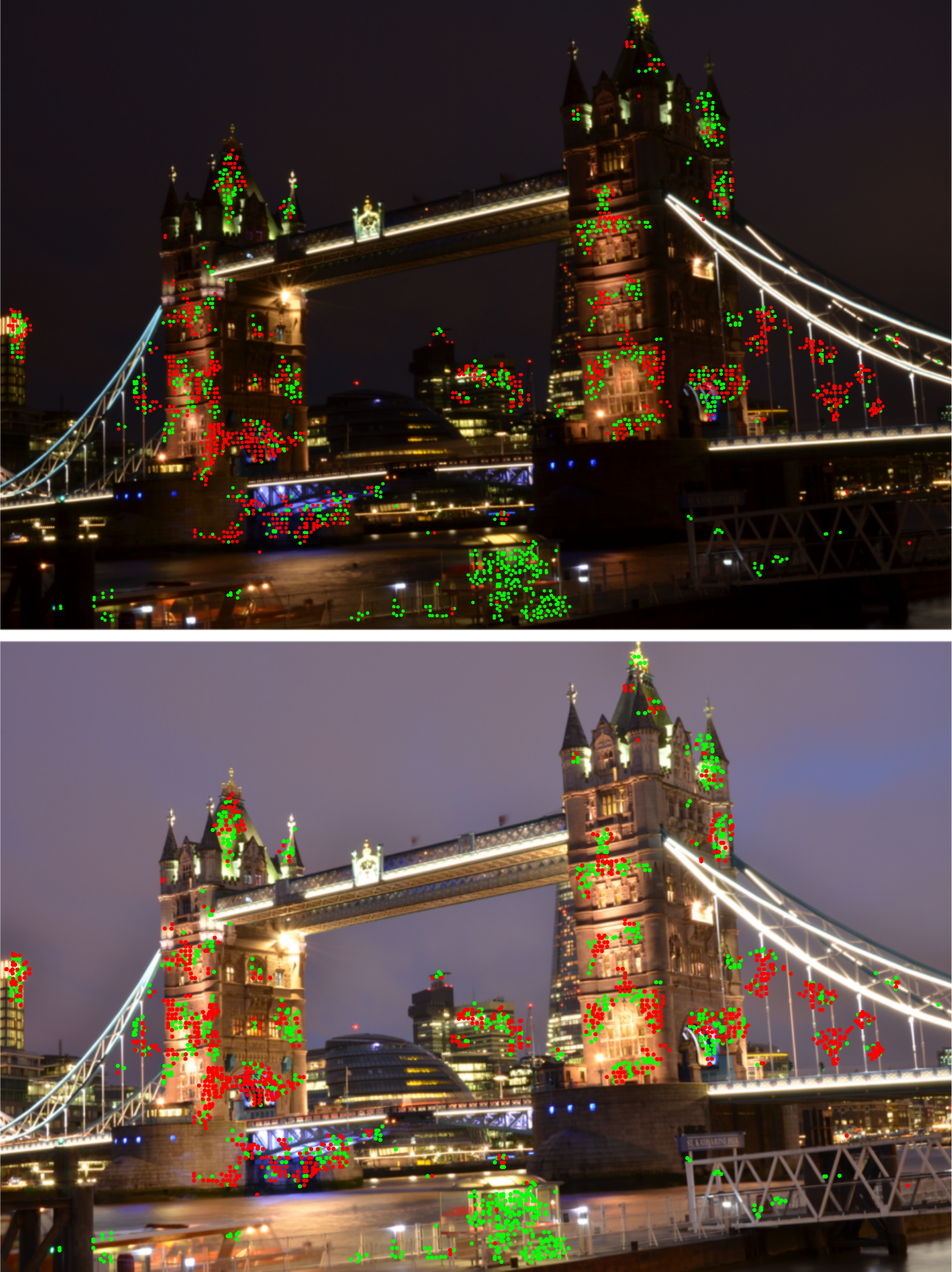}\hspace{0.01\textwidth}%
		\includegraphics[width=0.24\textwidth]{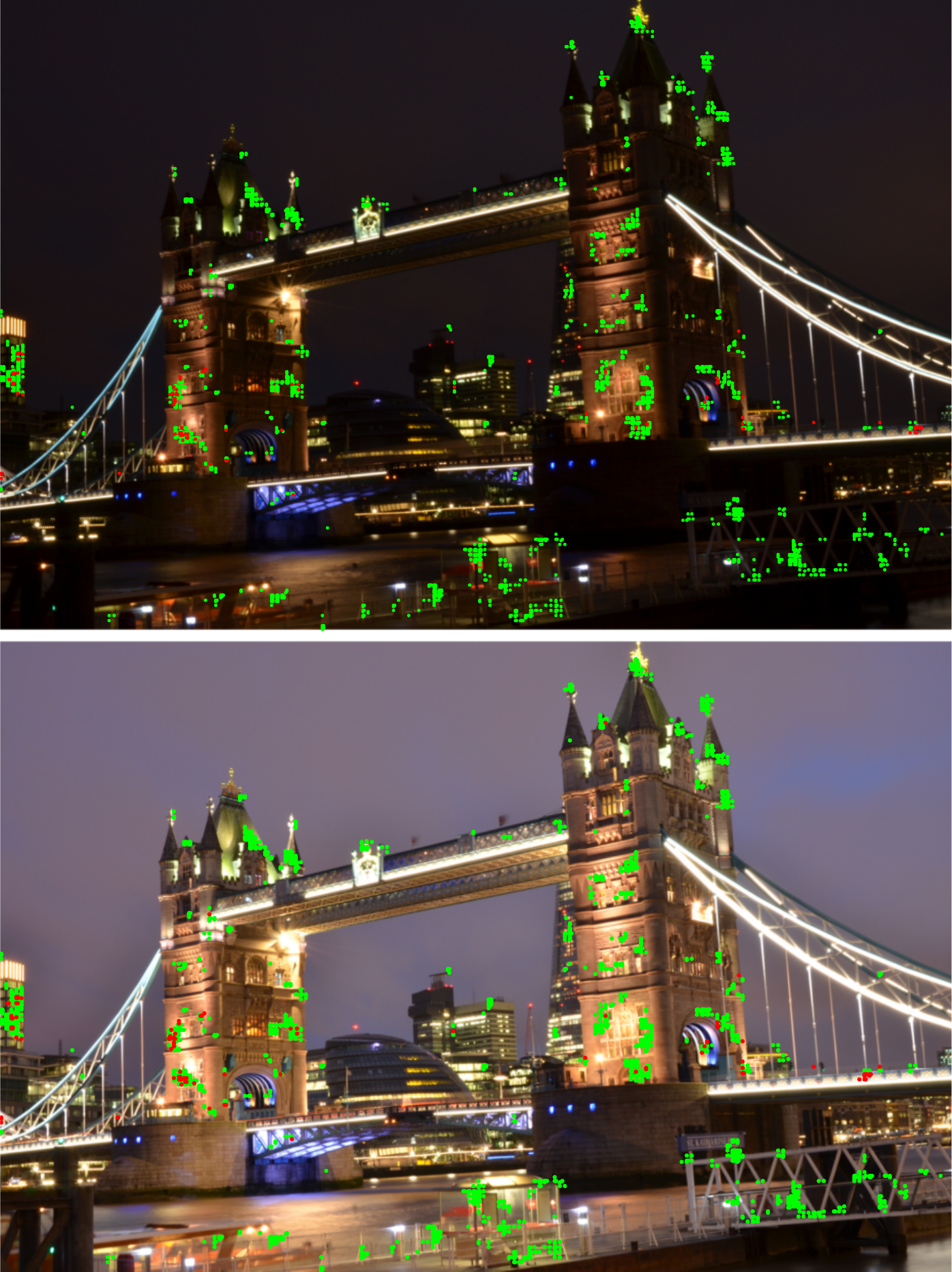}\hspace{0.01\textwidth}%
		\includegraphics[width=0.24\textwidth]{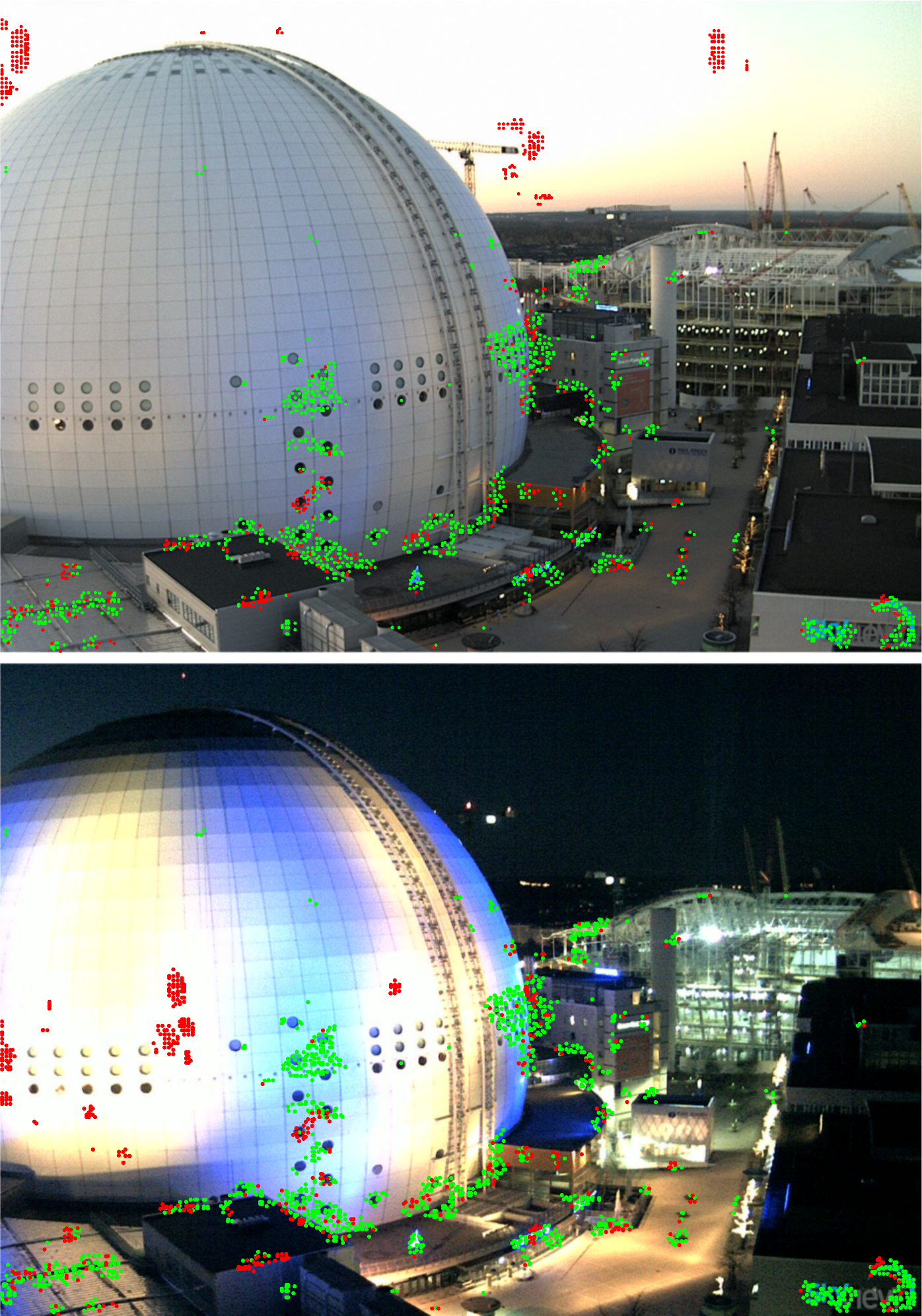}\hspace{0.01\textwidth}%
		\includegraphics[width=0.24\textwidth]{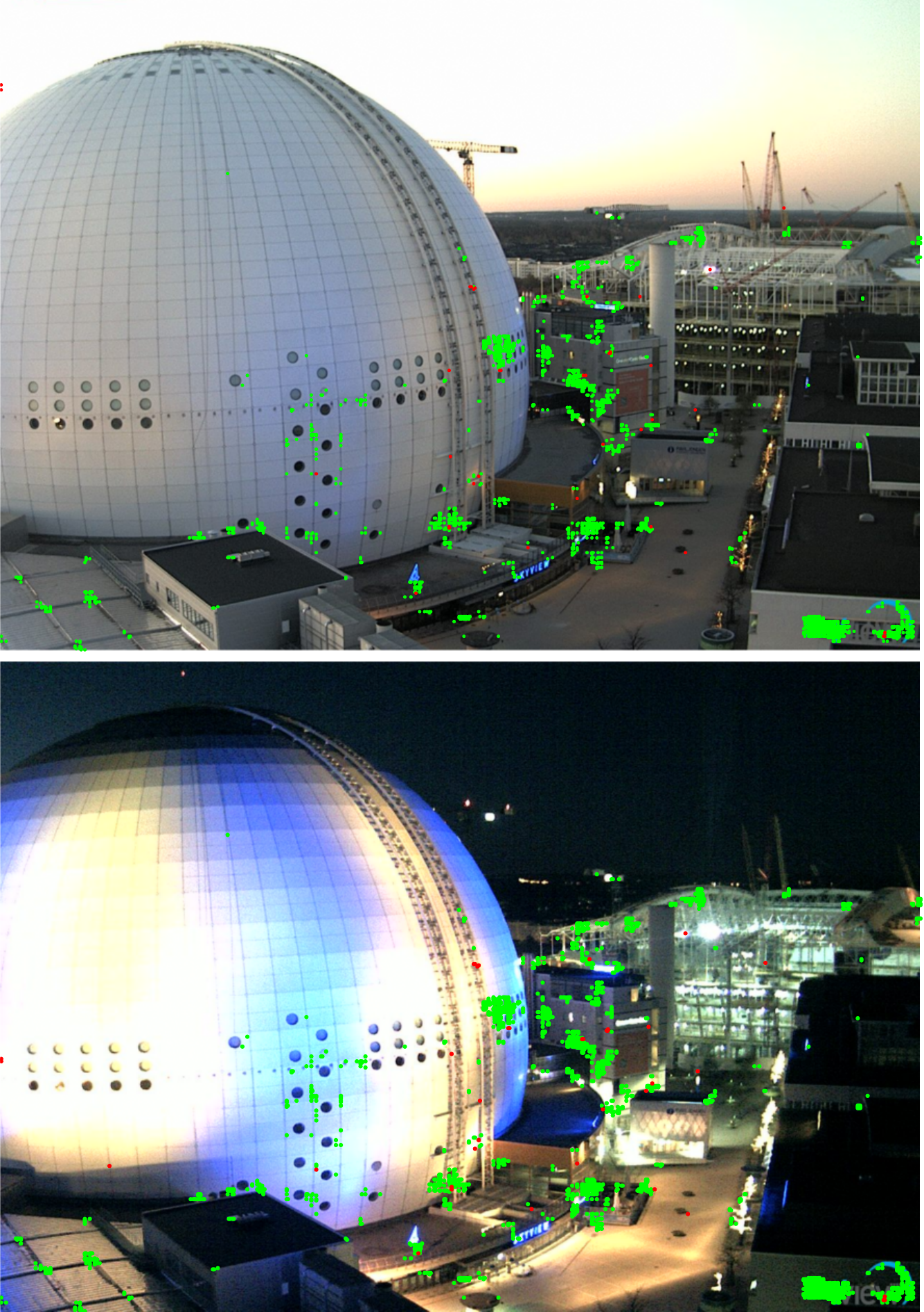}
		\\[1em]
		\includegraphics[width=0.24\textwidth]{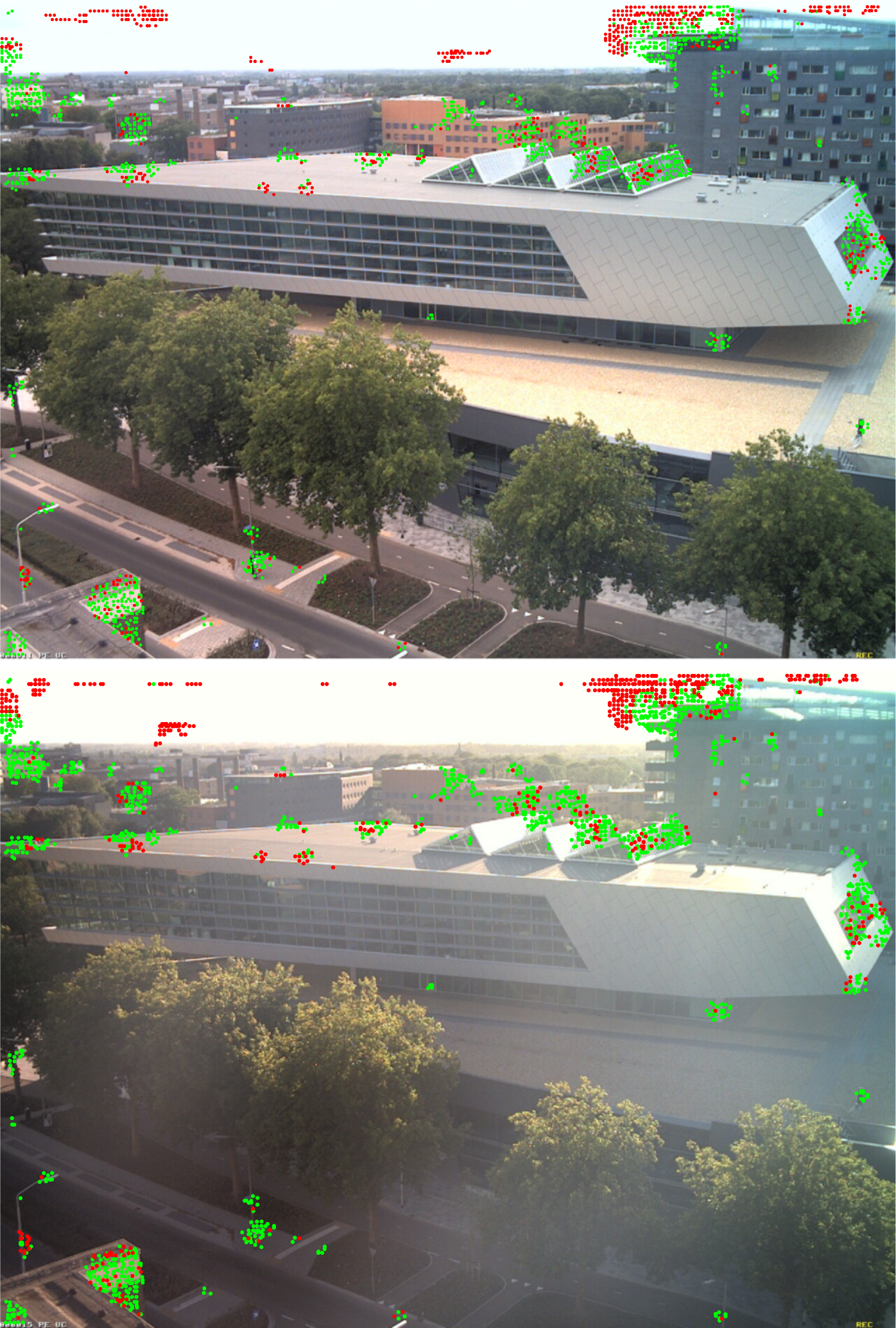}\hspace{0.01\textwidth}%
		\includegraphics[width=0.24\textwidth]{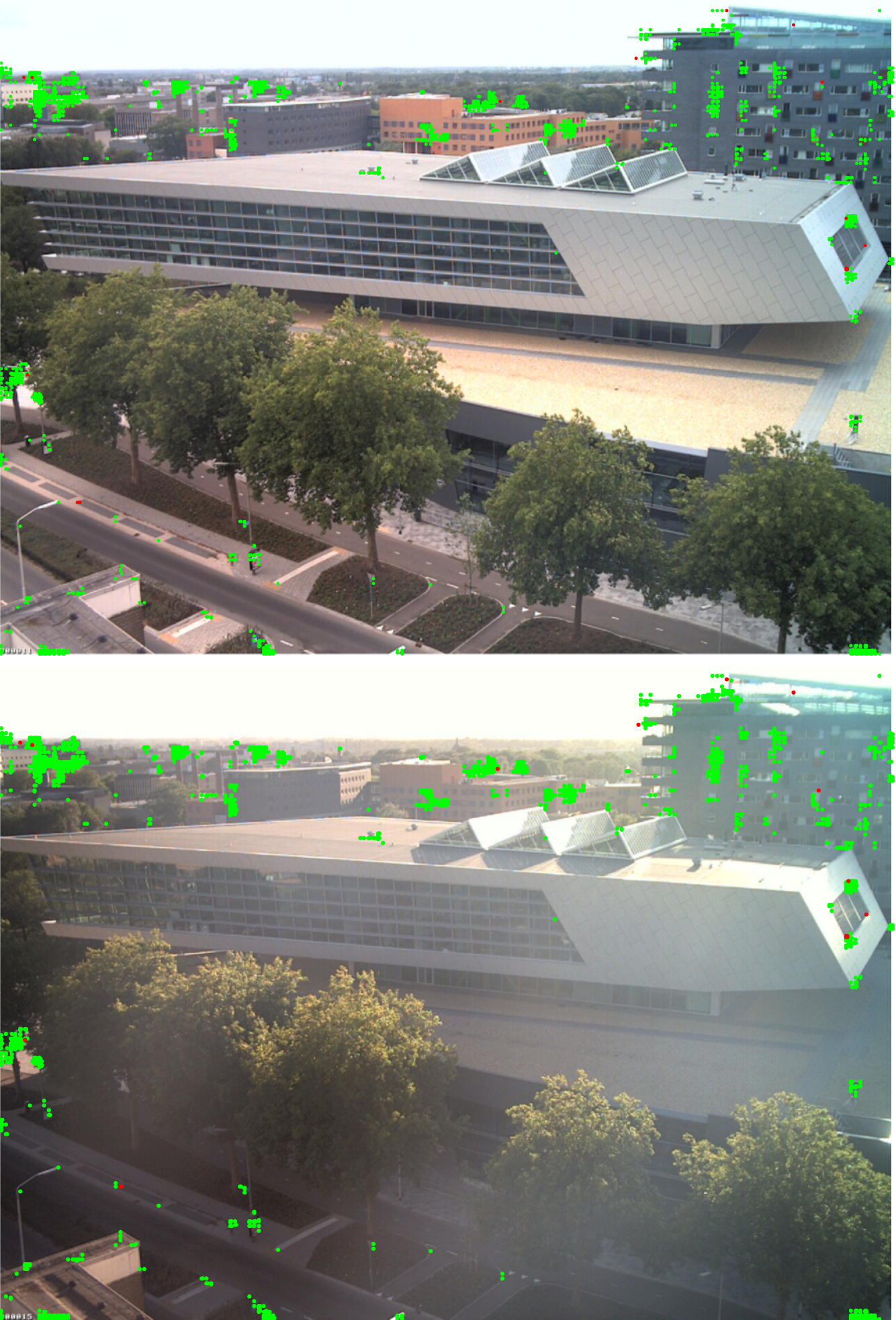}\hspace{0.01\textwidth}%
		\includegraphics[width=0.24\textwidth]{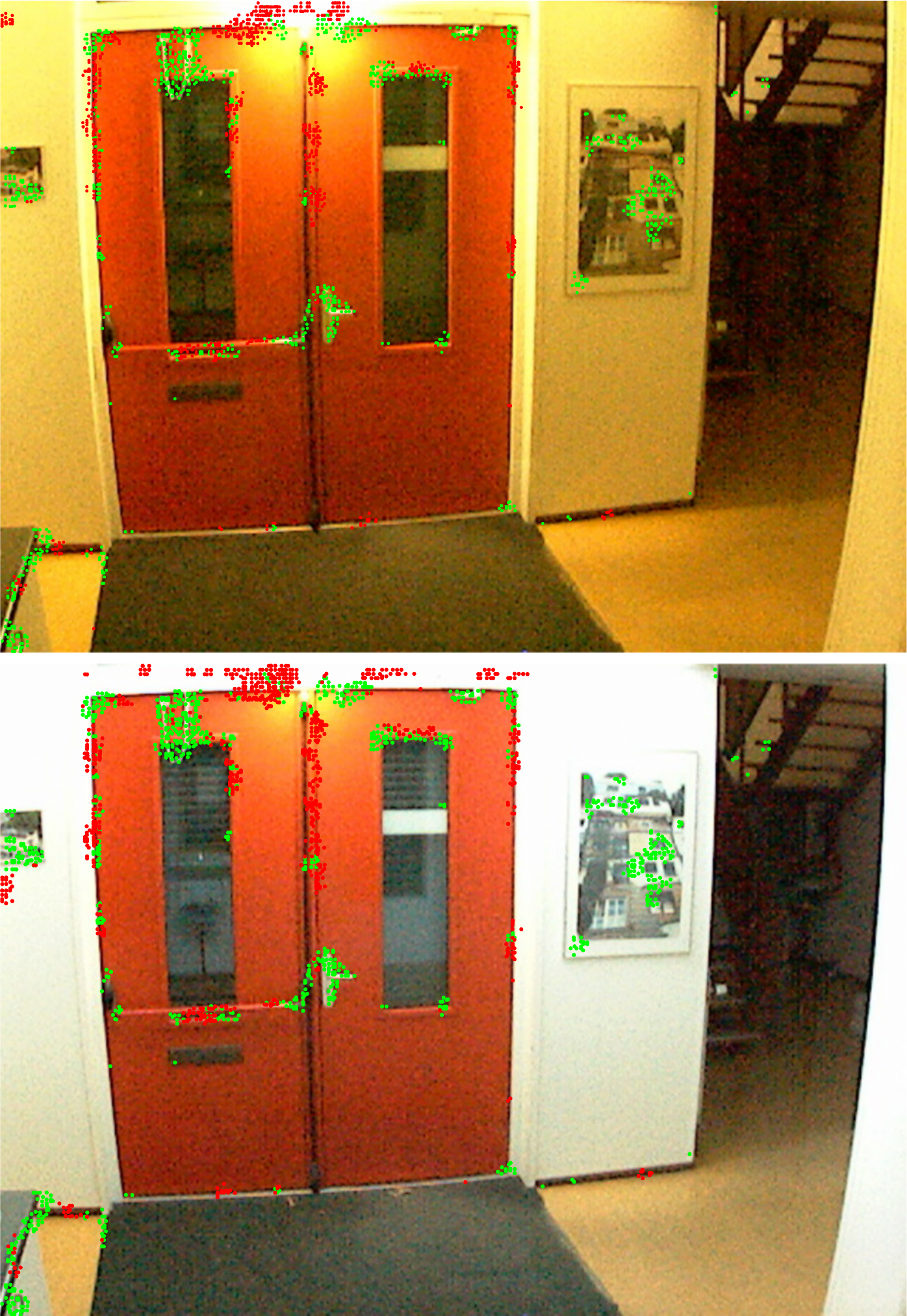}\hspace{0.01\textwidth}%
		\includegraphics[width=0.24\textwidth]{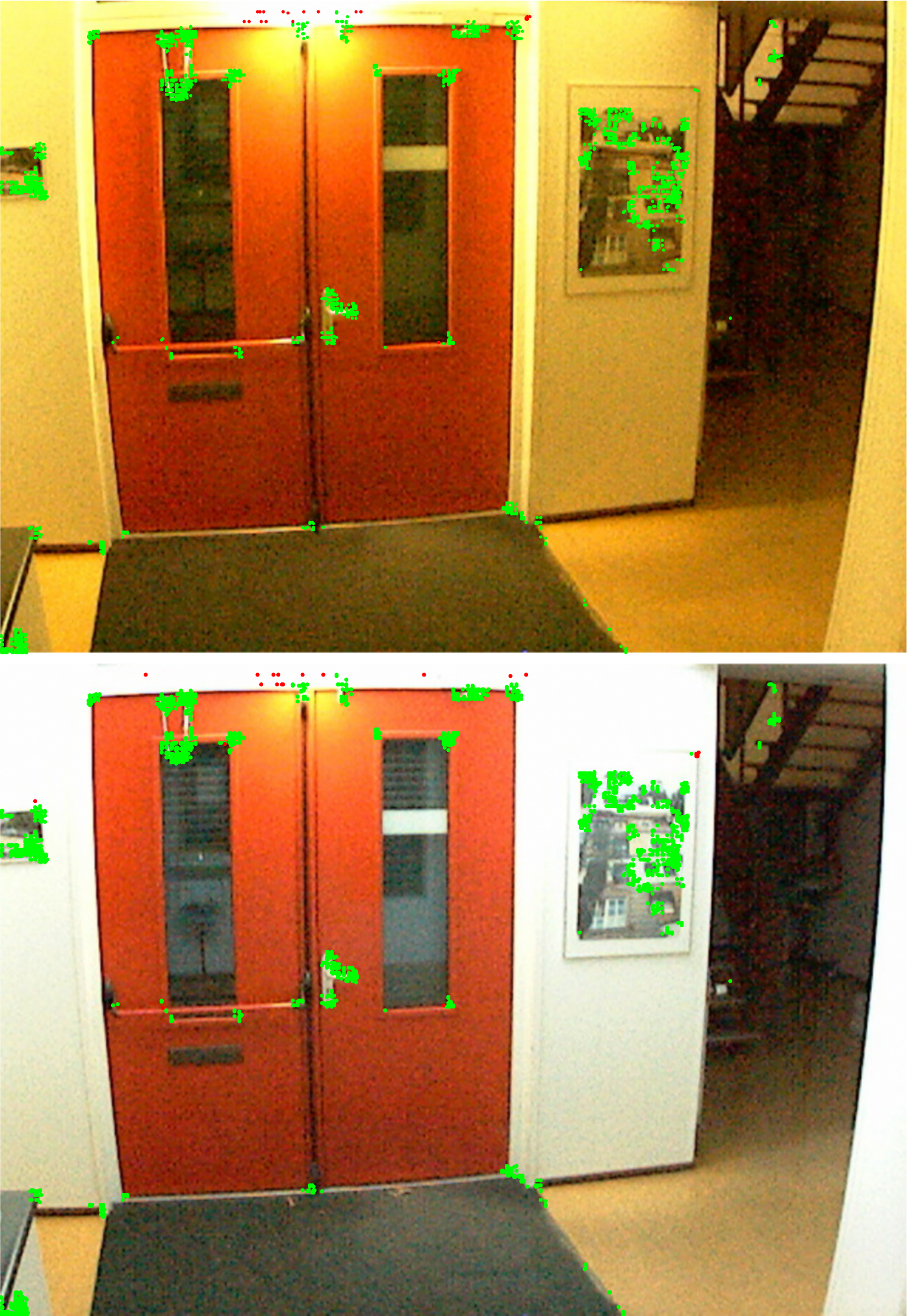}
		\\[1em]
		\includegraphics[width=0.24\textwidth]{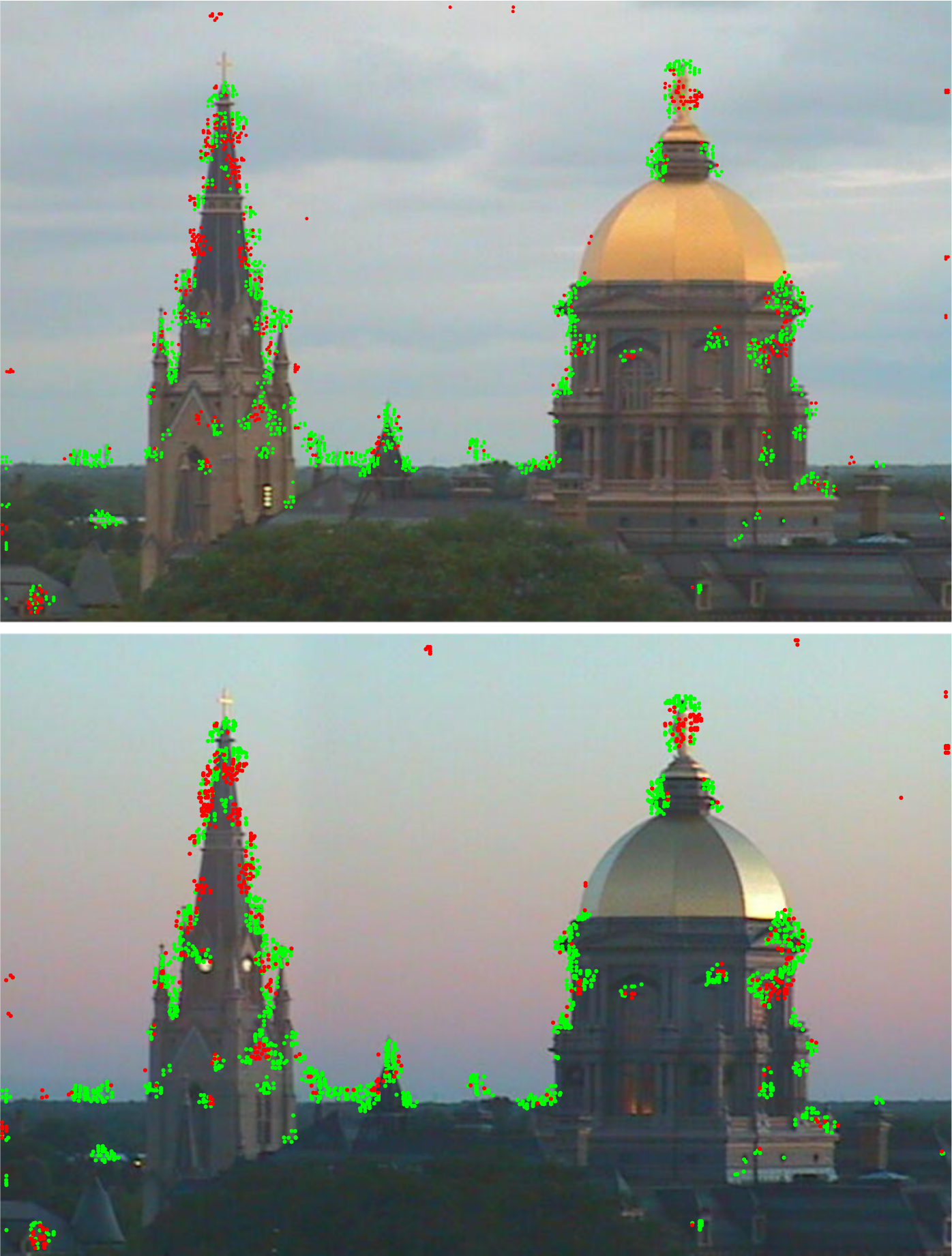}\hspace{0.01\textwidth}%
		\includegraphics[width=0.24\textwidth]{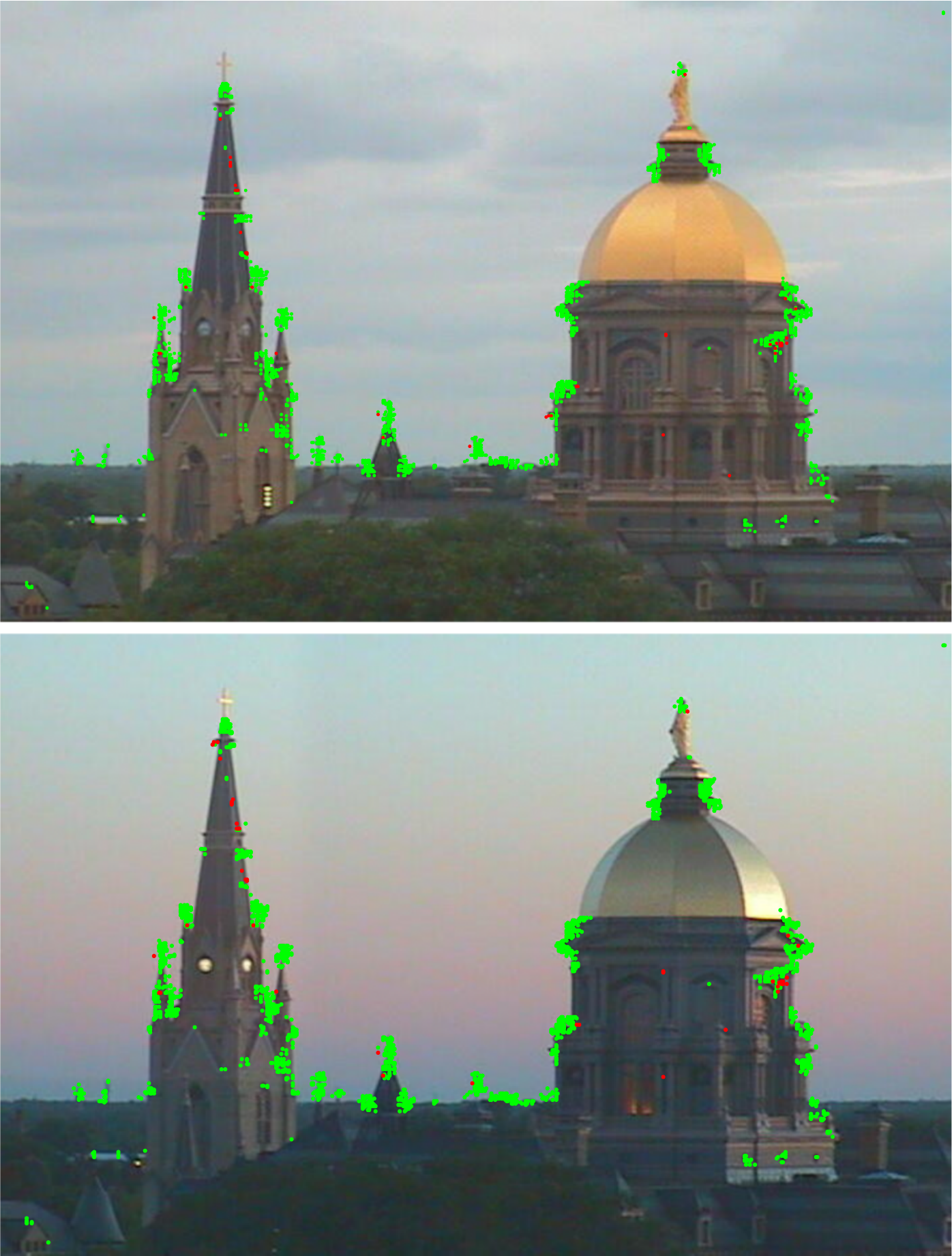}\hspace{0.01\textwidth}%
		\includegraphics[width=0.24\textwidth]{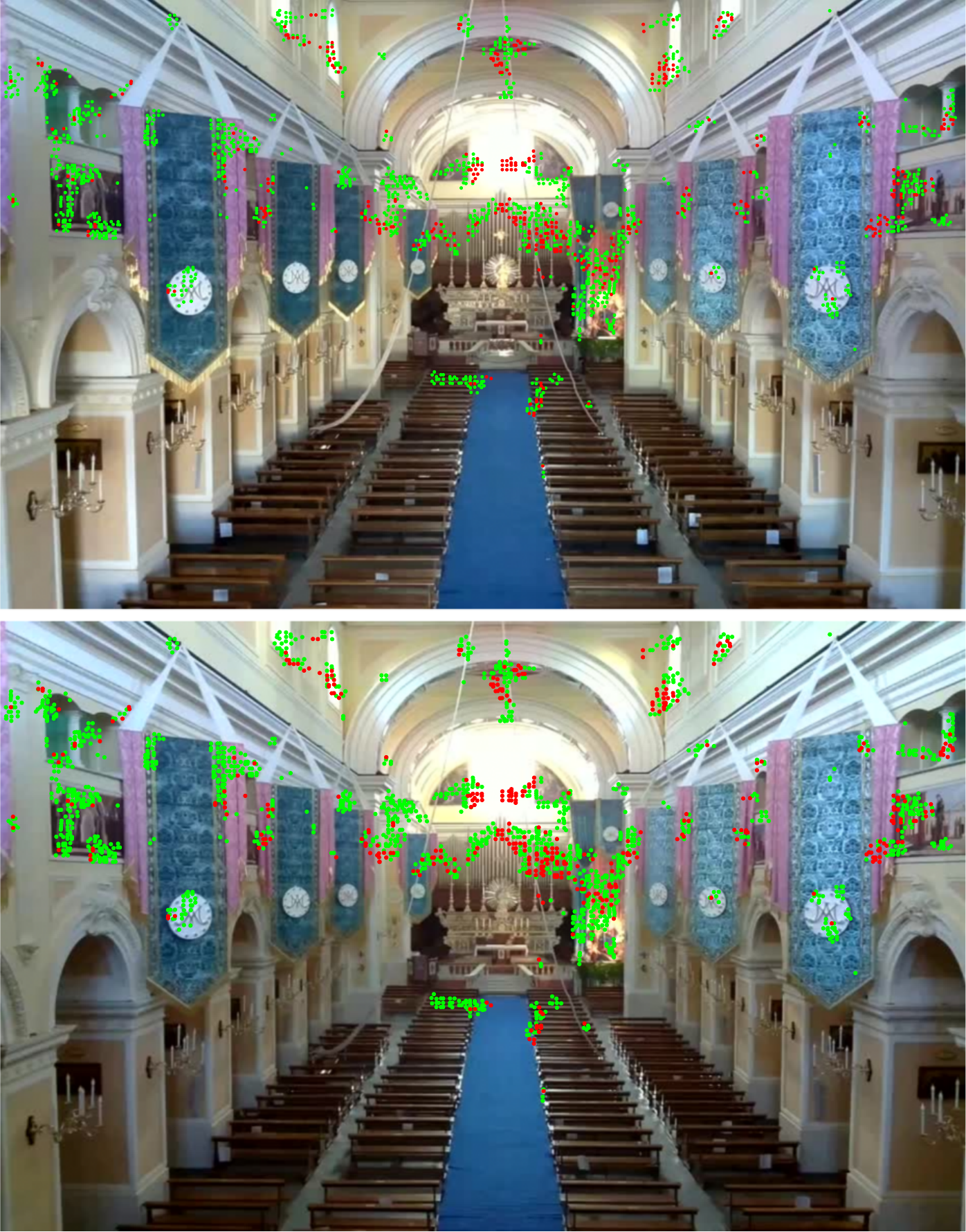}\hspace{0.01\textwidth}%
		\includegraphics[width=0.24\textwidth]{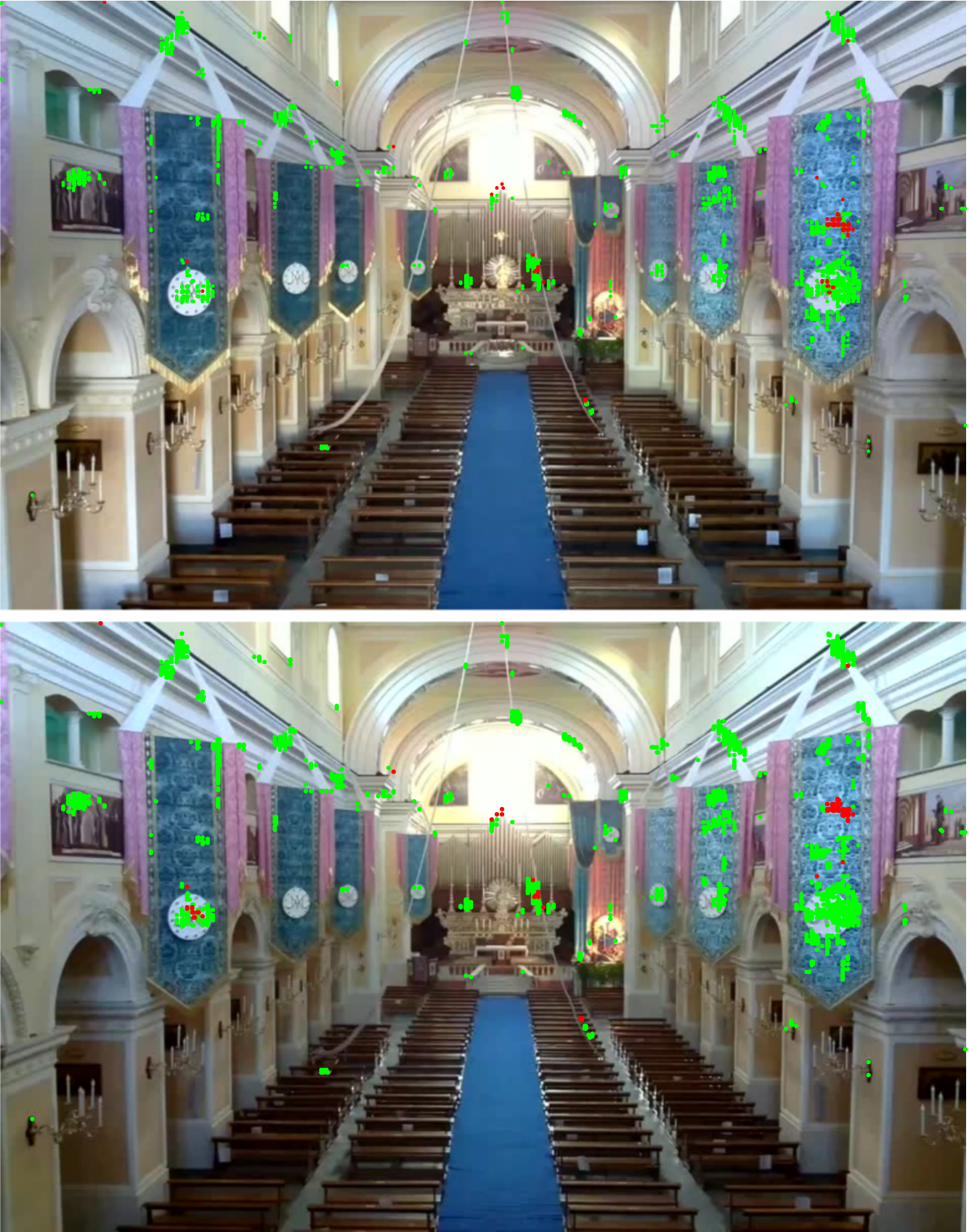}
		\par
		\caption{Qualitative comparison between $\mathbb X$RCNet and SparseNC on HPatches. The green dots represent the 
		correct matches whose errors are within 3 pixels, and red dots the incorrect matches. $\mathbb X$RCNet produces more 
		correct matches out of the top 2000 matches than SparseNC.}
		\label{fig:hpatches_qualitative_sparsenc}
	\end{figure}

	In Fig.~\ref{fig:hpatches_qualitative} and~\ref{fig:hpatches_qualitative_sparsenc}, we select six individual testing pairs to 
	demonstrate that $\mathbb X$RCNet outperforms DualRCNet~\cite{Xinghui_NeurIPS20_DRC} and 
	SparseNC~\cite{Rocco_ECCV20_SparseNC} respectively in terms of the ratio of correct matching predictions of top 2000 
	outputs~\footnote{2000 is a arbitrary number. Following previous works of D2Net, DualRCNet and SparseNC, we also adopt 
	2000 for a fair comparison.}. It can be seen that $\mathbb X$RCNet is capable of producing more reliable results than 
	previous works. In this section we propose a novel evaluation criterion in supplement to the main results along with the 
	qualitative comparison of Fig.~\ref{fig:hpatches_qualitative}. This new evaluation criterion can be formulated as:
	
	\begin{equation}
		\mathbb{N}(\tau^-; \tau^+, ^+, ^-) = \sum_{i}^{N}\mathbf{1}(c^+_i > \tau^+ \cap c^-_i < \tau^-), \label{eq:cherrypicking_bias}
	\end{equation}
	where $c^+_i$ and $c^-_i$ is the ratio of the correct matches out of all the predicted matches of two comparing methods 
	denoted as '$^+$' and '$^-$' respectively. $\tau^+$ and $\tau^-$ are thresholds of the corresponding ratio. $i\in \{1,2,...,N\}$ is 
	the index of the testing pairs in the dataset. $\mathbf{1}(\cdot)$ is a binary indicator function such that $\mathbf{1}(\text 
	{True})=1$ and $\mathbf{1}(\text{False})=0$. As long as there exists the pixel-wise ground truth label, we can always adopt 
	Equation~\ref{eq:cherrypicking_bias} to calculate the number of pairs that favour the '$^+$' method against the '$^-$' method.
	
\begin{figure}[t]
		\centering
		\includegraphics[width=0.33\textwidth]{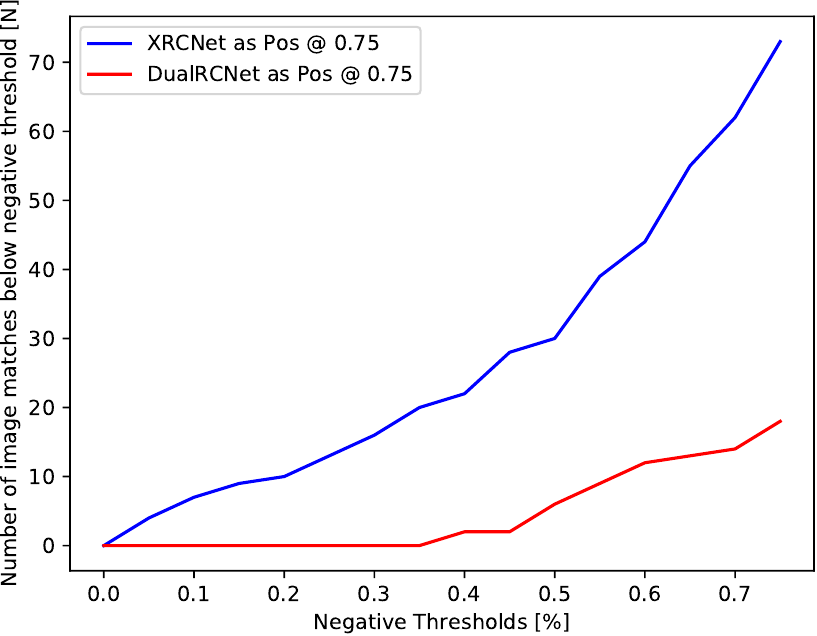}
		\includegraphics[width=0.33\textwidth]{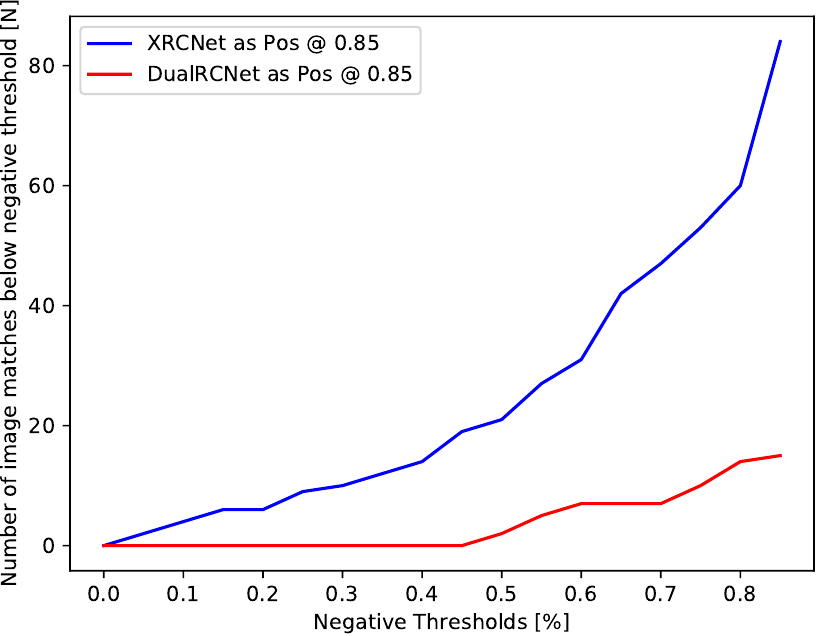}
		\includegraphics[width=0.33\textwidth]{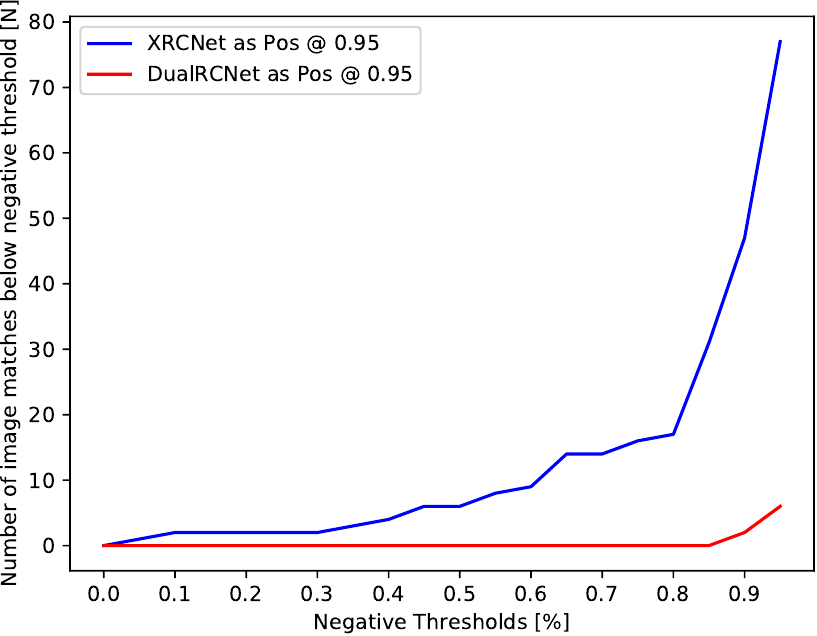}
		\\
		\includegraphics[width=0.33\textwidth]{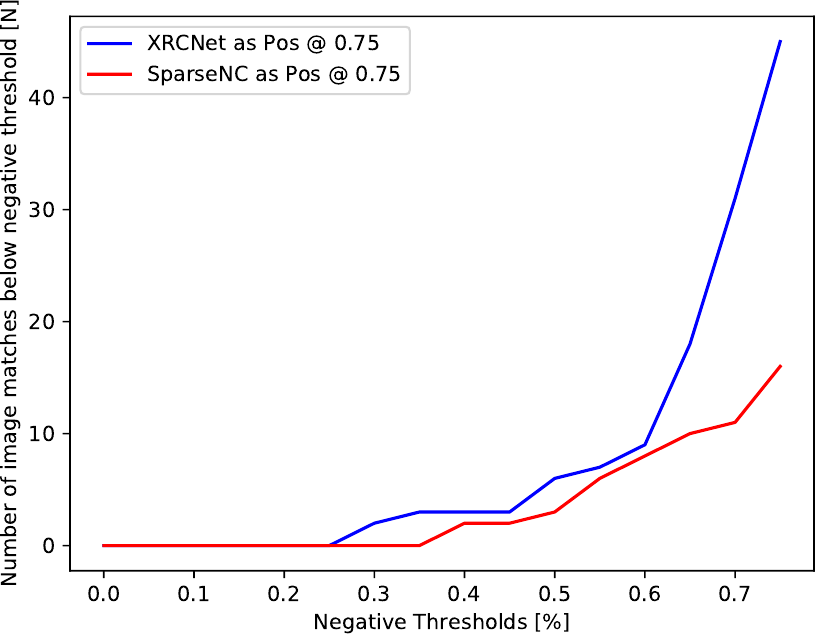}
		\includegraphics[width=0.33\textwidth]{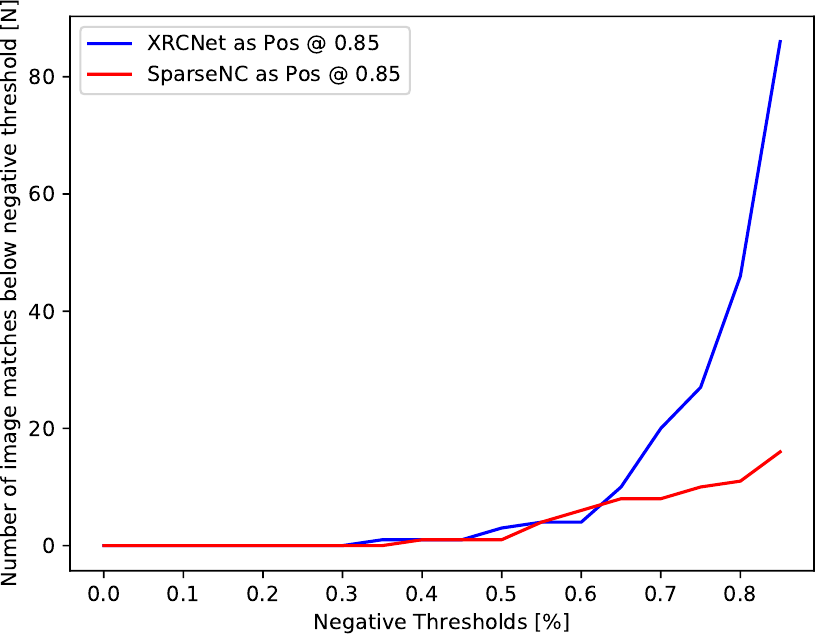}
		\includegraphics[width=0.33\textwidth]{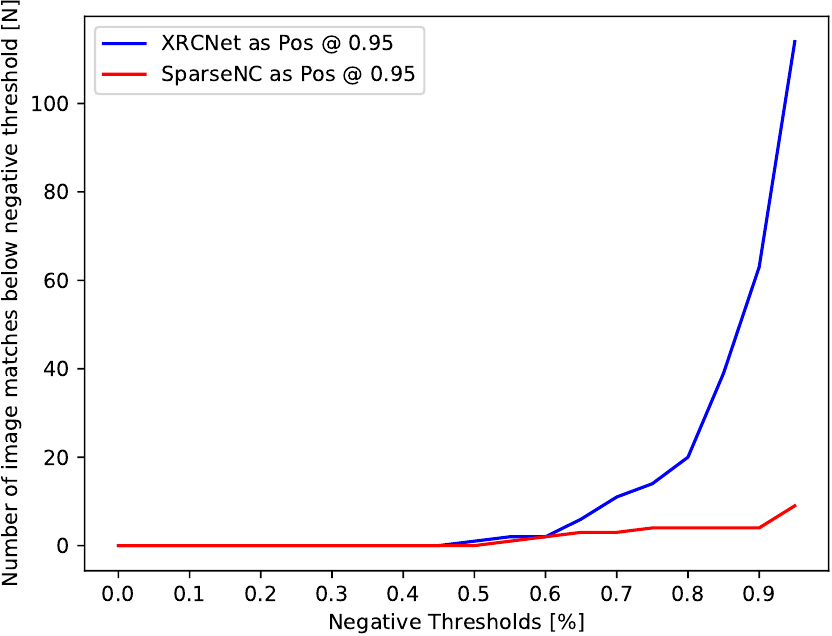}
		\caption{\textbf{Top row:} The comparison of the number of testing pairs that $\mathbb{X}$RCNet outperforms DualRCNet 
		(blue curve) and DualRCNet outperforms $\mathbb X$RCNet using Equation~\ref{eq:cherrypicking_bias}. \textbf{Bottom 
		row:} Similar comparison between $\mathbb{X}$RCNet and SparseNC. For all comparisons, the $\tau^+$ is chosen as 75\%, 
		85\%, and 95\% for both curves. $\tau^-$ in Equation~\ref{eq:cherrypicking_bias} is denoted as the Negative Threshold. 
		'Pos' denotes '+' method and \@0.75 represent the $\tau^+$ ratio threshold.}
		\label{fig:cherrypicking_bias}
	\end{figure}

	Equation~\ref{eq:cherrypicking_bias} measures the number of testing image pairs that favours method $^+$ with respect to 
	ratio $\tau^-$ at a specific positive $\tau^+$. In other words, $\mathbb N(\cdot)$ is a histogram of the testing pairs where the 
	first method $^+$ achieves accuracy higher than the threshold $\tau^+$ but the second method $^-$ achieves accuracy lower 
	than $\tau^-$. In the top row of Fig~\ref{fig:cherrypicking_bias}, we illustrate plotting both the $\mathbb N(\tau^-; \tau^+, 
	{\mathbb X\text{RCNet}}, {\text{DualRCNet}})$ the blue curve vs $\mathbb N(\tau^-; \tau^+, {\text{DualRCNet}}, {\mathbb 
	X\text{RCNet}})$ the red curve over the range of $\tau^- \in [0,\tau^+]$ with a step size of 0.1, and 
	$\tau^+$ is set to $0.75$, $0.85$ and $0.95$, respectively. Similarly, in the bottom row of Fig.~\ref{fig:cherrypicking_bias} we 
	present both $\mathbb{N}(\tau^-; \tau^+,{\mathbb{X}\text{RCNet}}, {\text{SparseNC}})$ as the blue curve vs 
	$\mathbb{N}(\tau^-; \tau^+,{\text{SparseNC}},{\mathbb{X}\text{RCNet}})$ as the red curve. The three sub-figures in 
	Fig.~\ref{fig:cherrypicking_bias} compare the number of testing data that favours $\mathbb X$RCNet against those favouring 
	DualRCNet/SparseNC. It demonstrates that the proposed $\mathbb{X}$RCNet consistently outperforms DualRCNet and 
	SparseNC for all combination of $\tau^-$ and $\tau^+$ values as the number favouring $\mathbb{X}$RCNet is significantly 
	higher than the number favouring DualRCNet/SparseNC.
	
	\section{Qualitative Analysis -- InLoc and Aachen Day-Night}\label{Sec_3}
\begin{figure}[t]
		\centering
		\includegraphics[width=0.49\textwidth]{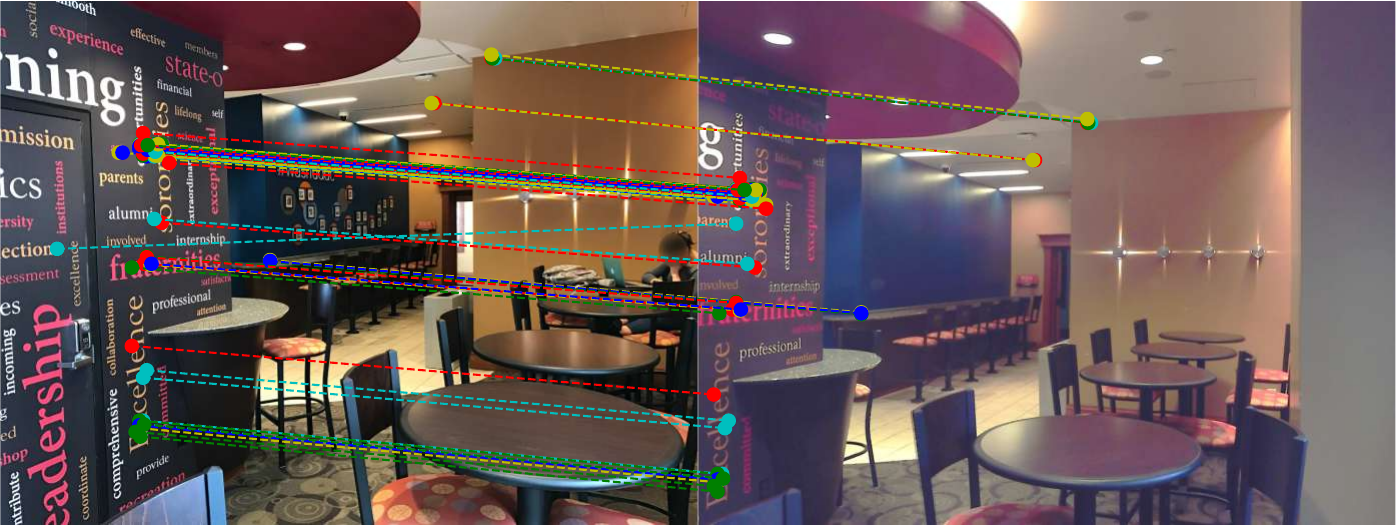}\hspace{0.01\textwidth}%
		\includegraphics[width=0.49\textwidth]{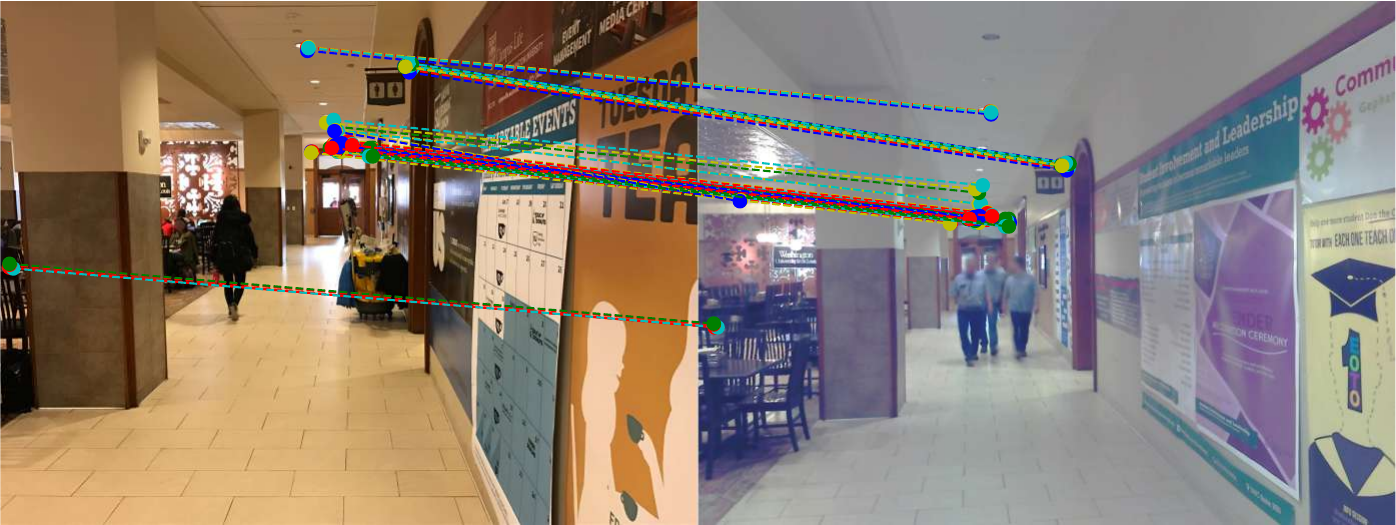}\\[1em]
		\includegraphics[width=0.49\textwidth]{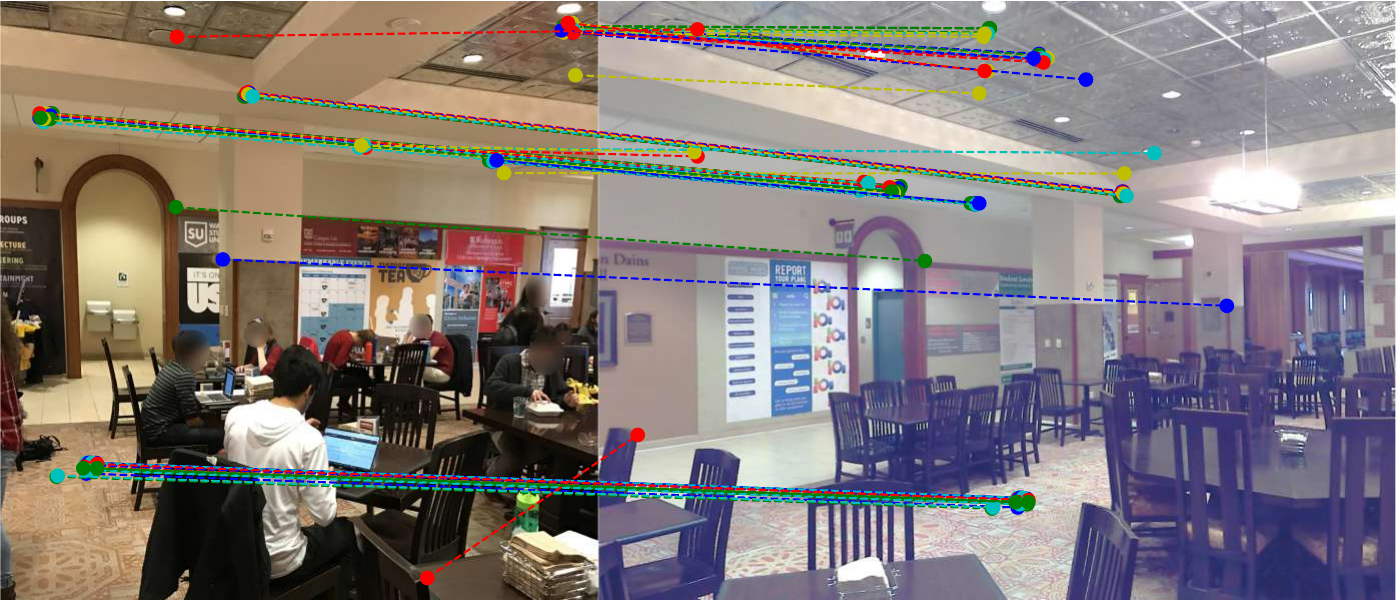}\hspace{0.01\textwidth}%
		\includegraphics[width=0.49\textwidth]{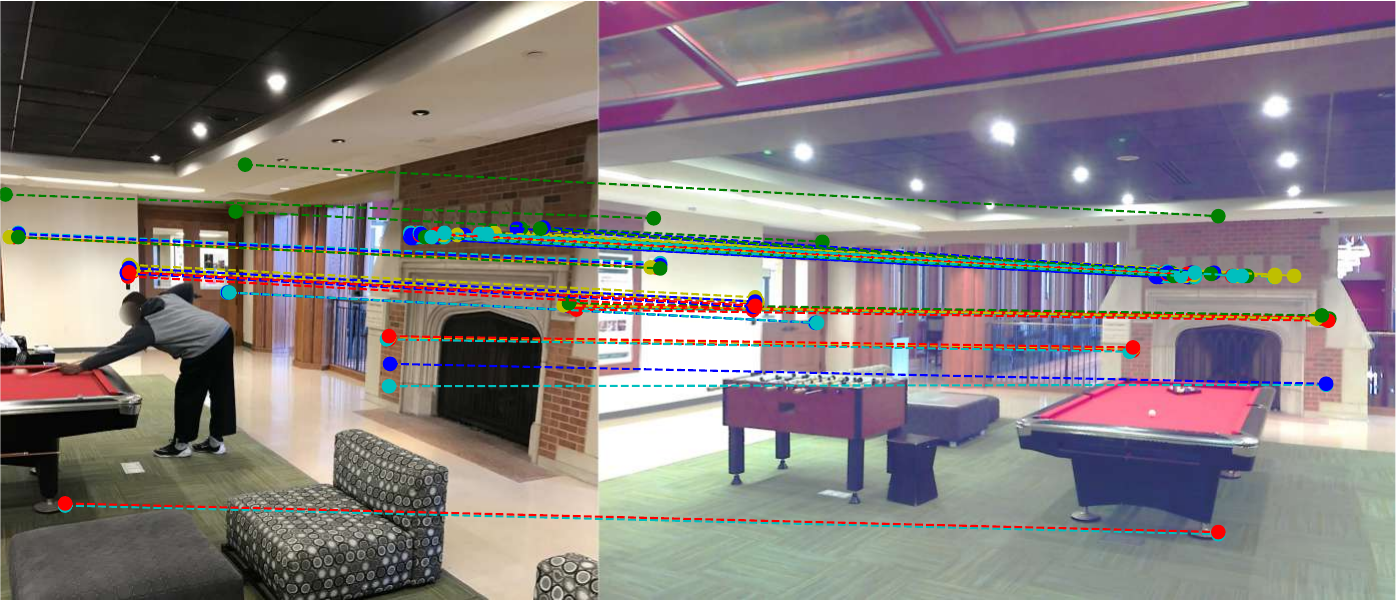}\par
		\caption{Examples of $\mathbb{X}$RCNet running on the Aachen Day-Night dataset - top 2000 matches are displayed. It is 
		worth pointing out the output matches with high reliability scores are heavily clustered in relatively small regions and may 
		overlap each other.}
		\label{fig:inloc_qualitative}
	\end{figure}

	Fig.~\ref{fig:inloc_qualitative} illustrates the performance of $\mathbb{X}$RCNet on the InLoc dataset. Similarly to HPatches, 
	the increase in resolution from 1600 to 3840 (4K) results in better performance. The 4K upsampling resolution for InLoc 
	dataset performs better in terms of relocalisation accuracy than the rest. As mentioned in the main article, we hypothesise this 
	due to the native resolution of testing images in InLoc is much higher than that of HPatches.
	
\begin{figure}[t]
		\centering
		\includegraphics[width=0.19\textwidth]{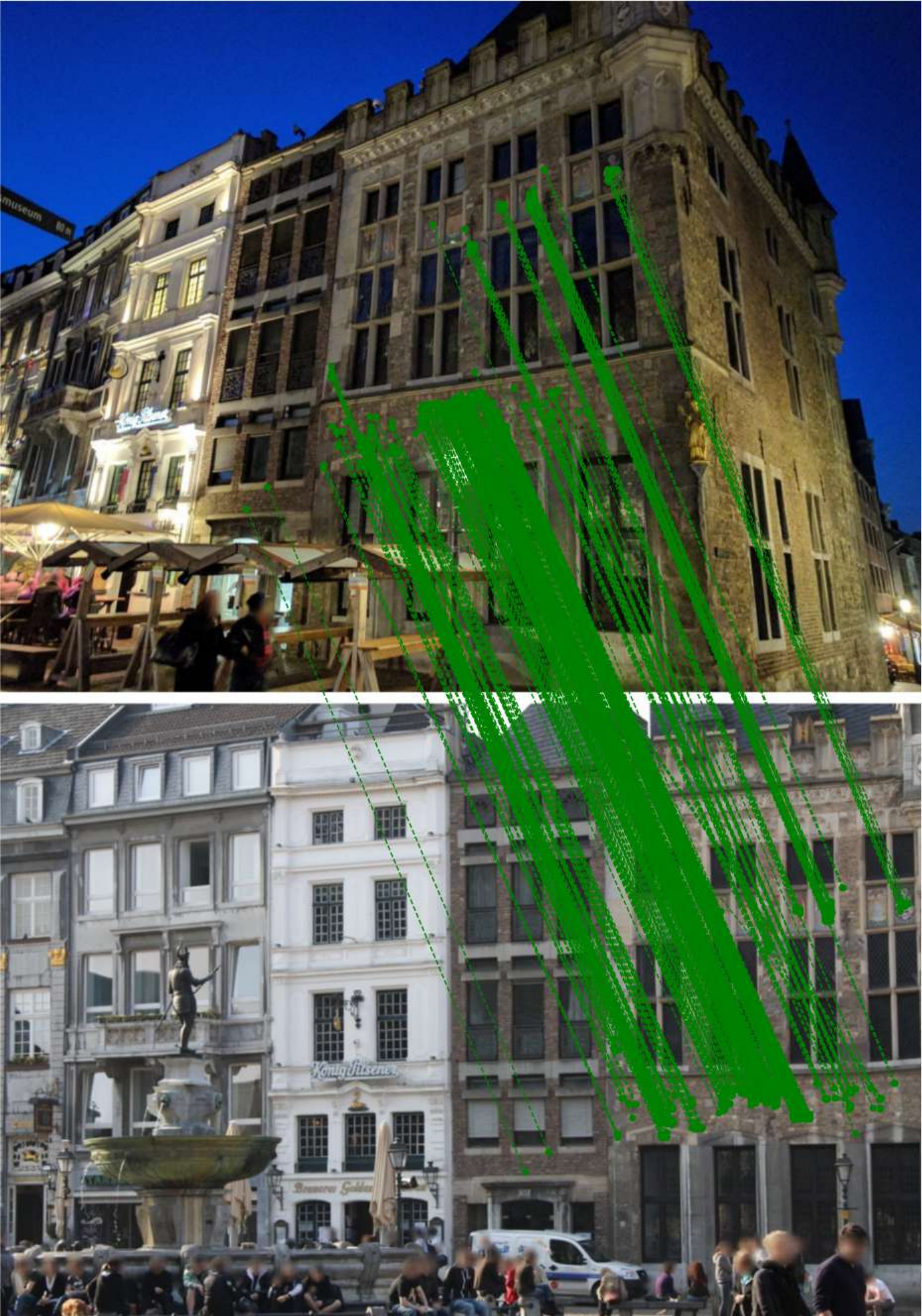}
		\includegraphics[width=0.19\textwidth]{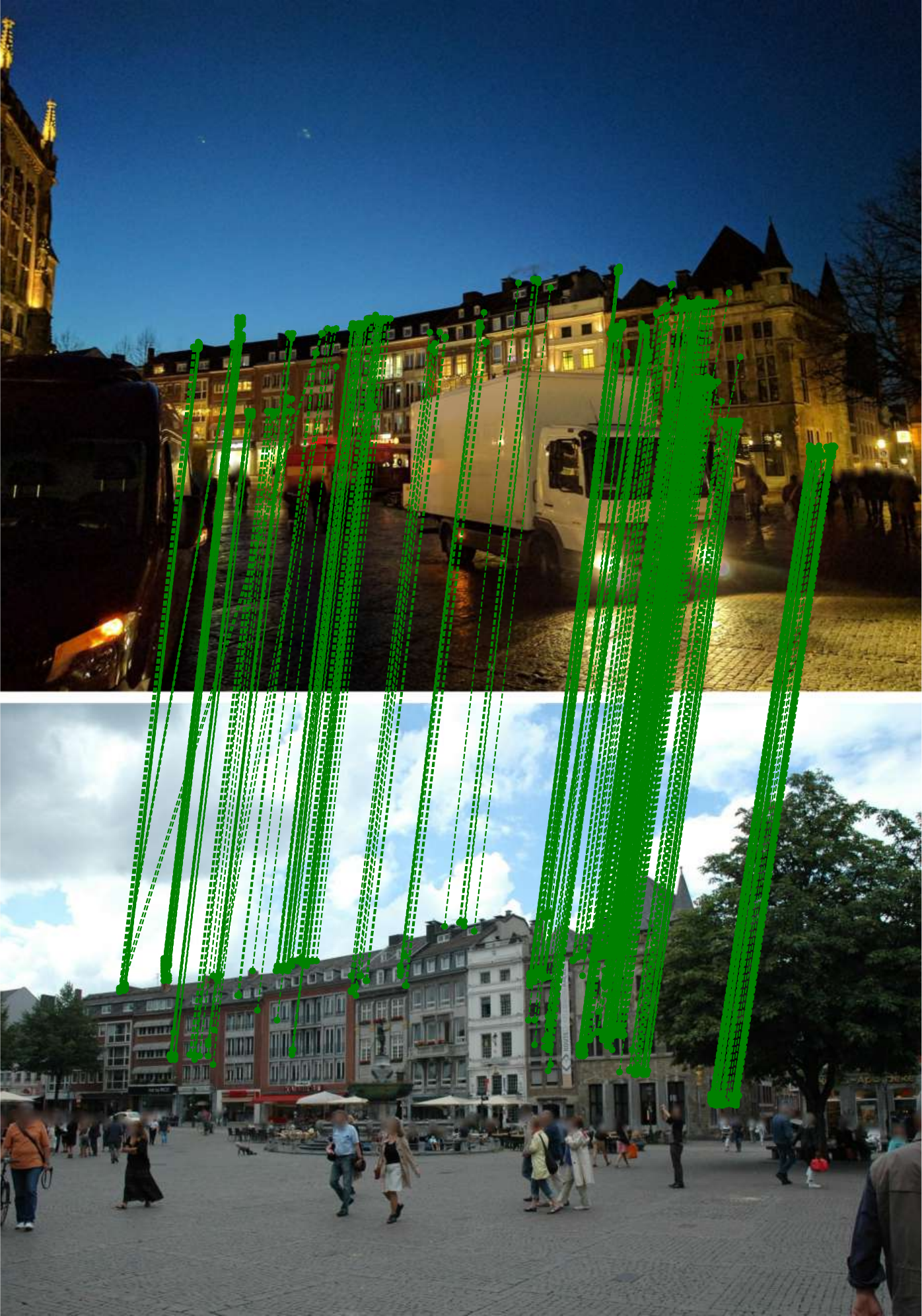}
		\includegraphics[width=0.19\textwidth]{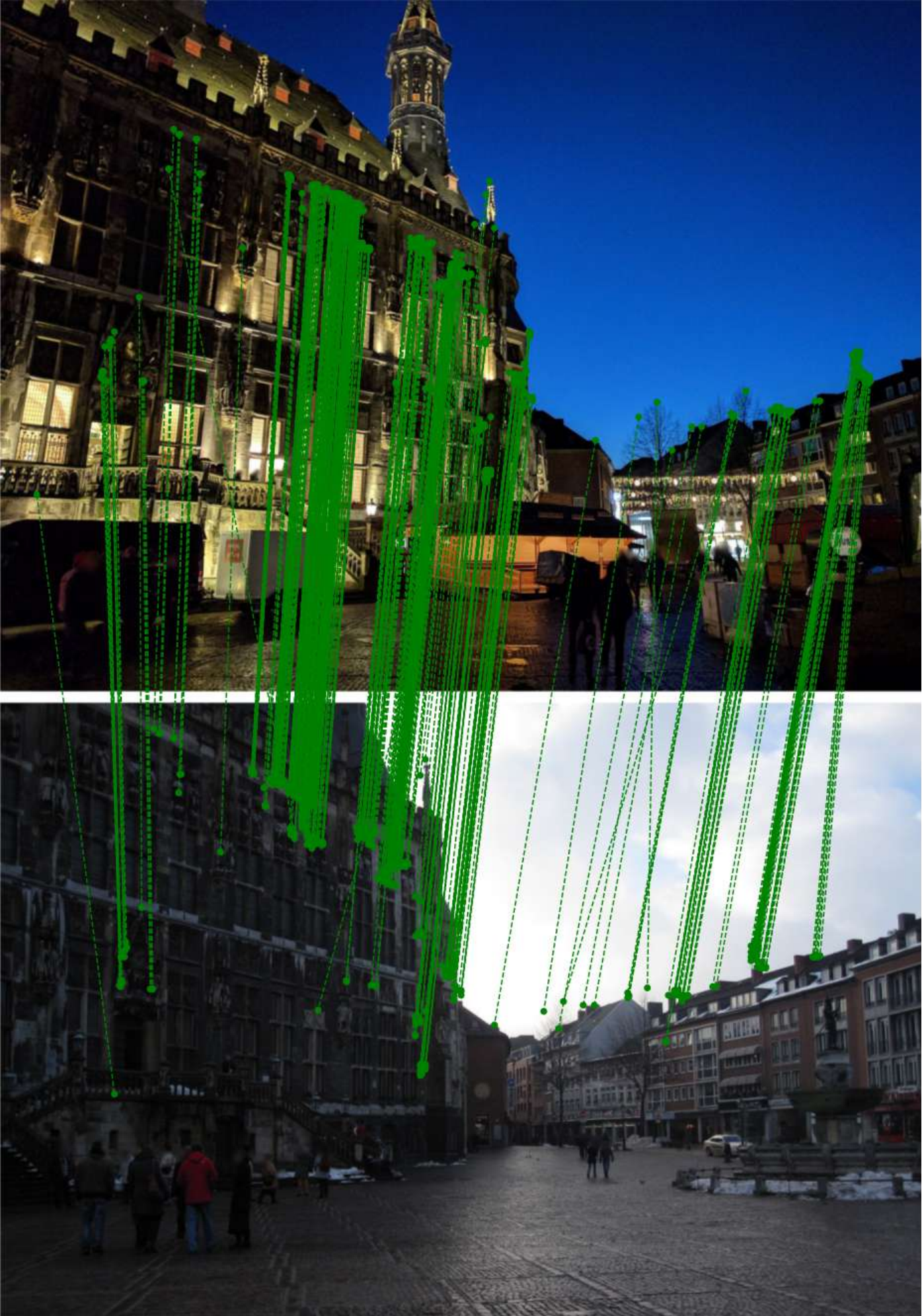}
		\includegraphics[width=0.19\textwidth]{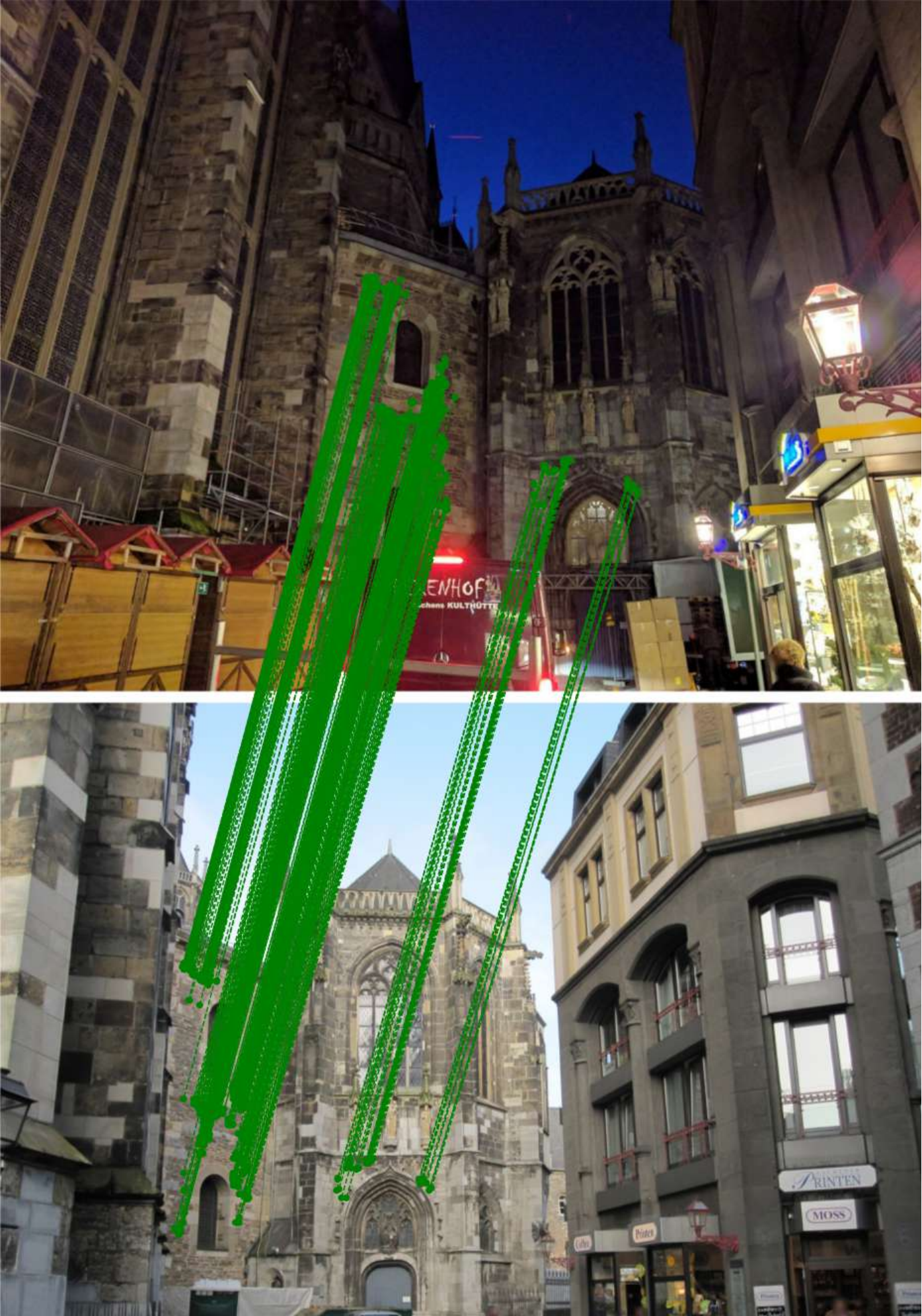}
		\includegraphics[width=0.158\textwidth]{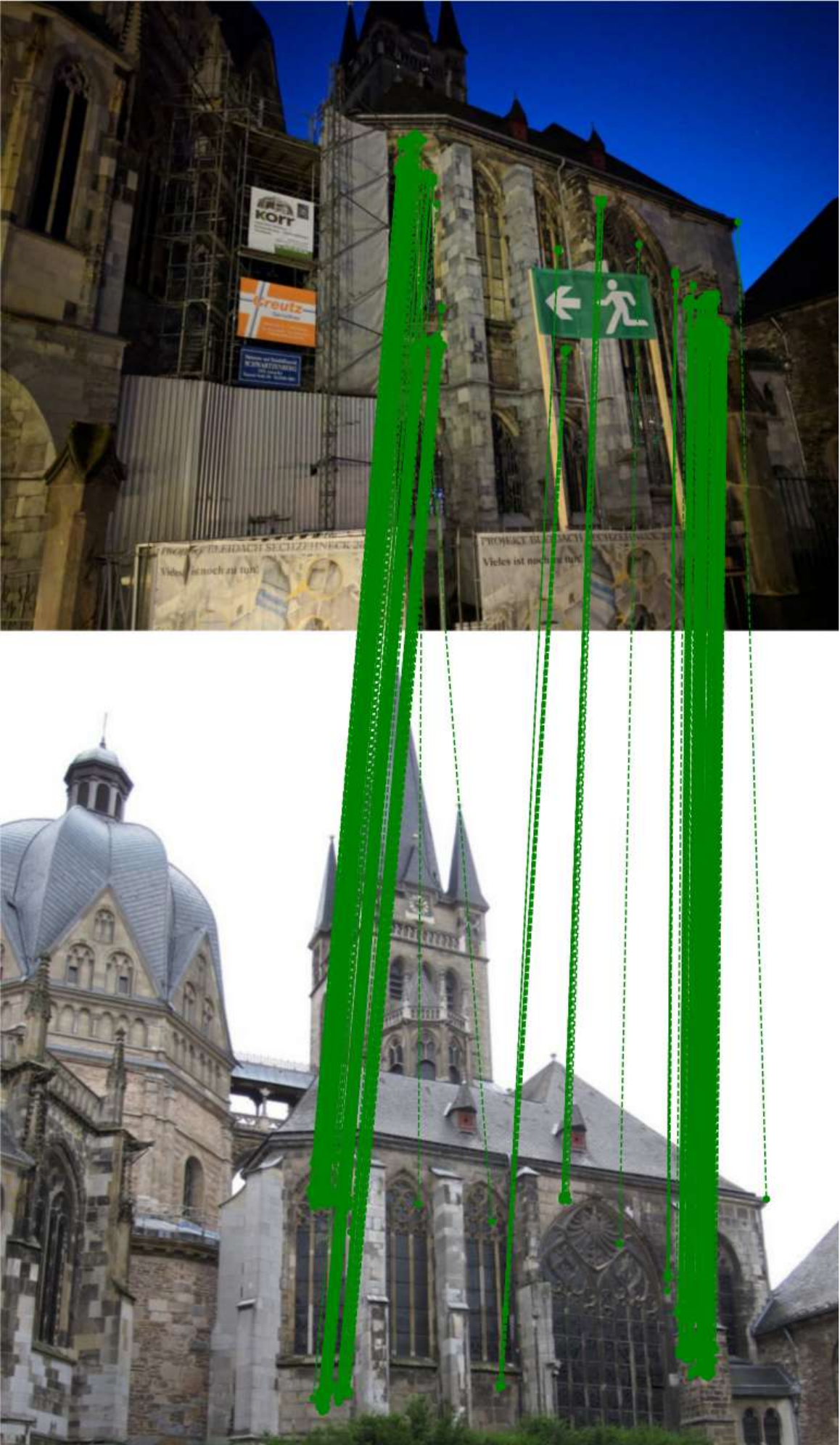}
		\caption{Examples of $\mathbb{X}$RCNet running on the Aachen Day-Night dataset - top 2000 matches are displayed. It is 
		worth pointing out the output matches with high reliability scores are heavily clustered in relatively small regions and may 
		overlap each other.}
		\label{fig:aachen_qualitative}
	\end{figure}

	Fig.~\ref{fig:aachen_qualitative} shows visual examples of the proposed model evaluated on the Aachen Day-Night 
	dataset~\cite{Sattler_CVPR17_LargeScaleLocalisation}.
	
	\section{3D Reconstruction Using Dense Correspondences}\label{Sec_4}
\begin{figure}[htb]
		\centering
		\includegraphics[width=1.0\textwidth, trim = 0cm 1cm 0cm 0cm,clip]{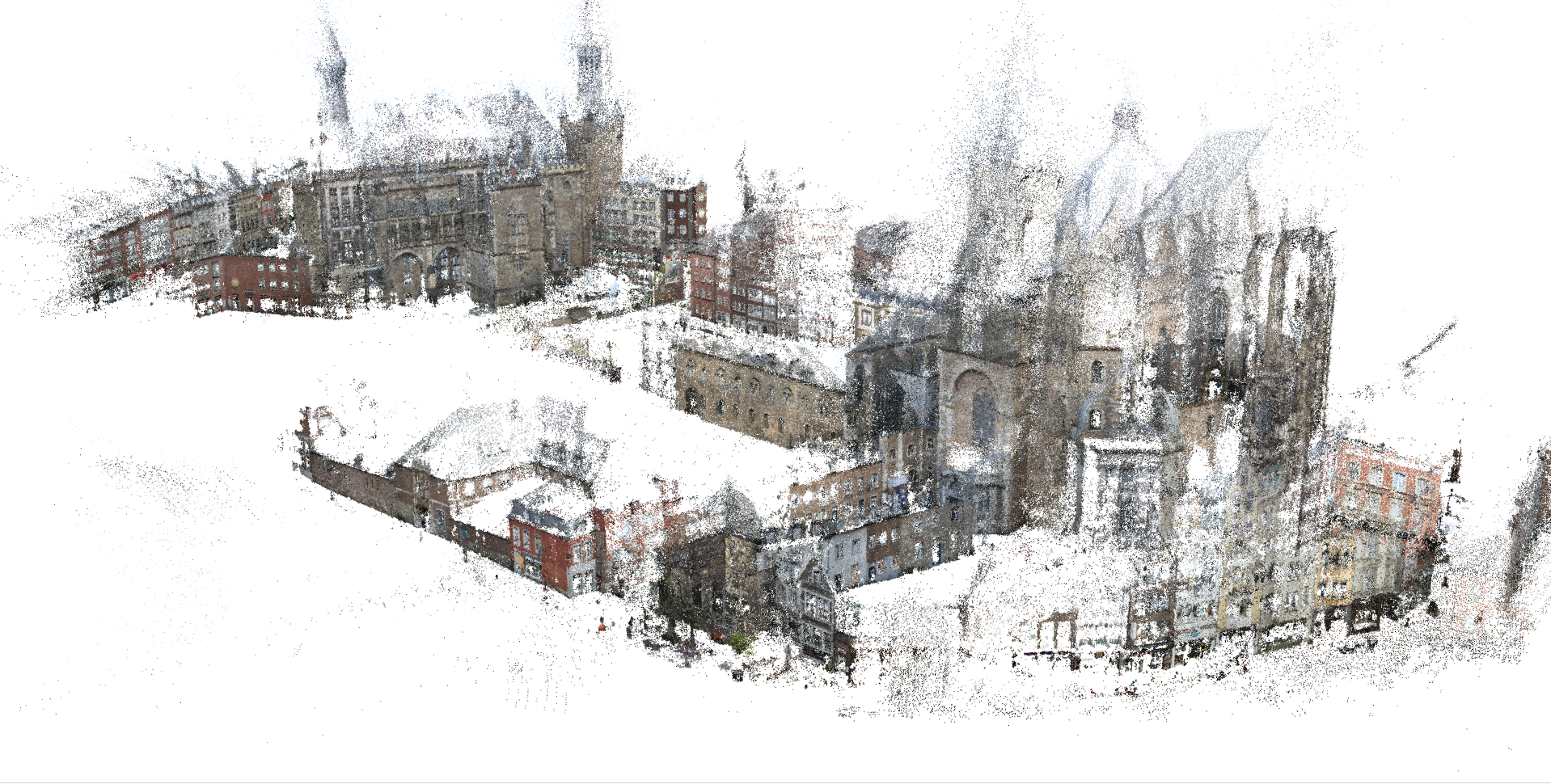}
		\includegraphics[width=1.0\textwidth, trim = 4cm 0cm 4cm 4cm,clip]{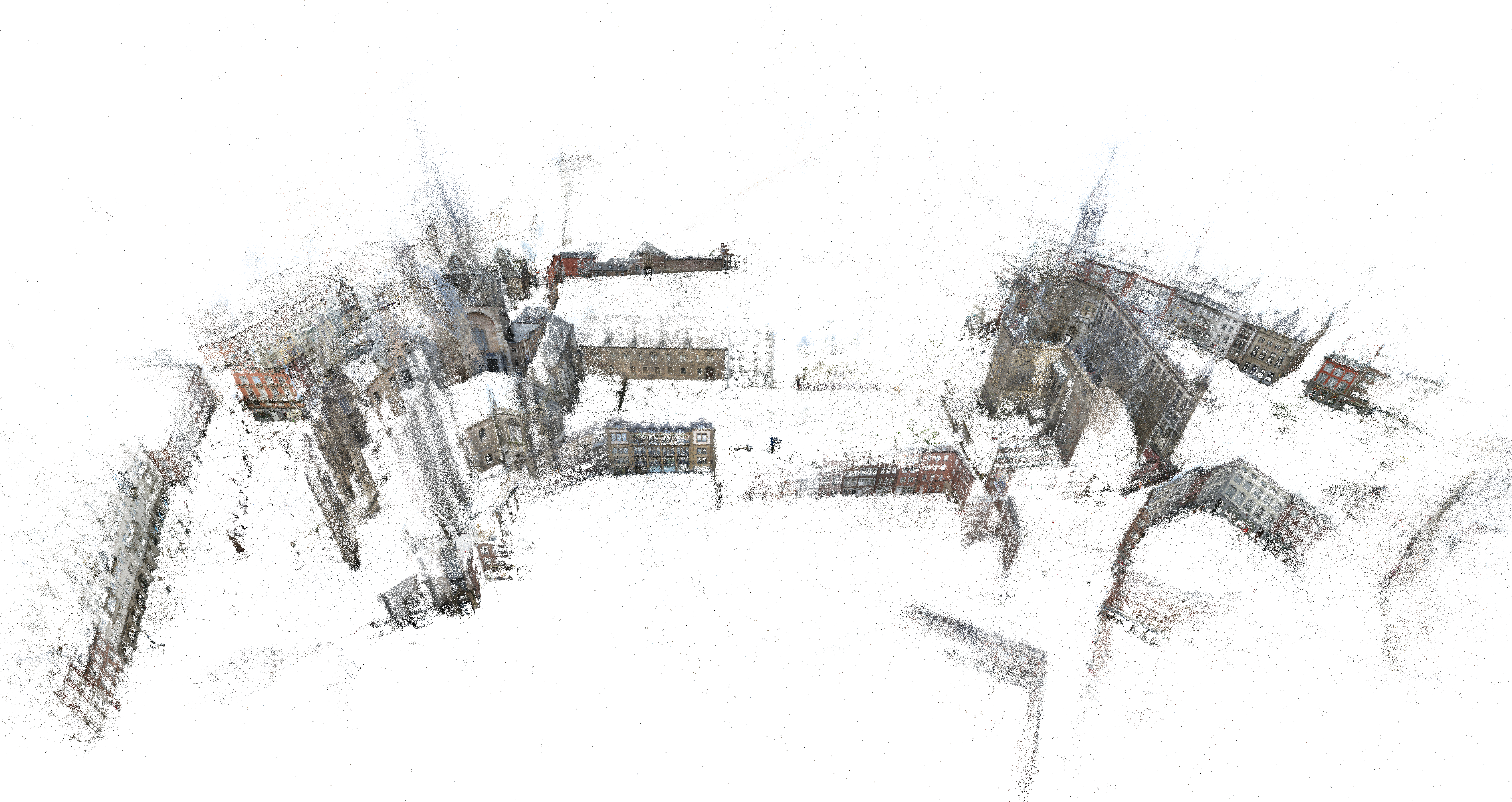}
		\caption{3D model reconstructed using correspondences obtained by $\mathbb{X}$RCNet for the Aachen Day-Night 
		dataset.}
		\label{fig:3D_recon_XRCNet}
	\end{figure}

\begin{figure}[htb]
		\centering
		\includegraphics[width=1.0\textwidth, trim = 0cm 0cm 0cm 0cm,clip]{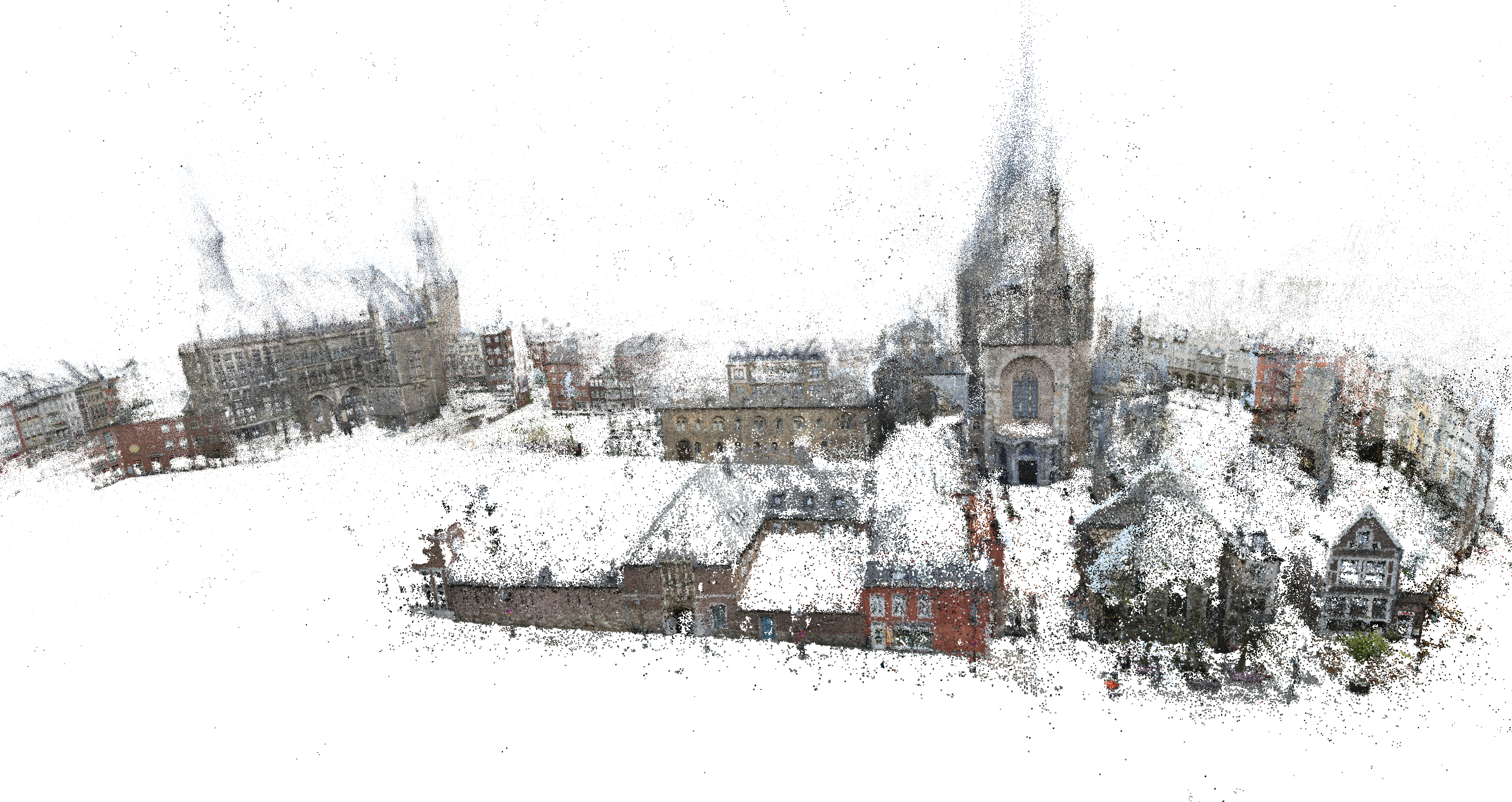}
		\includegraphics[width=1.0\textwidth, trim = 0cm 0cm 0cm 0cm,clip]{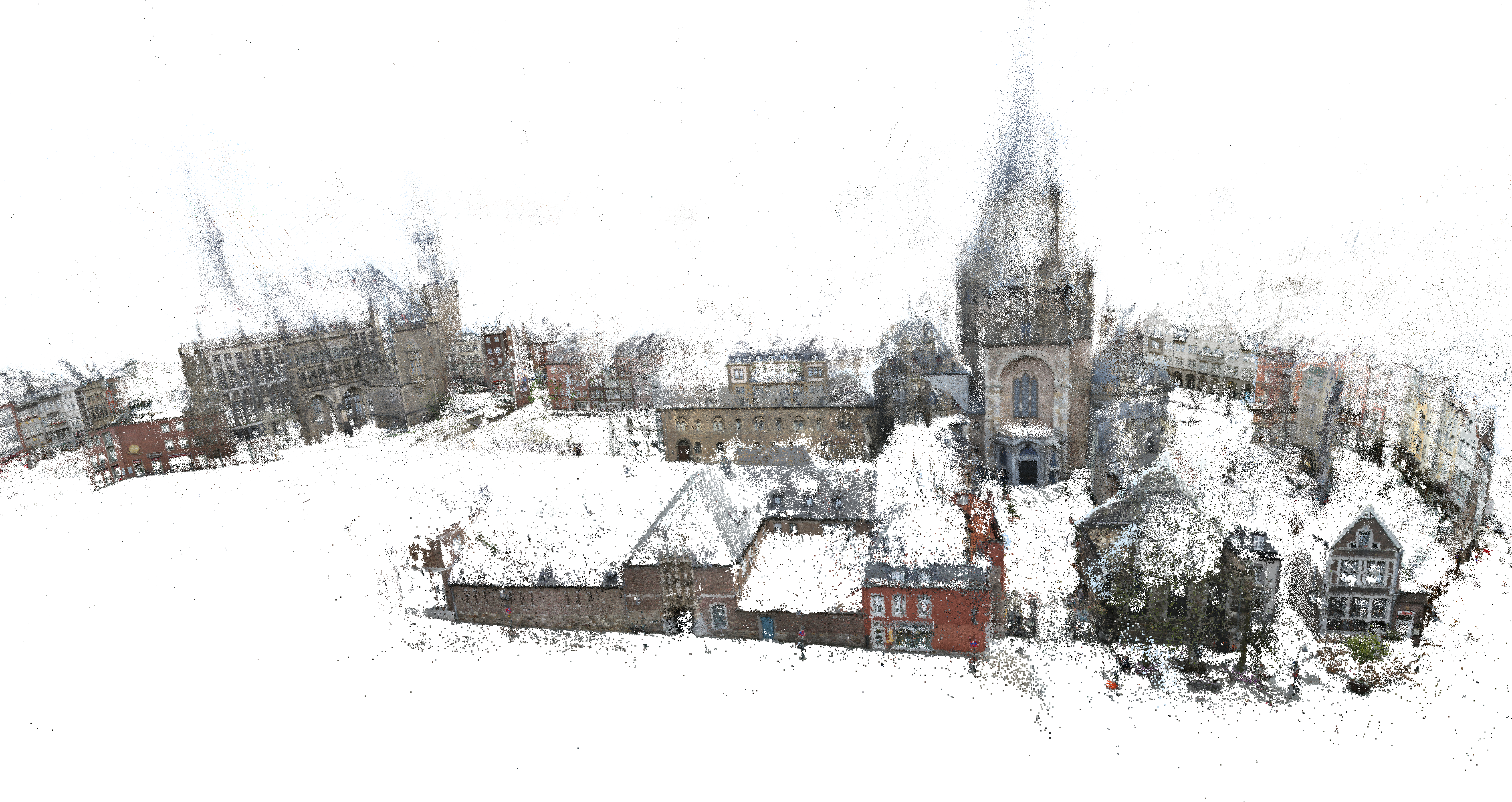}
		\caption{Qualitative comparison of $\mathbb X$RCNet (top) and DualRCNet (bottom) 3D model reconstructions on the 
		Aachen Day-Night dataset.}
		\label{fig:xrcnet_vs_dualrc_aachen}
	\end{figure}

	To demonstrate a potential application of using the correspondence network, we plot the 3D point cloud reconstructed using 
	the $\mathbb{X}$RCNet in Fig.~\ref{fig:3D_recon_XRCNet} on the Aachen Day-Night reference images. We also compare the 
	quality of the 3D reconstruction using $\mathbb{X}$RCNet and DualRCNet in Fig.~\ref{fig:xrcnet_vs_dualrc_aachen}. 
	It can be seen that the quality of the reconstructed models are fairly close for the two methods.
	
	\section{Code}\label{Sec_5}
	We include code as part of the supplementary material to allow for reproducibility of the results as well as retraining the 
	models. Our code is made publicly available here: \url{https://xyz-r-d.github.io/xrcnet}.
	
	\clearpage
	{\small
		\bibliographystyle{ieee_fullname}
		\bibliography{reference}
	}